\documentclass[letterpaper,twocolumn]{article}
\pdfoutput=1
\usepackage{arxiv}
\usepackage[margin=1in]{geometry}

\renewenvironment{abstract}{\centerline{\large\bf Abstract}\vspace{0.5ex}\begin{quote}}{\par\end{quote}\vskip 1ex} 

\usepackage[round, comma]{natbib}

\usepackage[utf8]{inputenc} 
\usepackage[T1]{fontenc}    
\usepackage{hyperref}       
\usepackage{url}            
\usepackage{booktabs}       
\usepackage{amsfonts}       
\usepackage{nicefrac}       
\usepackage{microtype}      
\usepackage[pdftex]{graphicx}
\usepackage{subcaption}
\usepackage{amsmath}
\usepackage{array}
\usepackage{amssymb} 
\usepackage{float}
\usepackage{bbm}
\usepackage[toc,page,titletoc]{appendix}

\newcommand{\R}{\mathbb{R}}

\newcommand*{\myspur}[1]{\mathrm{Tr}\!\left(#1\right)}
\newcommand*{\myexp}[1]{\mathrm{exp}\!\left(#1\right)}
\newcommand*{\mylog}[1]{\mathrm{log}\!\left(#1\right)}

\newcommand*{\mynorm}[1]{\left\Vert #1\right\Vert}
\newcommand*\mean[1]{\overline{#1}}
\let\arrowVec\overrightarrow
\let\vec\mathbf
\newcommand{\argmin}{\mathop{\mathrm{arg\,min}}}

\graphicspath{{./Images/}}

\usepackage{times}

\title{Uncertainty Estimates for Ordinal Embeddings}

\author{
	Michael Lohaus$^1$
	\and
	Philipp Hennig$^{1,2}$
	\and
	Ulrike von Luxburg$^{1,2}$
	\affiliations
	$^1$University of T\"ubingen, Germany\\
	$^2$Max-Planck-Institute for Intelligent Systems, T\"ubingen, Germany
	\emails
	\{michael.lohaus, philipp.hennig, ulrike.luxburg\}@uni-tuebingen.de
}

\begin{document}

\maketitle

\begin{abstract}
To investigate objects without a describable notion of distance, one can gather ordinal information by asking triplet comparisons of the form ``Is object $x$ closer to $y$ or is $x$ closer to $z$?'' In order to learn from such data, the objects are typically embedded in a Euclidean space while satisfying as many triplet comparisons as possible. 
In this paper, we introduce empirical uncertainty estimates for standard embedding algorithms when few noisy triplets are available, using a bootstrap and a Bayesian approach. In particular, simulations show that these estimates are well calibrated and can serve to select embedding parameters or to quantify uncertainty in scientific applications. 
\end{abstract}

\section{Introduction}
In a typical machine learning problem, we are given a data set of examples $\mathcal{D}$ and a (dis)similarity function $\delta$ that quantifies the distance between the objects in $\mathcal{D}$. The inductive bias assumption of machine learning is then that similar objects should lead to a similar output. However, in poorly specified tasks such as clustering the biographies of celebrities, an obvious computable dissimilarity function does not exist. Nonetheless, humans often possess intuitive information about which objects should be dissimilar. While it might be difficult to access this information by quantitative questions (``On a scale from $1$ to $10$, how similar are celebrities $x$ and $y$?''), it might be revealed using comparison-based questions of the type ``Is object $x$ closer to $y$ or closer to $z$?'' The answer to this distance comparison is called a \emph{triplet}. The clear advantage of such an approach is that it removes the issues of a subjective scale across different participants. A common application for such ordinal data is crowd sourcing, where many untrained workers complete given tasks \citep{tamuz11, heikinheimo13,ukkonen15}. Ordinal information emerges for example as real data in search-engine query logs  \citep{schultz04}. Given a list of search results, a document $x$, on which a user clicked, is closer to a second clicked document $y$ than to a document $z$, on which the user did not click. 

The standard approach to deal with ordinal data in machine learning is to create Euclidean representations of all data points using ordinal embedding.  In the classic literature, this approach is called non-metric multidimensional scaling (NMDS) \citep{borgGroenen05} and has its origins in the analysis of psychometric data by \citet{shepard62} and \citet{kruskal64}. In NMDS the objects are embedded into a low-dimensional Euclidean space while fulfilling as many triplet answers as possible. In recent years, such embedding methods have been extensively studied, and various optimization problems were proposed \citep{agarwal07,terada14,amid15,jain16,anderton18}, also including probabilistic models \citep{tamuz11,maaten12,karaletsos16}.

Theoretical aspects of ordinal embeddings have been studied for the scenario where the original data comes from a Euclidean space of dimension $d$. The consistency of ordinal embeddings in the large-sample limit $n \rightarrow \infty$ as well as the consistency for the finite sample case up to a similarity transformation have been established \citep{kleindessner14,arias-castro17}. It is also known that $\Omega\! \left( dn \mylog{n} \right)$ actively chosen triplets are necessary to learn an embedding of $n$ points in $\R^d$ \citep{jamieson11,jain16} .  As of now, there has been no theoretical analysis on the distortion of embeddings that are constructed from a set of much fewer than $\Omega\! \left( dn \mylog{n} \right)$ triplets. In this paper, we develop empirical methods that provide uncertainty estimates to characterize how certain a given embedding method is about its result when only few and possibly noisy training triplets are available.

We present two methods to obtain uncertainty estimates: one is based on a bootstrap approach, the other on a Bayesian approach. In both cases, we generate a distribution over embeddings which can then be used to compute the uncertainty about the answer of a triplet comparison or the point location. These uncertainty estimates improve upon ordinal embedding and address its limitations. In particular, we apply our uncertainty estimates to estimate the embedding dimension or to quantify uncertainty in psychophysics applications. In the experiments we examine whether the triplet uncertainty estimates are well calibrated. Lastly, we use the estimates to improve active selection of triplets. We find that our uncertainty estimates provide a valuable add-on to judge the quality of ordinal embeddings. 

\section{Setup and background on ordinal embedding}

\subsection{Setup}
Let $\left(\mathcal{X},\delta\right)$ be a metric space and $\mathcal{D} = \left\lbrace \xi_1,...\xi_n \right\rbrace \subset \mathcal{X}$ a finite data set. We consider a latent dissimilarity function $\delta: \mathcal{X} \times \mathcal{X} \rightarrow \R_0^+$ on $\mathcal{D}$ and abbreviate $\delta\!\left( \xi_i, \xi_j\right) =: \delta_{ij}$. In the following, we assume that no explicit dissimilarity values $\delta_{ij}$ nor any other representations of the objects in $\mathcal{D}$ are given to us. Instead, for three distinct points $\xi_i$, $\xi_j$ and $\xi_l$, we can observe the (possibly noisy) answer to the comparison: ``Is $\delta_{ij} < \delta_{il}$ or is $\delta_{ij} > \delta_{il}$?'' We call the answer to this question a \emph{triplet}. Formally, a triplet is an ordered set of three points $\left(i,j,l\right)$, encoding that the answer is ``$\delta_{ij} < \delta_{il}$''. Similarly, $\left(i,l,j\right)$ denotes that the answer is ``$ \delta_{il}< \delta_{ij}$''. This answer can be noisy in the sense that it might be wrong compared to the ground truth dissimilarity. However, this ground truth is considered unknown and we only observe a set $\mathcal{S}$ of noisy triplets, which we refer to as the training set.

A Euclidean embedding of the data set $\mathcal{D} = \{\xi_1, ..., \xi_n\}$ consists of $n$ points $\vec{x}_1, ..., \vec{x}_n \in \R^d$, written more compactly as $\vec{X} \in \R^{n \times d}$. The parameter $d$ is called the embedding dimension. 
The function $\rho: \R^d \times \R^d \rightarrow \R_0^+$ denotes the Euclidean distance function, abbreviated as $\rho_{ij} := \rho\!\left( \vec{x}_i, \vec{x}_j\right) := \mynorm{\vec{x}_i - \vec{x}_j}_2 $. Let $\vec{D} := \vec{D}\!\left(\vec{X}\right)$ be the Euclidean distance matrix corresponding to the embedding $\vec{X}$. 

Given a (noisy) triplet set $\mathcal{S}$, the goal of \emph{ordinal embedding} is to find a representation of the data set $\mathcal{D}$ in a low-dimensional space $\R^d$ that recovers as many triplets as possible, that is we want to construct points $\vec{x}_1, ..., \vec{x}_n \in \R^d$ such that 
\begin{equation*}
\delta\!\left(\xi_i,\xi_j\right) < \delta\!\left(\xi_i,\xi_l\right) \Rightarrow
\rho\!\left(\vec{x}_i,\vec{x}_j\right) < \rho\!\left(\vec{x}_i,\vec{x}_l\right).
\end{equation*}

\subsection{Ordinal embedding methods}
The goal of this paper is to provide uncertainty estimates for ordinal embeddings and improve upon existing embedding methods. In the following, we briefly recall two popular embedding methods.

\paragraph{Crowd Kernel.}
\citet{tamuz11} assume an explicit noise model by which the observed triplets have been generated. Given three points $\vec{x}_i, \vec{x}_j, \vec{x}_l$, the likelihood to obtain the triplet answer $(i,j,l)$ is given by 
\begin{equation}\label{eq:CKprobability}
p_{ijl} = \frac{\rho_{il}^2 + \mu}{\rho_{ij}^2+\rho_{il}^2+ 2\mu},
\end{equation}
where $\mu$ is a parameter for regularization and numerical stability. 
A high value of $p_{ijl}$ indicates that a triplet $(i,j,l)$ is well modeled by the current embedding. 
Given a noisy set of triplets $\mathcal{S}$, the final embedding $\vec{X}$ is obtained by minimizing the negative log-likelihood $\sum\limits_{\left( i,j,l \right) \in \mathcal{S}} \hspace{-8pt}-\mylog{p_{ijl}}$.

{\bf Stochastic Triplet Embedding} (STE, \citealp{maaten12} ) suggests a more \emph{local} variant where large violations receive nearly constant penalties and clearly satisfied triplets induce a nearly constant reward. Their Gaussian noise model leads to the triplet likelihood 
\begin{equation}\label{eq:STELikelihood}
p_{ijl} = \frac{\myexp{-\rho_{ij}^2}}{\myexp{-\rho_{ij}^2}+\myexp{-\rho_{il}^2}}.
\end{equation}
In the same paper, the authors also suggest a more robust version of STE that replaces the Gaussian distribution by the more heavy-tailed $t$-distribution, leading to what is called $t$-STE (see supplement). The final embedding is constructed by minimizing the negative log-likelihood.

\section{Computing uncertainty estimates of pairwise comparisons}

Assume we are given a data set $\mathcal{D}$ of $n$ abstract points, a noisy triplet set $\mathcal{S}$ and an embedding algorithm. The goal is to generate a set of embedding samples and use the variance of such samples to capture uncertainty over point locations or triplet answers. We consider two approaches to generate embeddings in order to compute uncertainty estimates for ordinal embedding algorithms.

In a {\bf Bayesian approach}, we use a prior distribution, either over embeddings or over distance matrices. As the second ingredient we use a likelihood function, encoding the probability of observing a certain triplet given an embedding or distances. Sampling from the corresponding posterior leads to a set of embeddings, or a set of distance matrices, respectively. 

Our {\bf bootstrap approach} proceeds like the following: we first subsample triplets (not points!) and secondly, we embed the points based on these triplets by the given embedding algorithm. Each repetition leads to a new embedding of the $n$ data points. As in the Bayesian approach, we can use this set of embeddings to evaluate uncertainties. 
	
These uncertainty estimates capture uncertainties caused by the limited number of observed triplets and their possibly noisy nature. By construction, they cannot account for systematic model bias, which could be introduced by an inapt choice of embedding algorithm in the bootstrap, or a bad choice of likelihood in the Bayesian approach. 
	
\subsection{Bayesian approach for sampling embeddings}
A Bayesian model is defined by a generative model that specifies prior and likelihood. We consider two variants. Our first attempt is to consider a prior over distance matrices. In order to calculate an uncertainty estimate for triplets, it is preferable to know directly the uncertainty of the two distances involved. However, this turns out to be challenging due to the specific properties of Euclidean distance matrices that would need to be modeled. Our second approach specifies a prior over embeddings. This turns out to be much more manageable despite the ``detour'' via embeddings to distances. In both cases, we use \emph{Gaussian} distributions for the prior. Gaussians are a frequent choice of prior in the Bayesian literature. They are closely related to $\ell_2$-norm regularizers in the statistical literature \citep{2018arXiv180702582K} and amount to a similar set of assumptions and restrictions. 

A {\bf prior over the distance matrix $\vec{D}$} is supposed to incorporate as much prior knowledge about Euclidean distance matrices as possible, yet it should be easy to handle.  Simple prior knowledge about Euclidean distance matrices is that they are symmetric, all entries are non-negative, diagonal entries are zero, and the entries fulfill the triangle inequality. These properties alone are not sufficient to characterize Euclidean distance matrices. We would need to incorporate even more complicated conditions \citep{dattorro05}. However, it is hard to find a distribution of matrices that has its support exactly on Euclidean distance matrices. See the supplementary material for a first approach that employs a multivariate Gaussian prior over distance matrices to encode symmetry. It is non-trivial to encode more specific properties of distance matrices in an expressible prior. Because Bayesian models based on this simplistic but tractable prior did not behave satisfactorily in initial experiments we turn to a different approach.

A {\bf prior over embeddings} is more manageable. We regard an embedding $\vec{X}$ as a vector $\arrowVec{\vec{X}}$ in $\R^{nd}$. We assume the prior  $\arrowVec{\vec{X}} \sim \mathcal{N}\!\left(\vec{0},\vec{\Lambda}\right)$ with $\vec{\Lambda} \in \R^{nd \times nd}$. Furthermore, we assume that the embedded points are independently and identically distributed, that is $\vec{\Lambda}$ is a block diagonal matrix. Overall, the Gaussian prior over embeddings is more restrictive than a prior over distance matrices since a distance matrix can originate from any point configuration. Yet, a prior over embeddings is a reasonable assumption because triplets put strong constraints on the embedding. Hence, the prior mainly provides a scale for the embeddings.

For a {\bf likelihood}, we can use any function $\ell\!\left(\vec{D}\right)$ that appropriately describes how well a triplet is modeled, for example the STE likelihood \eqref{eq:STELikelihood} or other likelihood functions used by other embedding algorithms.

Combining either of the two priors with the chosen likelihood function yields two {\bf posterior} distributions. When using the prior over embeddings we obtain
\begin{equation}
\label{eq:posteriorX}
p\!\left(\vec{X} | \mathcal{S}\right) \propto \mathcal{N}\!\left( \vec{0},\vec{\Lambda}\right) \ell\!\left(\vec{D}\!\left(\vec{X}\right)\right). 
\end{equation}
Both posterior distributions are not analytically tractable. However, it is possible to sample from them using a Markov Chain Monte Carlo algorithm or by applying variational methods. In this paper, we use the MCMC framework of elliptical slice sampling (ESS) by \citet{murray10}. It is applicable for multivariate Gaussian priors and any likelihood function and can therefore be used for both posteriors.  
Elliptical slice sampling requires an initial sample $\vec{X}_0$. We use the fact that some embedding methods maximize a likelihood; for example we use STE to generate  $\vec{X}_0$ when sampling from a posterior that relies on the STE likelihood. As a result, the MCMC random walk begins at a maximum likelihood estimate and discovers the posterior distribution from there. Once we can sample from the posterior, we obtain distance matrices $\vec{D}^{(1)}, \ldots, \vec{D}^{(b)}$, either directly from $p\!\left(\vec{D} | \mathcal{S}\right)$, or via embeddings $\vec{X}^{(1)}, \ldots, \vec{X}^{(b)} $ from $p\!\left(\vec{X} | \mathcal{S}\right)$.

\subsection{Bootstrap approach for generating embeddings} 

In this section we propose a bootstrap 
approach to generate a set of embeddings. Given a training set $\mathcal{S}$ of triplets and a sampling parameter $ r \in \left(0,1\right)$, we uniformly draw $\lfloor r \cdot |\mathcal{S}| \rfloor$ many triplets out of $\mathcal{S}$ without replacement to obtain a subset $\mathcal{S}_1\subset \mathcal{S}$. After that, the given embedding method uses $\mathcal{S}_1$ to embed the $n$ points into $\R^d$. Subsampling and embedding are repeated $b$ times, which results in $b$ embeddings. Note that we draw triplets without replacement for a single subset, but we do not divide $\mathcal{S}$ into $b$ subsets. Therefore, we can choose $b$ arbitrarily high. The resulting embeddings can widely differ in scale and orientation. To make the embeddings comparable, we randomly select one of them as a reference embedding and use a Procrustes analysis to rotate and scale all the other embeddings to match the reference embedding as well as possible \citep{borgGroenen05}. This procedure results in embeddings  $\vec{X}^{(1)}, \ldots, \vec{X}^{(b)} $, which give rise to distance matrices  $\vec{D}^{(1)}, \ldots, \vec{D}^{(b)} $. 
		
\subsection{Computing uncertainty estimates}

Given a set of distance matrices $\vec{D}^{(1)}, \ldots, \vec{D}^{(b)} $ computed with either the Bayesian approach or the bootstrap, we can now compute {\bf uncertainties for all triplet answers}. A naive approach is to compute the uncertainty of a triplet $(i,j,l)$  by evaluating in how many of the $b$ embeddings this triplet is true. However, this majority vote leads to bad results because it does not take the corresponding distances of a triplet into account. We propose the following model-based approach. From the set of $b$ distance matrices we first compute the entry-wise means $\mean{\rho}_{ij}$ and standard deviations $\mean{\sigma}_{ij}$: 
\begin{align*}
 \mean{\rho}_{ij} = \frac{1}{b} \sum\limits_{k = 1}^b \vec{D}^{(k)}_{ij},\hspace{7pt} &\mean{\sigma}_{ij}^2= \frac{1}{b-1} \sum\limits_{k = 1}^b \left( \vec{D}^{(k)}_{ij}- \mean{\rho}_{ij} \right)^2.
\end{align*}
We now assume that the distances $\rho_{ij}$ are distributed as $\mathcal{N}(\mean{\rho}_{ij},\mean{\sigma}_{ij}^2)$ (an assumption that approximately holds in many applications).
With these ingredients, a natural uncertainty estimate can be assigned to any triplet answer: 
the likelihood to observe the triplet answer $(i,j,l)$ is \begin{equation} \label{eq:uncertaintyEstimateWithAvgDistance}
\pi_{ijl} := \Phi \left(\frac{\mean{\rho}_{il} -\mean{\rho}_{ij} }{\mean{\sigma}_{il} + \mean{\sigma}_{ij} } \right),
\end{equation}
where $\Phi$ denotes the cdf of the standard normal distribution. If $\pi_{ijl}$ is close to $0.5$, it signifies high uncertainty about $(i,j,l)$. If $\pi_{ijl}$ is close to $1$ (or $0$), then the triplet $(i,j,l)$ is likely to be true (or the opposite $(i,l,j)$ is true). 

Rather than computing the uncertainty of triplets, we can also evaluate {\bf uncertainties for each point position}. Among embeddings $\vec{X}^{(1)}, \ldots, \vec{X}^{(b)} $ we consider the mean $\mean{\vec{x}}_{i} = \frac{1}{b} \sum\limits_{k = 1}^b \vec{x}^{(k)}_{i}$ and sample covariance of a single point:
\begin{align*} 
\label{eq:pointUncertainty}
\mean{C}_{i} &= \frac{1}{b-1} \sum\limits_{k = 1}^b \left( \vec{x}^{(k)}_{i}- \mean{\vec{x}}_{i} \right) \left( \vec{x}^{(k)}_{i}- \mean{\vec{x}}_{i} \right)^T,
\end{align*}
where $\vec{x}^{(k)}_{i} \in \R^d$ is the embedding of $\xi_i$ in the $k$-th embedding $\vec{X}^{(k)}$. 
Note that these estimates make sense because in both our approaches, the embeddings are well-aligned  and the distance matrices ``comparable'': due to Procrustes analysis in the bootstrap approach, and due to the model-based approach in the Bayesian setting. 
	
\section{Using uncertainty estimates}
\label{sec:UsingEstimates}
In this section, we discuss four applications for our uncertainty estimates. First, we use our triplet uncertainty estimates to predict the true answers of triplet comparisons. Secondly, we estimate an appropriate embedding dimension for ordinal training data. Thirdly, we quantify the uncertainty of a psychophysics experiment by applying the point uncertainties. Finally, we query the most uncertain triplets in an active triplet selection setting.
	
\subsection{Triplet prediction}
\label{sec:TripletPrediction}

Given a set $\mathcal{S}$ of noisy triplets, the aim of triplet prediction is to predict the true answers both for unobserved triplet comparisons and noisy observed triplets. To use our uncertainty estimates for the triplet prediction problem, we introduce the possibility to abstain from a prediction. To this end, we introduce an ``uncertainty threshold'' $t > 0.5$. We predict $\delta_{ij} < \delta_{il}$  if $\pi_{ijl} > t$ and  $\delta_{il} < \delta_{ij}$ if $\pi_{ijl} < 1-t$. If $\pi_{ijl} \in [1-t,t]$ there is no prediction and we ``abstain''.
Note that by construction of the values $\pi_{ijl}$ we have that  $\pi_{ijl}>t \Leftrightarrow \pi_{ilj} < 1-t$ and thus, the above construction cannot lead to any inconsistencies. 

The threshold $t$ can be used for a trade-off between the triplet prediction error and the abstention rate: when the threshold $t$ increases, we abstain more often, but the prediction error on the remaining predicted triplets decreases. It can also be observed that for a fixed threshold $t$, the abstention rate decreases when we increase the number of training triplets because we are more and more certain about our predictions. See Section~\ref{sec:Experiments} for results. 

\subsection{Estimating the embedding dimension}
When we employ an ordinal embedding approach and we are given real world ordinal information, it is often not clear how to choose the embedding dimension $d$ because the true dimension of the data is unknown. The naive approach is to use some measure to quantify the error of the embedding, and choose the dimension which leads to the best result. However, many such measures improve trivially with increasing dimension. This is analogous to overfitting -- we achieve a low error for training triplets, but suffer a high error for unobserved triplets.

Instead, we now use uncertainty estimates to find a good embedding dimension. Given an embedding dimension $d$ we sample many embeddings as described above and compute the average uncertainty over all triplet answers. In regions of overfitting the uncertainty estimates reveal a higher variance due to many degrees of freedom, and in regions of underfitting the estimates reveal uncertainty due to contradicting triplets. As we can see in the experiments in Section~\ref{sec:ExpEmbDim}, the average uncertainty is typically minimized for the correct dimension when using the bootstrap method. In other cases it can give an indication for a good embedding dimension depending on the intrinsic dimension of the data set. Additionally, the uncertainty estimates do not trivially decrease by increasing the embedding dimension.

\subsection{Application in psychophysics}
\label{sec:Psycho}
\begin{figure}[b]
	\centering
	\includegraphics[width=.37\textwidth]{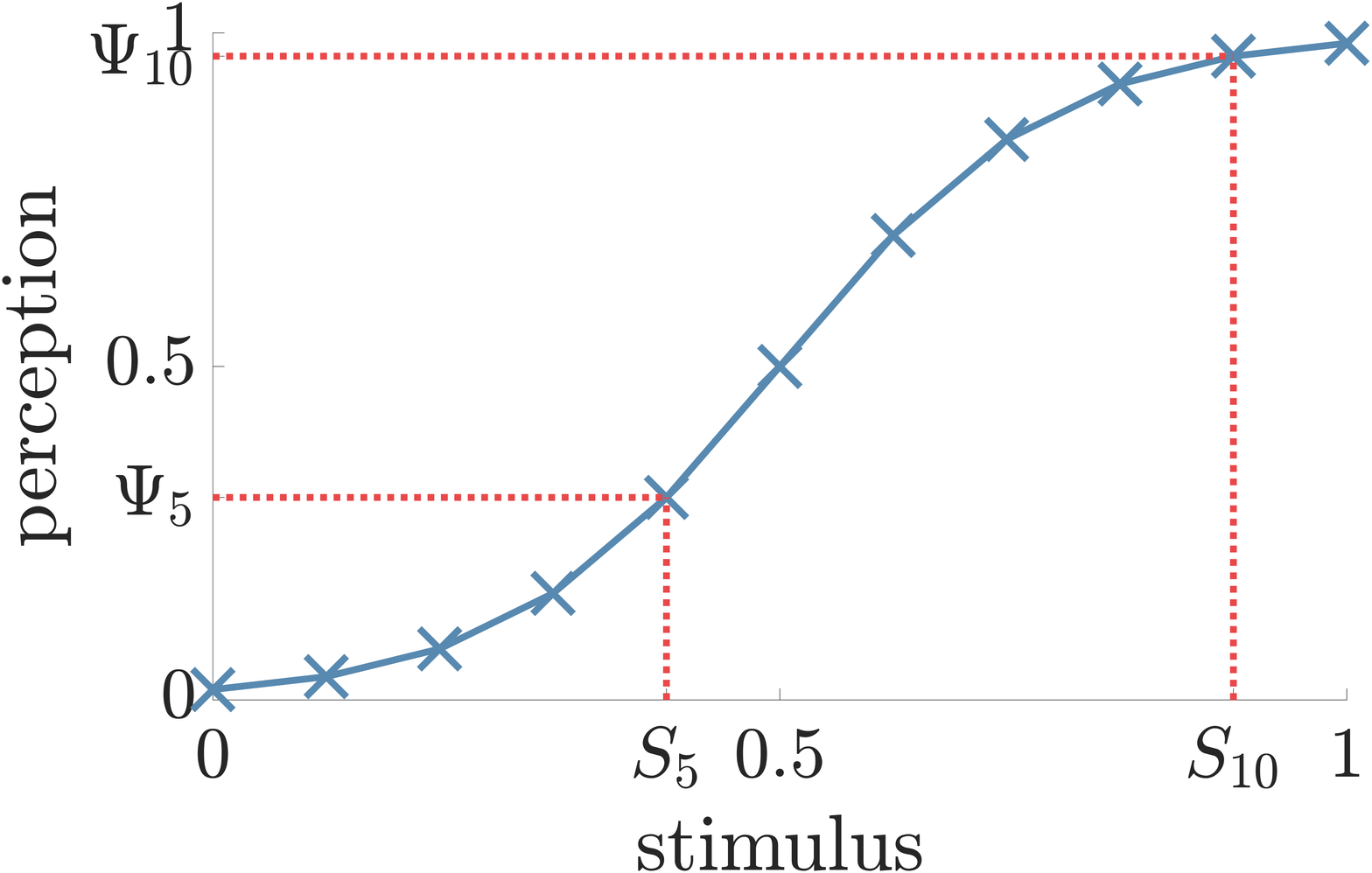}
\caption{Differences of physical stimuli can be perceived differently. The difference between stimulus $S_5$ and $0.5$ is $0.1$ whereas the difference between perception $\Psi_5$ and $0.5$ is $0.2$. On the other hand, stimulus $S_{10}$ is perceived closer to stimulus $1$.}\label{fig:psychoExample}
\end{figure}

When ordinal embedding is used in a context of scientific data analysis, an analysis of uncertainties is often a crucial step. We want to illustrate this with an application in psychophysics. A standard task in this field is to estimate how physical stimuli are perceived by human observers. For example, the input might consist of images of increasing contrast, or of sounds of increasing frequency. The outcome of an experiment is supposed to show how this translates to the perceived increase of contrast, or sounds. See Figure~\ref{fig:psychoExample} for an illustration where we plot a non-linear relation between an exemplary stimulus and its perception. While traditional methods such as the Steven's Magnitude estimation \citep{gescheider97} try to quantitatively measure aspects of this relationship, a different approach is to use ordinal embeddings.
Here, the idea is to ask participants to compare triplets of stimuli, and generate a one-dimensional embedding based on their triplet answers. This embedding corresponds to the projection of the curve in Figure~\ref{fig:psychoExample} on the y-axis. We can now use our uncertainty estimates to additionally evaluate the uncertainties along the estimated perception curve. 

To simulate a psychophysics experiment with a couple of human observers we assume that each observer perceives a stimulus following a fixed perception function (similar to Figure~\ref{fig:psychoExample}). The triplets that the human observer provides are created based on her corresponding perception function. Since every human observer perceives the stimuli differently, each observer has a different perception function and hence, gives different triplet answers. Finally, the set of all observed triplets contains contradictions depending on how different the perception functions of the observers are. In Section~\ref{sec:ExpPsycho} we show that the uncertainties over point positions reveal how much the perception curves differ.

\subsection{Active learning}
\label{sec:ActiveLearning}
For human observers providing triplets can be monotonous and boring after a while \citep{bijmolt95}, and crowd sourcing experiments are time consuming. Therefore, instead of asking all triplet comparisons, it can be advantageous to collect a small amount of the most informative triplets by an active approach. We propose our uncertainty estimates as a means to obtain a practical procedure for active triplet selection. 

In this section we assume that there is a limited budget of triplets that can be collected. Before we query, the crowd provides a small set of random ``seed triplets''. These triplets are answers to comparisons that have been drawn randomly with replacement from the set of all possible comparisons. Our estimates derived in (\ref{eq:uncertaintyEstimateWithAvgDistance}) capture the uncertainty, which was introduced by the given seed set, for all triplets and determine which comparisons we should query next. The intuition is that the seed triplets determine the location of some points, while other points still have a lot of freedom. Therefore, we ask comparisons for which the answers are considered uncertain.

\begin{figure*}[t]
\centering

\subcaptionbox*{}[.3\textwidth]{
	\centering
	\includegraphics[width=.3\textwidth]{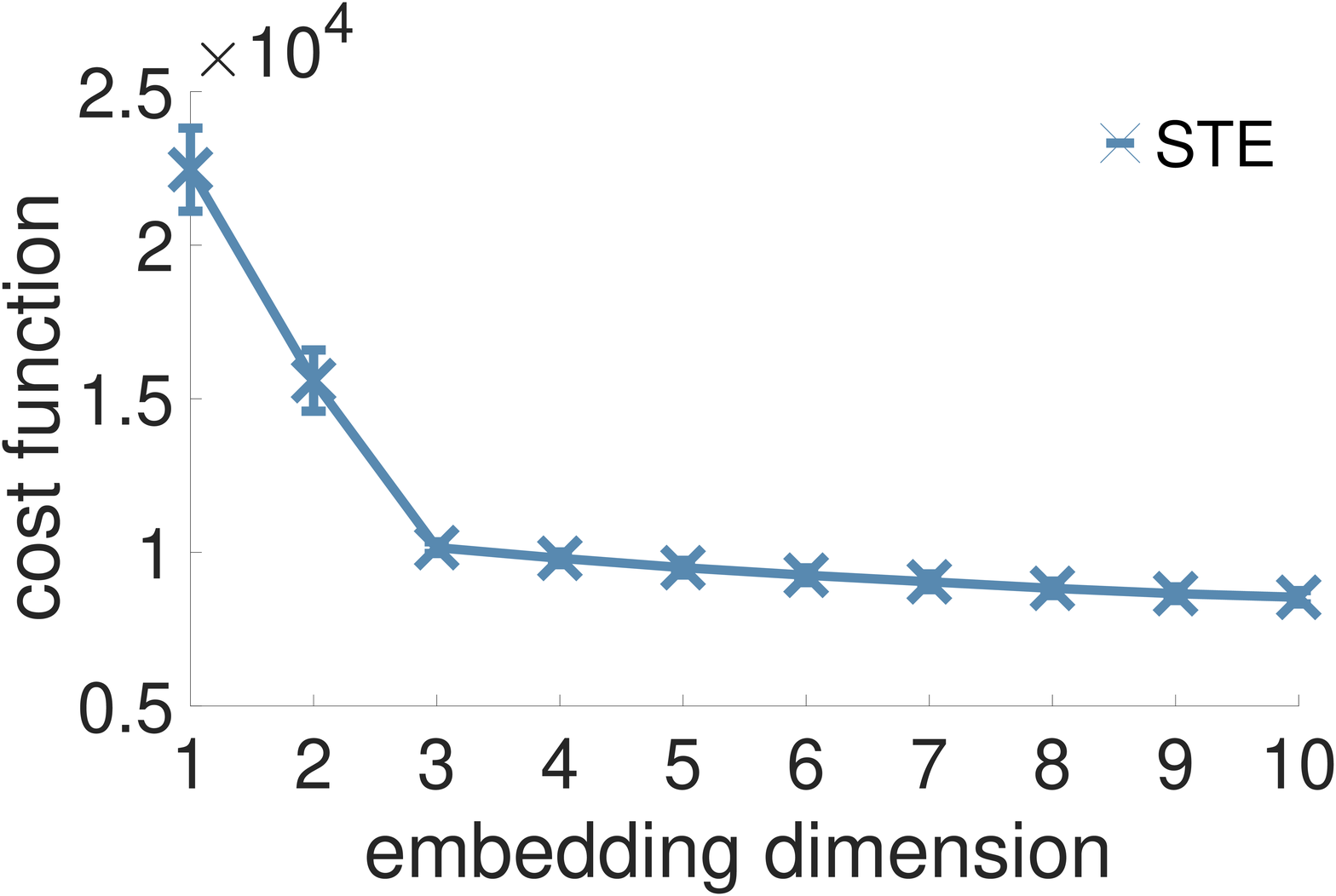}
}\hspace{15pt}
\subcaptionbox*{}[.3\textwidth]{
	\centering
	\includegraphics[width=.3\textwidth]{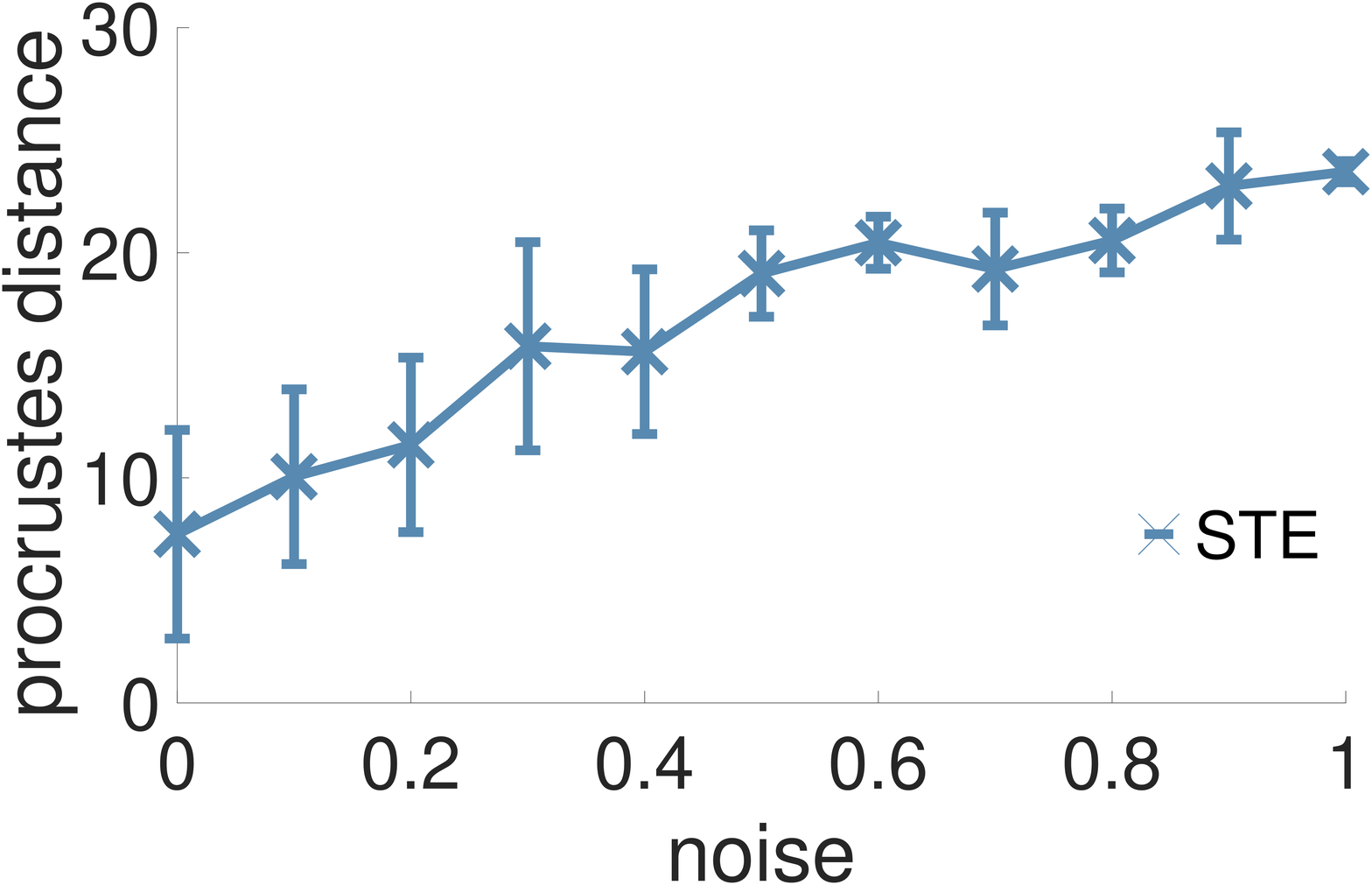}
}\hspace{15pt}
\subcaptionbox*{}[.3\textwidth]{
	\centering
	\includegraphics[width=.3\textwidth]{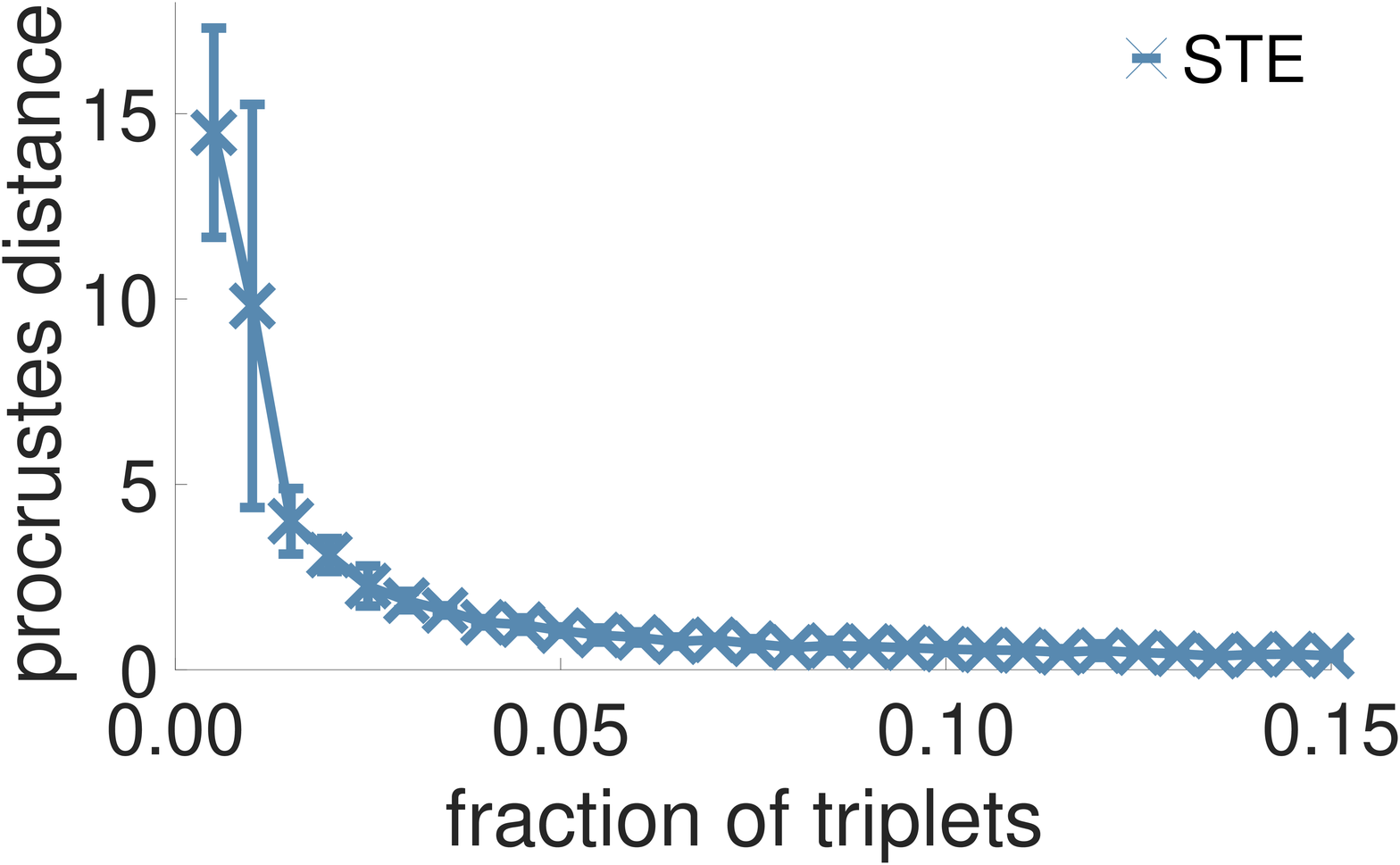}
}\\
\subcaptionbox{Increasing embedding dimension.\label{fig:increasingDimension}}[.3\textwidth]{
	\centering
	\includegraphics[width=.3\textwidth]{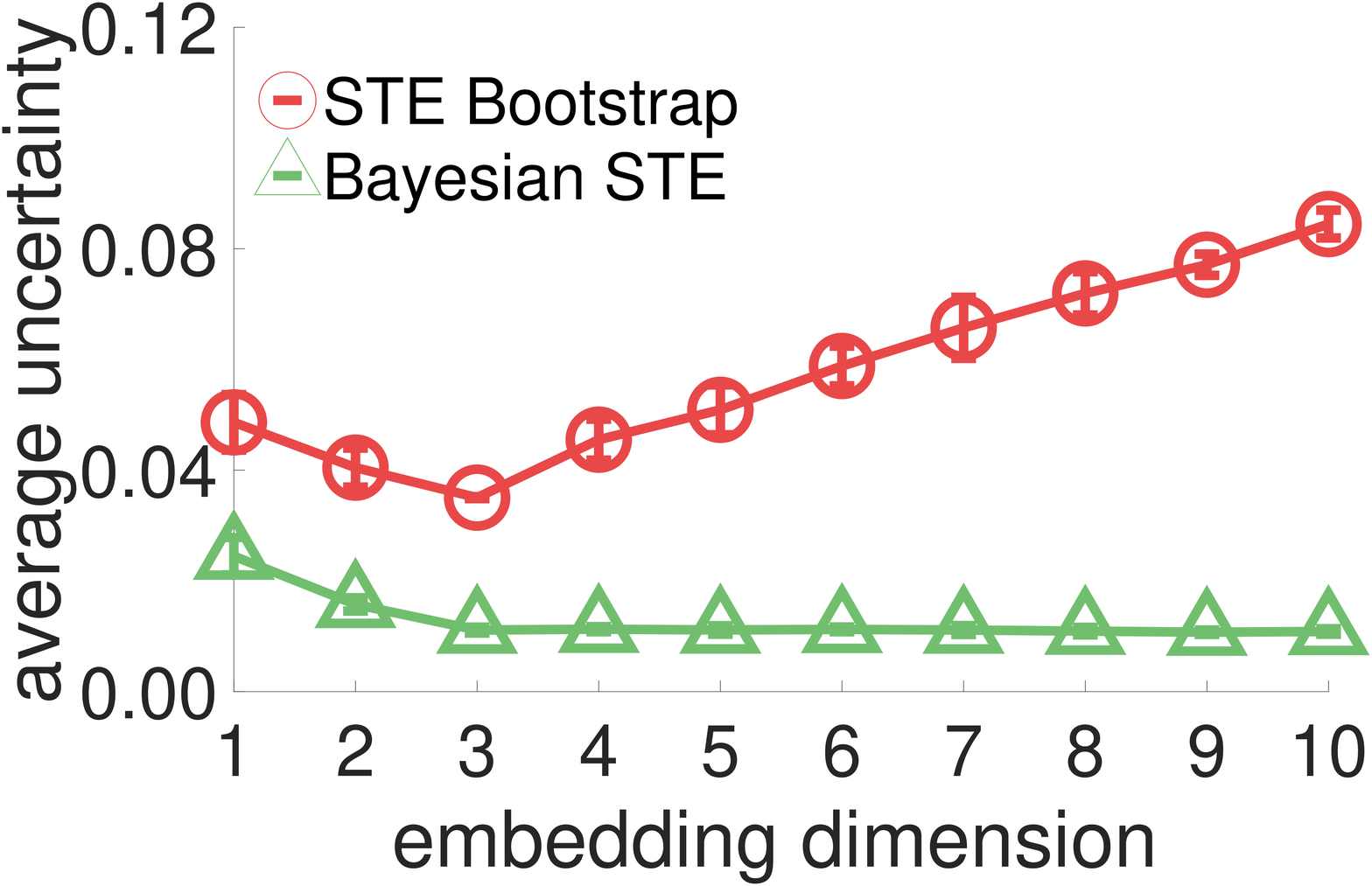}
}\hspace{15pt}
\subcaptionbox{Increasing noise.\label{fig:increasingNoise}}[.3\textwidth]{
	\centering
	\includegraphics[width=.3\textwidth]{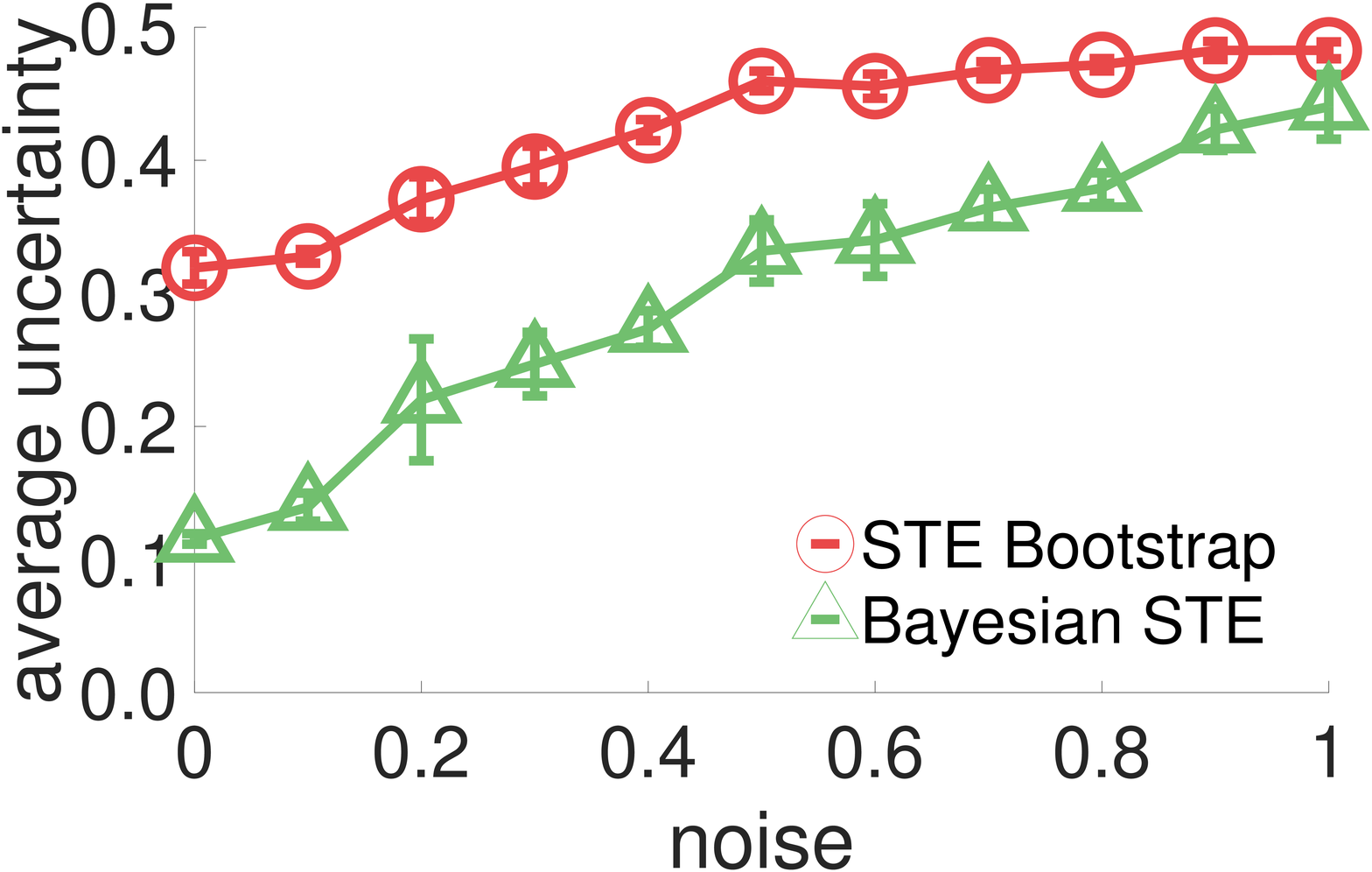}
}\hspace{15pt}
\subcaptionbox{Increasing fraction of triplets.\label{fig:increasingTriplets}}[.3\textwidth]{
	\centering
	\includegraphics[width=.3\textwidth]{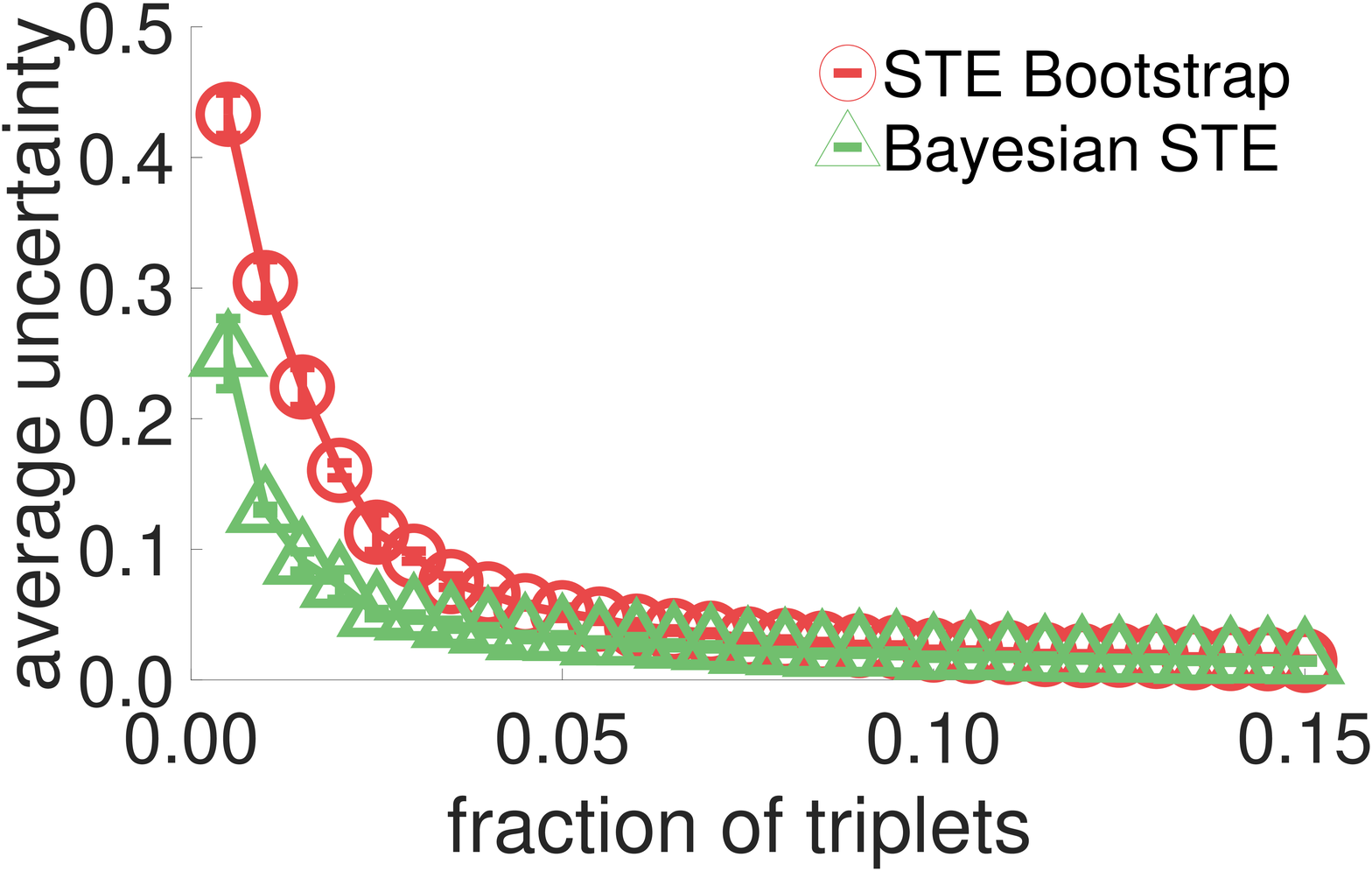}
}
\caption{a) We embed noisy triplets that were created from a projection of MNIST on $d_{\mathrm{true}} = 3$ principal components. The average triplet uncertainty of the boostrap is minimal for $d_{\mathrm{true}}$ whereas the STE cost function decreases further. b) and c) We select $50$ points from a mixture of Gaussians. The top row shows the mean and standard deviation of the embedding errors of STE embeddings measured by the procrustes distance. The second row shows the mean and standard deviation of the average uncertainty. b) We select one percent of all true triplets. Increasing the noise in them induces a higher embedding error. Simultaneously, the uncertainty estimates of both our approaches increase and go to $0.5$ (completely uncertain).  c) Here, we have no noise in the training triplets. The embedding error decreases when more triplets are available. Along with it, the uncertainty decreases.}
\end{figure*}

Given a seed training set $\mathcal{S}$, we augment $\mathcal{S}$ in the following way. First, we detect the triplets with highest uncertainty:
\begin{equation}
\label{eq:MinCertainty}
 \left(i,j,l\right) = 	\argmin\limits_{i,j,l = 1,...,n }\left(\pi_{ijl} - 0.5\right)^2.
\end{equation}
Triplets are binary answers, and our estimates assess the uncertainty about them: if  $(i,j,l)$ is uncertain, then $(i,l,j)$ is uncertain to the same degree, since $\pi_{ilj} = 1 - \pi_{ijl}$. Therefore, two complementary triplets will minimize (\ref{eq:MinCertainty}). Then, we query the corresponding comparison from the crowd in order to observe the triplet answer, and add it to the training set $\mathcal{S}$. When collecting more triplets, we can query one by one, each time updating the uncertainty estimates, or we query bigger batches. Also note that a comparison might be queried several times.

In Section \ref{sec:Experiments}, we perform classification, clustering, and triplet prediction using actively queried triplet comparisons. Unfortunately, the results are disappointing: triplet prediction improves only by a small amount, which does not translate to an improvement for classification and clustering. This finding is in accord with our own experience in many simulations with a large variety of triplet selection mechanisms (landmark settings, geometry-based sampling): often, none of the intricate triplet sampling schemes significantly beat the naive strategy of random selection \citep{jamieson15}. However, it is in fact good news for practitioners. The advice based on the experiments in our paper is to simply query as many random comparisons as possible.

\paragraph{Related work on active triplet selection.}
Some active approaches have already been reported in the literature on ordinal embeddings, however they are not practical for various reasons. \citet{jamieson11} consider an active triplet selection mechanism that is motivated from a theoretical point of view. It relies on a subroutine that outputs a \emph{consistent embedding} if one exists (that is, an embedding that satisfies {\em all} true triplet answers). However, in a setting of noisy triplets, no algorithm can solve this problem. 

\citet{tamuz11} propose an adaptive selection algorithm, which is similar to our Bayesian perspective. In order to collect a triplet for any point $a$, they consider the posterior distribution over the location of $a$ in  $\R^d$, involving all those triplets that have $a$ as the anchor. Then a pair of points $b, c \neq a$ is selected by a procedure that maximizes an \emph{information gain criterion}, and the corresponding comparison is queried. Unfortunately, no code for this method is made available, but a data-driven approximation of the information gain can be used from \cite{jamieson15}. However, it was not possible to produce competitive results within our framework. For the sake of completeness see the supplementary material for results comparing to the information gain criterion.

\begin{figure*}
	\centering
	\subcaptionbox{Triplet prediction error.\label{fig:predictionError}}{
		\centering
		\includegraphics[width=0.45\textwidth]{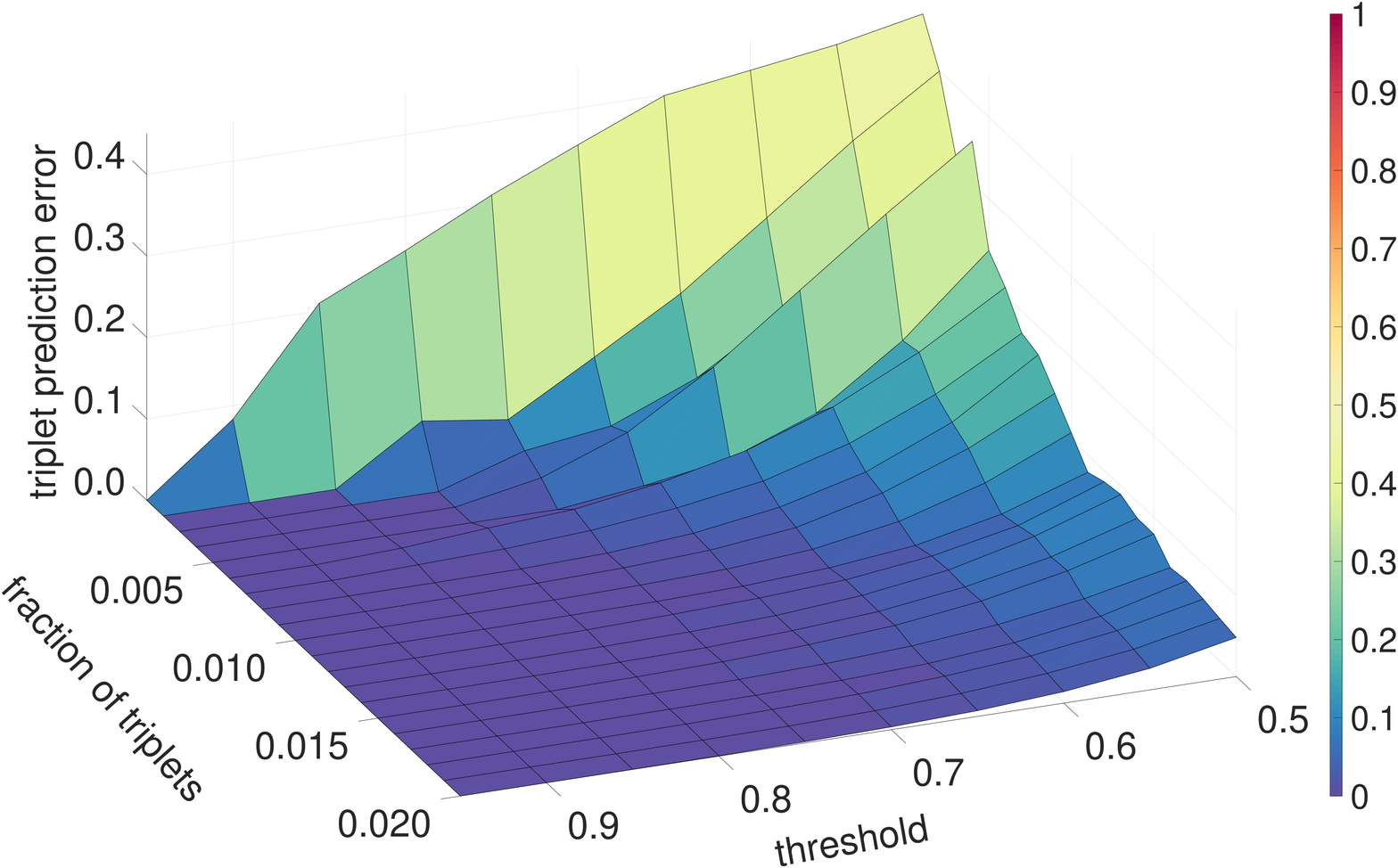}
	}
	\subcaptionbox{Abstention rate.\label{fig:abstentionRate}}{
		\centering
		\includegraphics[width=0.45\textwidth]{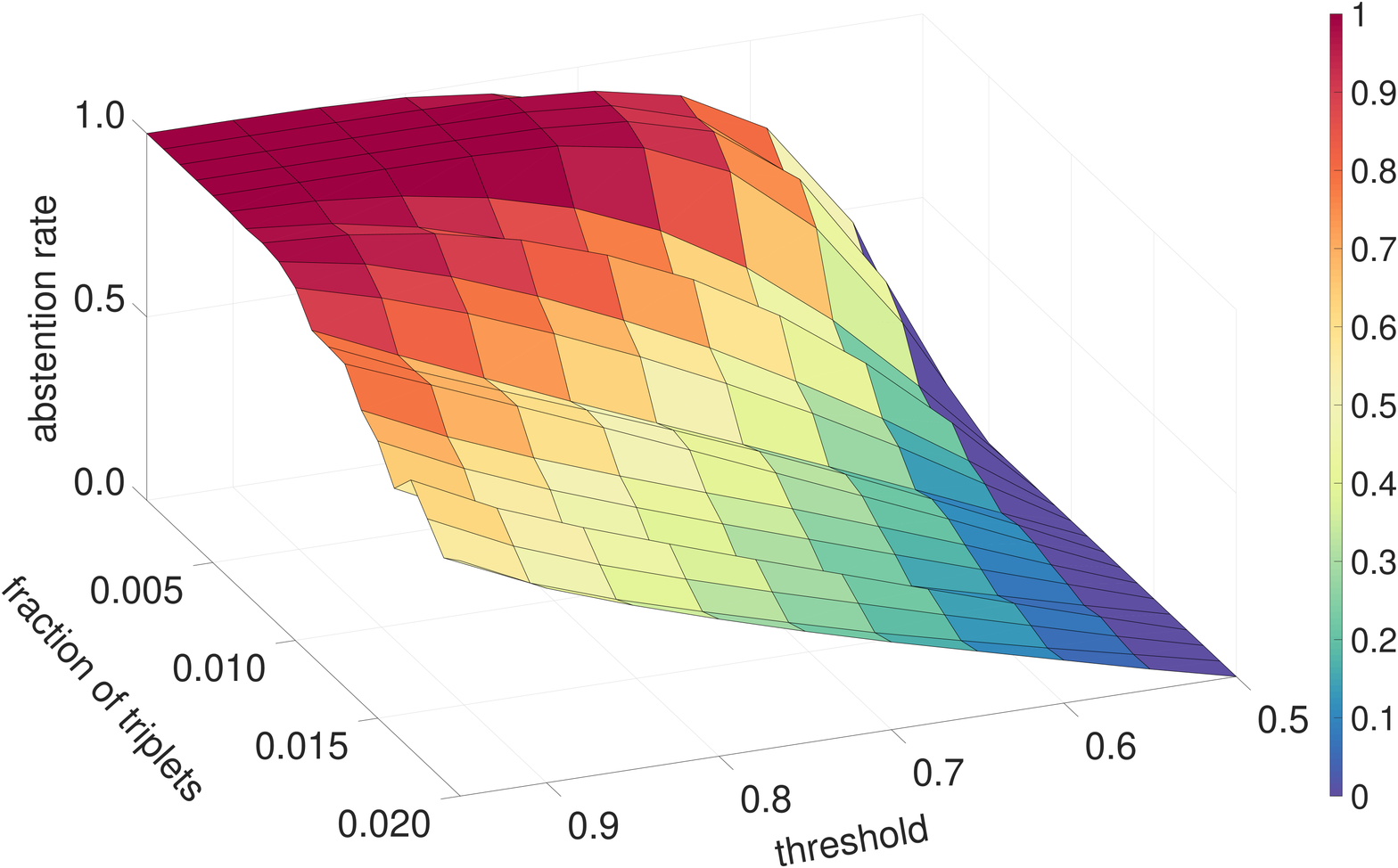}
	}
	\caption{Triplet prediction with STE Bootstrap performed on $50$ points from a two-dimensional mixture of three Gaussians. A triplet set of variable size is used to generate uncertainty estimates for all triplets. If the estimate is above a threshold $t$ (or below $1-t$), we make a prediction, otherwise we abstain. a) The triplet prediction error decreases when a higher threshold for certainty is required, or when the number of triplets increases. b) The abstention rate increases, when the threshold increases, but decreases when more triplets are used.} \label{fig:tripletPrediction}
\end{figure*}

\section{Experiments}
\label{sec:Experiments}
In this section, we first examine the behavior of our uncertainty estimates when the number of training triplets or the noise level changes. Secondly, we consider the task of triplet prediction. Thirdly, we use our estimates to find an appropriate embedding dimension. Additionally, we simulate a psychophysics experiment and estimate point uncertainties. Lastly, we employ our estimates for an active learning approach and examine the performance with respect to triplet prediction, classification and clustering. 

\paragraph{Method setup.}
With \emph{STE Bootstrap} we denote our bootstrap approach based on the STE embedding method. We use the code by \citet{maaten12} for the embedding algorithms. With random subsamples of $40$ percent of the training triplets (without replacement), we generate $b=20$ bootstrap embeddings. 

By \emph{Bayesian STE} we denote our Bayesian approach (\ref{eq:posteriorX}) using the STE likelihood (\ref{eq:STELikelihood}). Here we use the MCMC method as described above to sample $500$ embeddings. Since we start with a maximum likelihood solution as described above, no burn-in phase is necessary. For the prior covariance matrix $\vec{\Lambda}$ in (\ref{eq:posteriorX}), we simply use $15\cdot \vec{\mathrm{I}}$.

All experiments except for triplet prediction and the application in psychophysics have been repeated $5$ times, using the same set of Euclidean points, but independent training triplets. The figures always display means and standard deviations. Given a fixed number of points from the data sets in question, we use the Euclidean distance for the underlying distance $\delta$. The training set $\mathcal{S}$ of noisy triplets is generated as follows: we model two random variables $z_{ij}$ via $\mylog{z_{ij}} \sim \mathcal{N}\!\left( \mylog{\delta_{ij}}, \sigma^2\right)$ and $z_{il}$ via $\mylog{z_{il}} \sim \mathcal{N}\!\left( \mylog{\delta_{il}}, \sigma^2\right)$. The two positive random variables  $z_{ij}$ and $z_{il}$ take the role of distances between points; if $ z_{ij} < z_{il}$, then the triplet $(i,j,l)$ is added to $\mathcal{S}$, or vice versa. This ensures that the $z_{ij}$ are positive, and two similar distances can cause a wrong triplet independent of their magnitude. 

The number of triplets that we use in these experiments is in the range of $0.1$ up to $10$ times $dn\mylog{n}$. If the number of triplets grows much further, there is no uncertainty for the estimates to reveal except for high noise levels. Besides, the goal of active learning is to query as few triplets as possible. Moreover, computing uncertainty estimates requires repeated embedding or sampling, and hence, running time profits from a low number of triplets. 

\subsection{Are the uncertainty estimates well calibrated?}
We first need to verify that our estimates are well calibrated, in the sense that they react ``correctly'' under the influence of noise and the amount of training triplets. To this end, we compare the changes of both the embedding error and the average uncertainty over triplets when increasing the noise or the amount of training triplets. 

In the following, we use a two-dimensional mixture of three Gaussians. The means of the three components are $\mu_1 = \left[2,2\right] , \mu_2 = \left[-2,-1\right], \mu_3 = \left[4,-2\right]$ and the covariance matrices are $\Sigma_1 = [2,0; 0,1], \Sigma_2 = [1,0; 0,1]$ and  $\Sigma_3 = [1, 0.7; 0.7,2] $. We randomly draw $50$ points, generate triplets as described above, and embed again in $\R^2$.  We compute the error of an embedding $\vec{X} \in \R^2$ in relation to the ground truth $\vec{X}^\ast\in \R^2$ by performing a Procrustes analysis. We then report the \emph{procrustes distance} $\min\limits_{\vec{U}}\mynorm{\vec{X}\vec{U} - \vec{X}^\ast}^2_F$, which uses the Frobenius norm and minimizes over orthogonal $\vec{U}$. For the overall uncertainty, we compute the average over the uncertainties of all true triplets. The values displayed in the figures are the averages of the overall uncertainties over the $5$ independent runs of the experiments.

\paragraph{Increasing noise.}
Given a fixed fraction of true triplets, we generate noisy training sets by the mentioned noise model for a range of values for $\sigma$. We expect that increasing the noise level leads to an increase in the embedding errors. Simultaneously, we expect that the uncertainty increases, as the data contains more contradicting triplets. In Figure \ref{fig:increasingNoise} we can see both effects for the example of the STE embedding algorithm: both the Bayesian and the bootstrap uncertainty increases with increasing noise and converge to $0.5$ (completely uncertain). Both our uncertainty measures qualitatively behave similarly, but the bootstrap approach seems to be more conservative in the sense that it produces higher values of uncertainty. Experiments with other embedding algorithms yield similar results, see the supplementary for details.

\paragraph{Increasing the number of triplets.}
In this case, we have $\sigma = 0$, but we increase the number of all true triplets in $\mathcal{S}$ from $0.5 \%$ to $15 \%$. In Figure \ref{fig:increasingTriplets}, we can see that the procrustes distance decreases as rapidly as the average uncertainty when the size of the noise-free training set $\mathcal{S}$ increases. When the number of triplets is roughly $10 dn\mylog{n}$, the uncertainty estimates identify that sufficiently many triplets are used.

\subsection{Triplet prediction.}
On the same mixture of Gaussians, we perform the task of triplet prediction as described in Section \ref{sec:UsingEstimates}. We consider noise-free training sets $\mathcal{S}$ with varying size, and we consider different choices for the abstention threshold $t$. For all combinations of parameters we then report the trade-off between abstention rate and triplet prediction error (the proportion of wrongly predicted triplets). 

First, we expect that the prediction error decreases with an increasing threshold $t$, because we predict triplets with low uncertainty. Surely, the abstention rate increases with increasing threshold $t$ and if we abstained on every prediction, the triplet prediction error would be zero. Therefore, we expect, secondly, the abstention rate to decrease, when the number of training triplets increases. In Figure \ref{fig:tripletPrediction} we show that both effects take place. 

\subsection{Estimating the embedding dimension}
\label{sec:ExpEmbDim}

To test whether our uncertainty estimates are a good tool to estimate the embedding dimension, we first generate a data set with known ground truth. To this end, we use the digits $6$ and $8$ of MNIST \citep{lecun98} and project the data set on its first principal components using $d_{\mathrm{true}} = 3$ components. Then we sample $50$ points from this low-dimensional data set, evaluate $20$\% of all corresponding triplets, and add noise with the described noise model ($\sigma = 0.1$). Then we use ordinal embedding to embed into various dimensions, and we report the cost function of STE and evaluate the average triplet uncertainty with our bootstrap and Bayesian approach.

In Figure~\ref{fig:increasingDimension} we can see the results for the case of $d_{\mathrm{true}} = 3$ (other results look similar, see the supplement for different $d_{\mathrm{true}}$). The cost function of the STE algorithm decreases with an increasing embedding dimension, but in case of the Bayesian approach, the average uncertainty drops and stays constant at $d_{\mathrm{true}}$. Furthermore, we find that the bootstrap method picks up the true dimension because the average triplet uncertainty is low or minimal for the correct dimension. 

\begin{figure}[t]
\centering

\subcaptionbox{ }[.35\textwidth]{
	\centering
	\includegraphics[width=.35\textwidth]{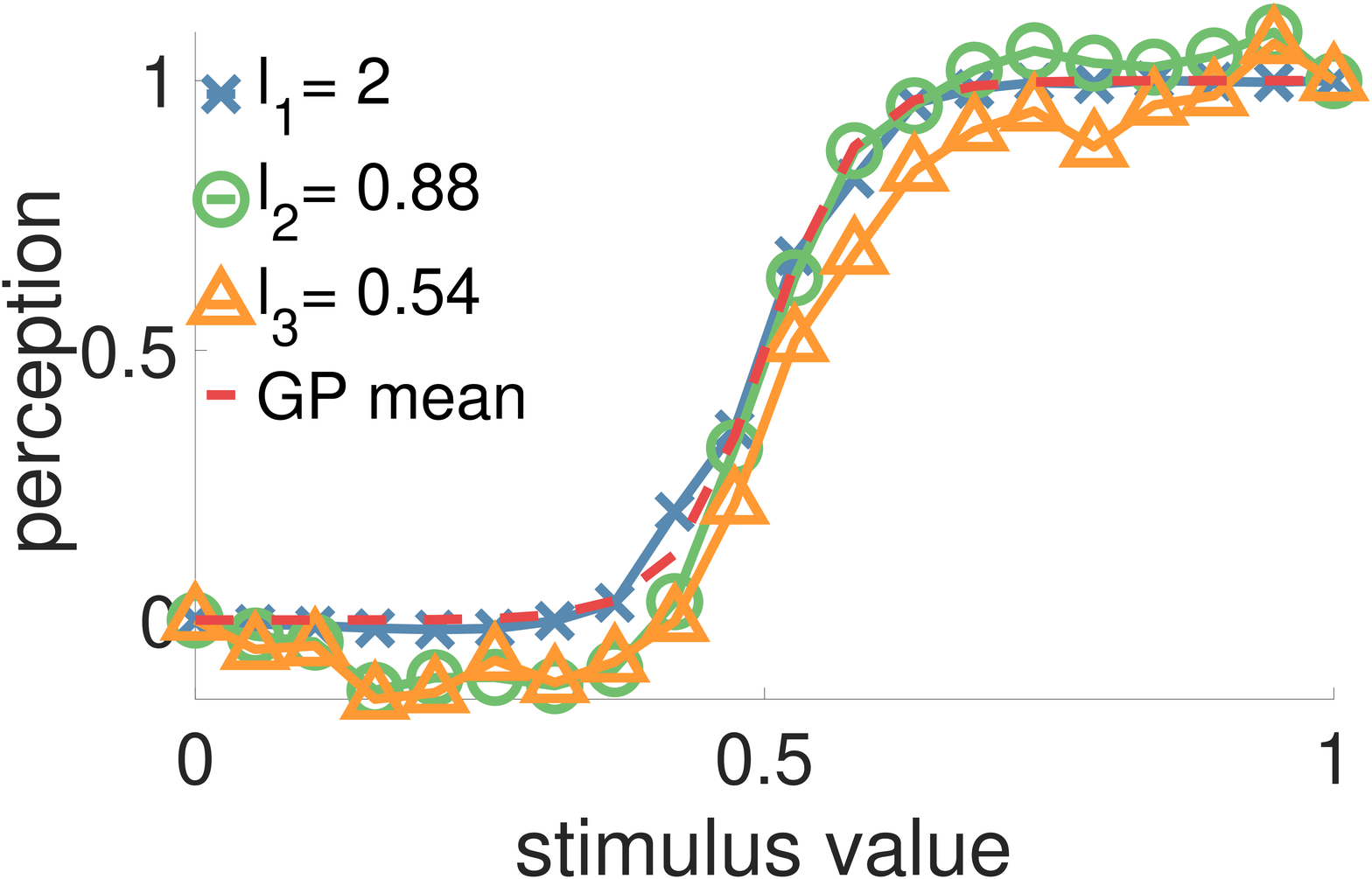}
}\\
\subcaptionbox{}[.35\textwidth]{
	\centering
	\includegraphics[width=.35\textwidth]{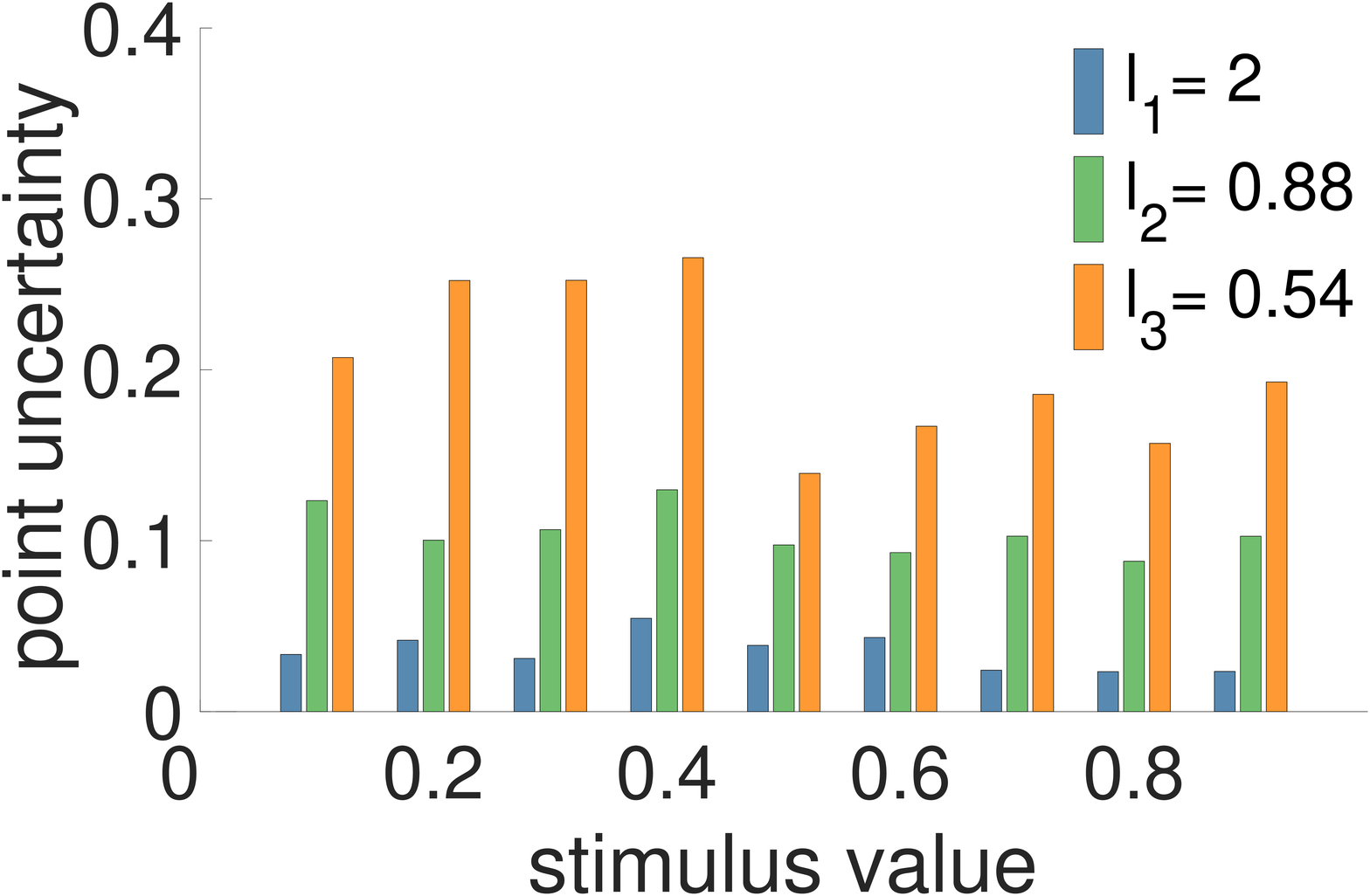}
}
\caption{(a) We compare the mean embeddings resulting from the boostrap for three lengthscales. The embedding is worse for more contradicting triplets, that is when $l$ is low. (b) For nine stimuli we compare the resulting point uncertainties. With decreasing $l$, they increase since the observers have more different perception functions.}\label{fig:psycho}
\end{figure}

\label{sec:ExpPsycho}
\begin{figure*}[t]
\subcaptionbox{Triplet Prediction.}[.32\textwidth]{
	\centering
	\includegraphics[width=.32\textwidth]{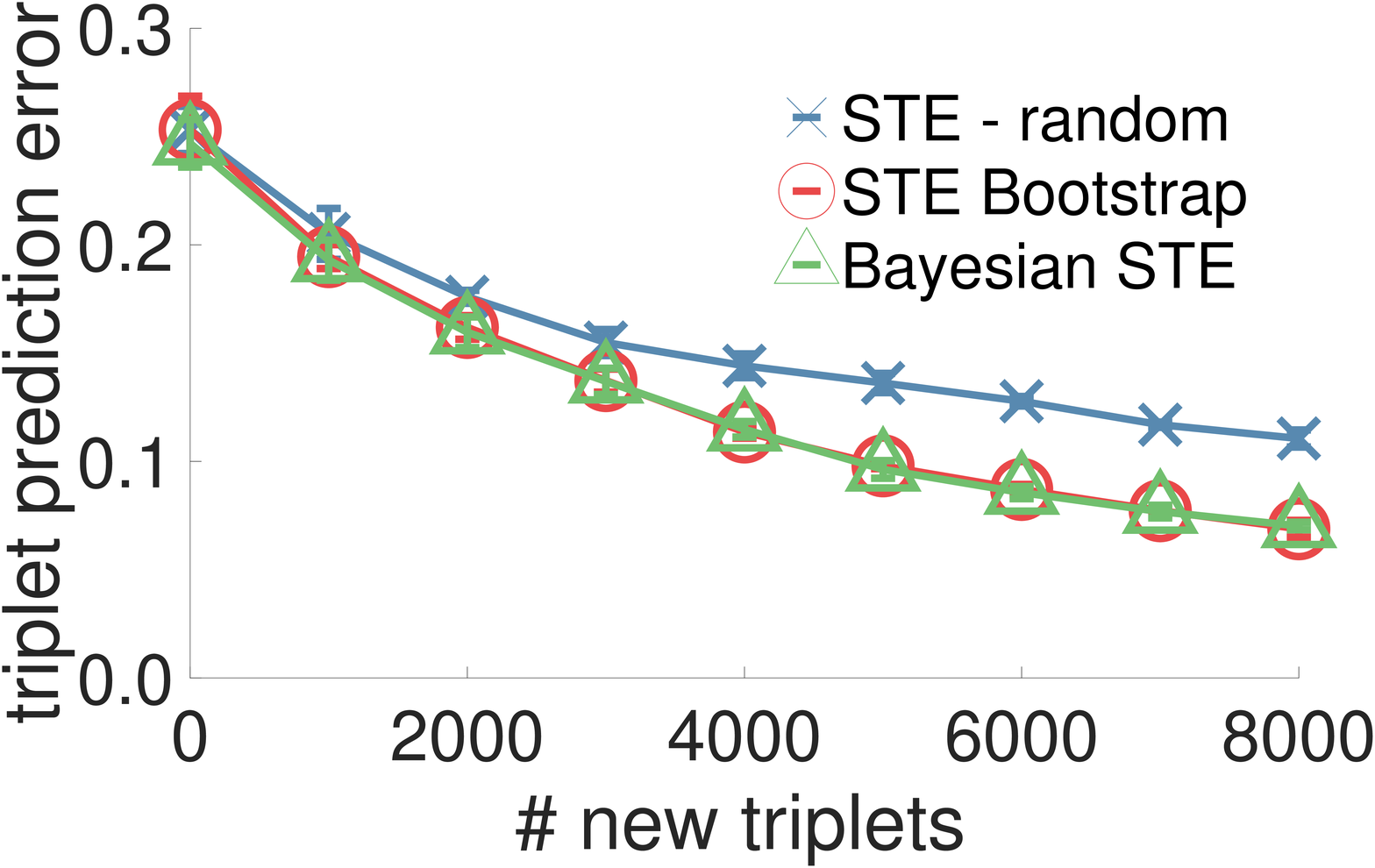}
}
\subcaptionbox{Classification.}[.32\textwidth]{
	\centering
	\includegraphics[width=.32\textwidth]{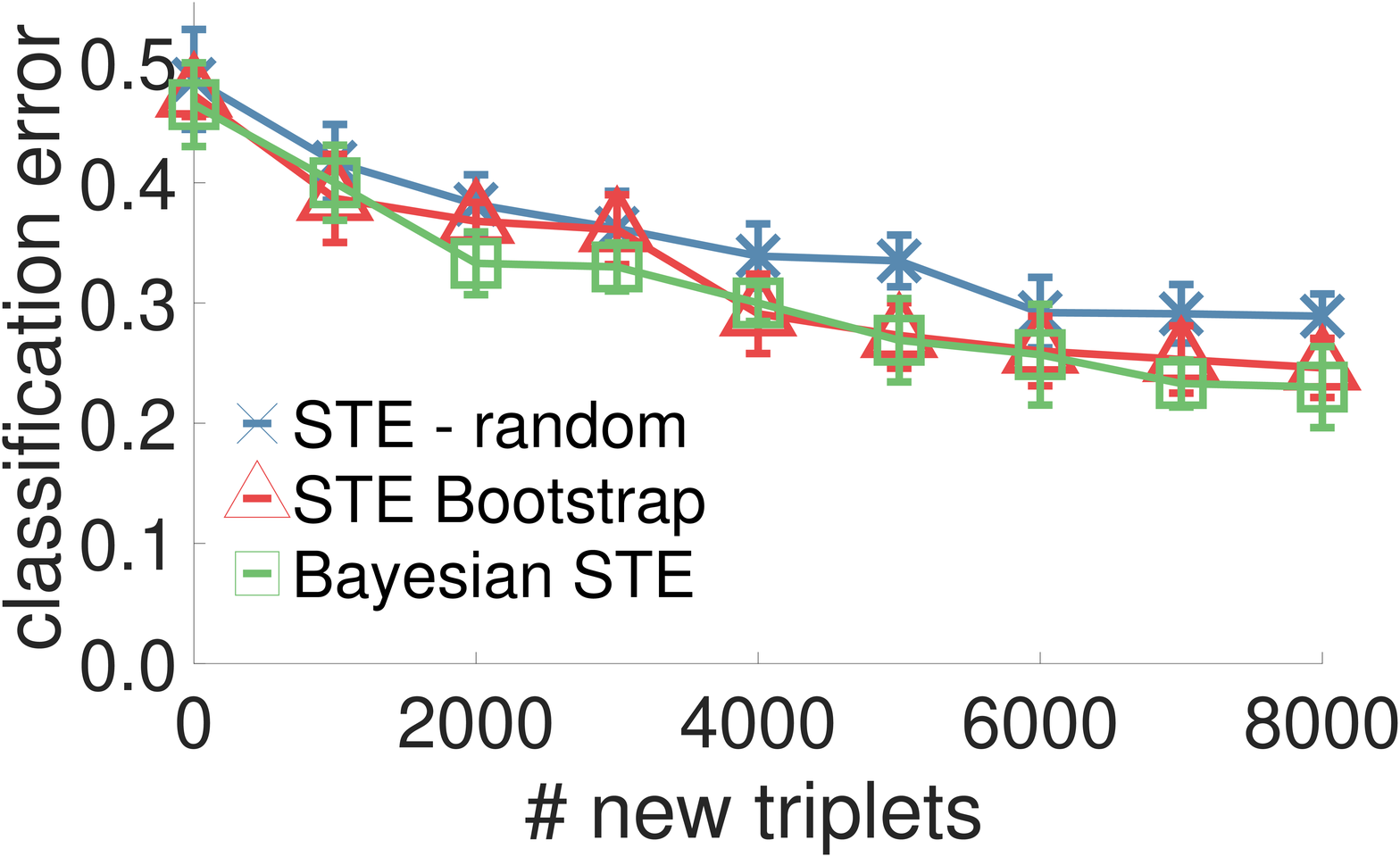}
}
\subcaptionbox{Clustering.}[.32\textwidth]{
	\centering
	\includegraphics[width=.32\textwidth]{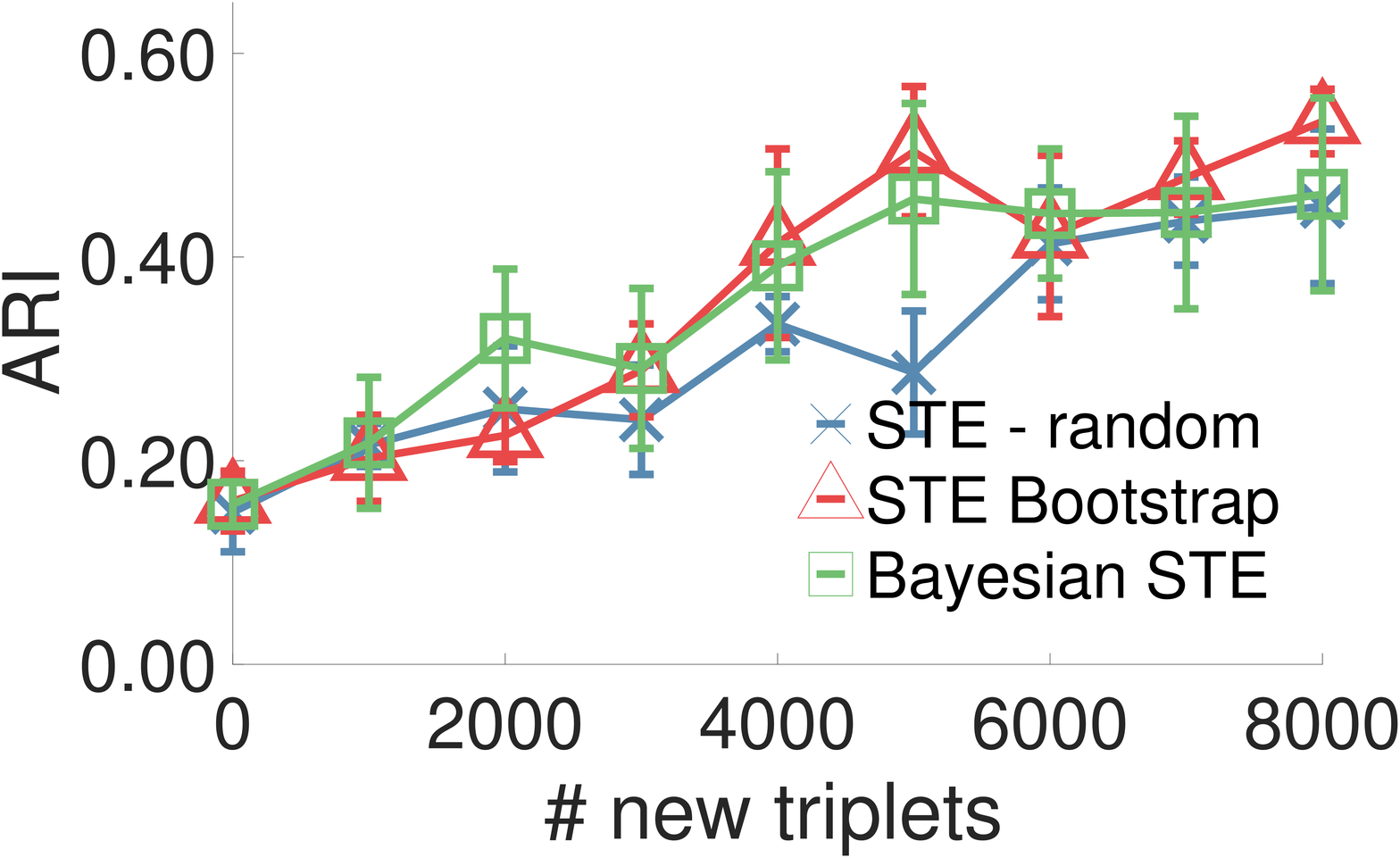}
}
\caption{We compare random selection of triplets with our active approaches on the satellite data set using the STE algorithm. The original dimension is $36$, the embedding dimension is $5$. This figure has been created with $n = 200$ points and noise level $\sigma = 0.1$. Note that the ARI is good when it is high. 
} \label{fig:Active}
\end{figure*}

\subsection{Application in psychophysics}

In order to simulate a psychophysics experiment as described in Section~\ref{sec:Psycho} we discretize the stimulus in $20$ steps in $[0,1]$ and represent every human observer with her individual perception function.  The triplet answers of each observer are created based on their perception function. The overall training data is the collection of all triplet answers and is used to construct the one-dimensional ordinal embedding, which estimates the relative positions of the perceived stimuli on the y-axis. Due to different perception functions the training data contains inconsistent triplet answers and hence, there is uncertainty about these locations. We use our Bootstrap approach and the STE embedding method to estimate the point uncertainties. 

We control how different each observers perception function is by drawing them from a Gaussian Process $\mathcal{GP}(m,k)$ \citep{rasmussen06} and changing the covariance function $k$.  The mean function for the Gaussian Process $\mathcal{GP}(m,k)$ is given by a logistic function $m(x) = (1+\myexp{-25(x-0.5)})^{-1}$. For the covariance function $k$ we choose the squared exponential kernel $k(x_i,x_j) = \myexp{(-2l^2)^{-1} \mynorm{x_i-x_j}^2}$ where $l>0$ denotes the lengthscale, which we use to control the covariances. Additionally, we fix all functions to be the identity at stimulus $0$ and at stimulus $1$. See the supplementary material for the exact posterior Gaussian Process and an illustration for a fixed lengthscale.

For $50$ observers we draw perception functions on $n = 20$ stimuli and create $100$ triplet answers from each function. On a total of $5000$ triplets we perform a bootstrap (with $b = 50$ and $r = 0.1$) and obtain several embeddings. Since ordinal embeddings are invariant to scale and rotation, we normalize each embedding such that the stimulus $0$ is perceived as $0$ and the stimulus $1$ is perceived as $1$. Lastly, we compute the mean position $\mean{x}_i$ and standard deviation $\mean{s}_i := \sqrt{\mean{C}_i}$ for each embedded stimulus $i = 1,...,20$. We repeat this procedure for three different lengthscales $l_1 = 2,l_2 = 0.88$ and $l_3 = 0.54$. When we decrease the lengthscale $l$, the perception functions differ more from the mean and we obtain more contradicting triplet answers. Correspondingly, the uncertainty of each point position increases (see Figure~\ref{fig:psycho}).

\subsection{Active learning}

We now evaluate our active learning approach using three different learning tasks on different data sets. The tasks are (i) triplet prediction, evaluated by the triplet prediction error, which in this context is the ratio of true triplets that are not satisfied in an embedding; (ii) classification on the embedded data points using a simple kNN classifier, evaluated by classification error; (iii) clustering using the normalized spectral clustering algorithm, evaluated with the Adjusted Rand Index (ARI) against the clusters provided by the true labels.

The active learning approach is realized as follows. As described in Section \ref{sec:ActiveLearning}, we start with a random seed of $2,000$ triplets and the noise level
$\sigma = 0.1$. We then use our uncertainty estimates to identify the $1,000$ most uncertain triplet comparisons, query them, and add the noisy triplet answers to the training set. We then evaluate the resulting embedding corresponding to the respective task. Subsequently, we compute new uncertainty estimates and, based on them, query more triplet comparisons. We repeat this procedure of adding batches of $1,000$ uncertain triplets until $10,000$ triplets are in the training set. For the passive experiments, $1,000$ random triplets are added each step. For both the active and passive approach, we embed the current triplet set with the respective embedding method into $d = 5$ dimensions. Note that in these experiments we compute the uncertainty estimates for all possible triplets. Therefore, the framework allows for repeatedly querying the same triplet comparison. 

 In Figure \ref{fig:Active} we examine the results of our two active strategies with the STE embedding method on the Landsat Satellite data set \citep{UCI17}. Results with $t$-STE and other data sets like MNIST \citep{lecun98}, and Breast Cancer \citep{wolberg90} are reported in the supplementary. In terms of triplet prediction error the active approaches improve the embeddings by a noticeable margin, although not overwhelmingly. For the other tasks the active methods do not show an effect on classification error and ARI.

\section{Discussion}
We presented both a bootstrap and a Bayesian approach that compute uncertainty estimates for a given embedding model when few and noisy triplets are available. They are well calibrated, and they can be used for the triplet prediction problem. In a negative result, we found that the uncertainty estimates are not overly helpful for active learning. This agrees with our experience that random triplet selection often outperforms elaborate ideas. In an application in psychophysics the estimates quantified the uncertainty about the outcome of the experiment. Additionally, the uncertainty estimates are helpful for embedding parameters: when using the bootstrap they can give an indication about an appropriate embedding dimension.

\newpage

\subsubsection*{Acknowledgments}

This work has been supported by the International Max Planck Research School for Intelligent Systems (IMPRS-IS), the Institutional Strategy of the University of Tübingen (DFG, ZUK 63) and the DFG Cluster of Excellence “Machine Learning – New Perspectives for Science”, EXC 2064/1, project number 390727645. Philipp Hennig gratefully acknowledges financial support by the European Research Council through ERC StG Action 757275-PANAMA.

\bibliographystyle{plainnat}

\begin{thebibliography}{28}
	\providecommand{\natexlab}[1]{#1}
	\providecommand{\url}[1]{\texttt{#1}}
	\expandafter\ifx\csname urlstyle\endcsname\relax
	\providecommand{\doi}[1]{doi: #1}\else
	\providecommand{\doi}{doi: \begingroup \urlstyle{rm}\Url}\fi
	
	\bibitem[Agarwal et~al.(2007)Agarwal, Wills, Cayton, Lanckriet, Kriegman, and
	Belongie]{agarwal07}
	S.~Agarwal, J.~Wills, L.~Cayton, G.~Lanckriet, D.~Kriegman, and S.~Belongie.
	\newblock Generalized non-metric multidimensional scaling.
	\newblock In \emph{Artificial Intelligence and Statistics}, 2007.
	
	\bibitem[Amid and Ukkonen(2015)]{amid15}
	E.~Amid and A.~Ukkonen.
	\newblock Multiview triplet embedding: Learning attributes in multiple maps.
	\newblock In \emph{International Conference on Machine Learning}, 2015.
	
	\bibitem[{Anderton} et~al.(2018){Anderton}, {Pavlu}, and {Aslam}]{anderton18}
	J.~{Anderton}, V.~{Pavlu}, and J.~{Aslam}.
	\newblock {Revealing the Basis: Ordinal Embedding Through Geometry}.
	\newblock \emph{arXiv e-prints}, art. arXiv:1805.07589, 2018.
	
	\bibitem[Arias-Castro(2017)]{arias-castro17}
	E.~Arias-Castro.
	\newblock Some theory for ordinal embedding.
	\newblock \emph{Bernoulli}, 2017.
	
	\bibitem[Bartels and Hennig(2016)]{bartels16}
	S.~Bartels and P.~Hennig.
	\newblock Probabilistic approximate least-squares.
	\newblock In \emph{Artificial Intelligence and Statistics}, 2016.
	
	\bibitem[Bijmolt and Wedel(1995)]{bijmolt95}
	T.~H.A. Bijmolt and M.~Wedel.
	\newblock The effects of alternative methods of collecting similarity data for
	multidimensional scaling.
	\newblock \emph{International Journal of Research in Marketing}, 1995.
	
	\bibitem[Borg and Groenen(2005)]{borgGroenen05}
	I.~Borg and P.J.F. Groenen.
	\newblock \emph{{Modern Multidimensional Scaling: Theory and Applications}}.
	\newblock Springer, 2005.
	
	\bibitem[Dattorro(2005)]{dattorro05}
	J.~Dattorro.
	\newblock \emph{Convex Optimization \& Euclidean Distance Geometry}.
	\newblock 2005.
	
	\bibitem[Dheeru and Karra~Taniskidou(2017)]{UCI17}
	D.~Dheeru and E.~Karra~Taniskidou.
	\newblock {UCI} machine learning repository, 2017.
	
	\bibitem[Gescheider(1997)]{gescheider97}
	G.~A. Gescheider.
	\newblock \emph{Psychophysics: The Fundamentals}.
	\newblock Lawrence Erlbaum Associates, 1997.
	
	\bibitem[Heikinheimo and Ukkonen(2013)]{heikinheimo13}
	H.~Heikinheimo and A.~Ukkonen.
	\newblock The crowd-median algorithm.
	\newblock In \emph{HCOMP}, 2013.
	
	\bibitem[Jain et~al.(2016)Jain, Jamieson, and Nowak]{jain16}
	L.~Jain, K.~G. Jamieson, and R.~D. Nowak.
	\newblock Finite sample prediction and recovery bounds for ordinal embedding.
	\newblock In \emph{Advances in Neural Information Processing Systems}. 2016.
	
	\bibitem[Jamieson and Nowak(2011)]{jamieson11}
	K.~G. Jamieson and R.~D. Nowak.
	\newblock Low-dimensional embedding using adaptively selected ordinal data.
	\newblock In \emph{49th Annual Allerton Conference on Communication, Control,
		and Computing}, 2011.
	
	\bibitem[Jamieson et~al.(2015)Jamieson, Jain, Fernandez, Glattard, and
	Nowak]{jamieson15}
	K.~G. Jamieson, L.~Jain, C.~Fernandez, N.~J. Glattard, and R.~D. Nowak.
	\newblock Next: A system for real-world development, evaluation, and
	application of active learning.
	\newblock In \emph{Advances in Neural Information Processing Systems}. 2015.
	
	\bibitem[{Kanagawa} et~al.(2018){Kanagawa}, {Hennig}, {Sejdinovic}, and
	{Sriperumbudur}]{2018arXiv180702582K}
	M.~{Kanagawa}, P.~{Hennig}, D.~{Sejdinovic}, and B.K. {Sriperumbudur}.
	\newblock {Gaussian Processes and Kernel Methods: A Review on Connections and
		Equivalences}.
	\newblock \emph{arXiv e-prints}, art. arXiv:1807.02582, 2018.
	
	\bibitem[{Karaletsos} et~al.(2016){Karaletsos}, {Belongie}, and
	{R{\"a}tsch}]{karaletsos16}
	T.~{Karaletsos}, S.~{Belongie}, and G.~{R{\"a}tsch}.
	\newblock Bayesian representation learning with oracle constraints.
	\newblock In \emph{International Conference on Learning Representations}, 2016.
	
	\bibitem[Kleindessner and von Luxburg(2014)]{kleindessner14}
	M.~Kleindessner and U.~von Luxburg.
	\newblock Uniqueness of ordinal embedding.
	\newblock In \emph{Conference on Learning Theory}, 2014.
	
	\bibitem[Kruskal(1964)]{kruskal64}
	J.~B. Kruskal.
	\newblock Nonmetric multidimensional scaling: A numerical method.
	\newblock \emph{Psychometrika}, 1964.
	
	\bibitem[LeCun et~al.(1998)LeCun, Bottou, Bengio, and Haffner]{lecun98}
	Y.~LeCun, L.~Bottou, Y.~Bengio, and P.~Haffner.
	\newblock Gradient-based learning applied to document recognition.
	\newblock \emph{Proceedings of the IEEE}, 1998.
	
	\bibitem[Murray et~al.(2010)Murray, Adams, and MacKay]{murray10}
	I.~Murray, R.~P. Adams, and D.~J.~C. MacKay.
	\newblock Elliptical slice sampling.
	\newblock In \emph{Artificial Intelligence and Statistics}, 2010.
	
	\bibitem[Rasmussen and Williams(2006)]{rasmussen06}
	C.~E. Rasmussen and C.~K.~I. Williams.
	\newblock \emph{{G}aussian {P}rocesses for {M}achine {L}earning}.
	\newblock MIT Press, 2006.
	
	\bibitem[Schultz and Joachims(2004)]{schultz04}
	M.~Schultz and T.~Joachims.
	\newblock Learning a distance metric from relative comparisons.
	\newblock In \emph{Advances in Neural Information Processing Systems}. 2004.
	
	\bibitem[Shepard(1962)]{shepard62}
	R.~N. Shepard.
	\newblock The analysis of proximities: Multidimensional scaling with an unknown
	distance function.
	\newblock \emph{Psychometrika}, 1962.
	
	\bibitem[Tamuz et~al.(2011)Tamuz, Liu, Belongie, Shamir, and Kalai]{tamuz11}
	O.~Tamuz, C.~Liu, S.~Belongie, O.~Shamir, and A.~T. Kalai.
	\newblock Adaptively learning the crowd kernel.
	\newblock In \emph{International Conference on Machine Learning}, 2011.
	
	\bibitem[Terada and {von~Luxburg}(2014)]{terada14}
	Y.~Terada and U.~{von~Luxburg}.
	\newblock Local ordinal embedding.
	\newblock In \emph{International Conference on Machine Learning}, 2014.
	
	\bibitem[Ukkonen et~al.(2015)Ukkonen, Derakhshan, and Heikinheimo]{ukkonen15}
	A.~Ukkonen, B.~Derakhshan, and H.~Heikinheimo.
	\newblock Crowdsourced nonparametric density estimation using relative
	distances.
	\newblock In \emph{HCOMP}, 2015.
	
	\bibitem[van~der Maaten and Weinberger(2012)]{maaten12}
	L.~van~der Maaten and K.~Q. Weinberger.
	\newblock Stochastic triplet embedding.
	\newblock In \emph{{IEEE} International Workshop on Machine Learning for Signal
		Processing}, 2012.
	
	\bibitem[Wolberg and Mangasarian(1990)]{wolberg90}
	W.~H. Wolberg and O.~L. Mangasarian.
	\newblock Multisurface method of pattern separation for medical diagnosis
	applied to breast cytology.
	\newblock \emph{Proceedings of the National Academy of Sciences}, 1990.
	
\end{thebibliography}


\newpage

\renewcommand\appendixpagename{Supplementary material}

\begin{appendices}
In the following, we shortly recap the GNMDS and $t$-STE embedding methods.

\section{Embedding algorithms}

{\bf Generalized Non-Metric Multidimensional Scaling} (GNMDS) by \citet{agarwal07}
aims to find a kernel matrix $\vec{K} = \vec{X}\vec{X}^T$ that satisfies the given triplet constraints in $\mathcal{S}$ with a large margin. It minimizes the trace norm of the corresponding kernel matrix to obtain a low dimensional embedding, which leads to the following semidefinite program:
\begin{align*}
\begin{array}{ll}
\min\limits_{\vec{K} \in \vec{S}^+}  \myspur{\vec{K}} + C \cdot \sum\limits_{\left( ijl \right) \in \mathcal{S}} \xi_{ijl}  &\vspace{3pt}\\
\hspace{4pt}k_{jj} - 2 \cdot k_{ij} + k_{ll} + 2 \cdot k_{il} \leq 1 +  \xi_{ijl}\vspace{3pt}& \forall \left( i,j,l \right) \in \mathcal{S}\\
\hspace{4pt}\xi_{ijl} \geq 0\vspace{3pt}& \forall \left( i,j,l \right) \in \mathcal{S}.
\end{array}
\end{align*}
The final embedding $\vec{X}$ can be computed by an SVD of the kernel matrix $\vec{K} = \vec{X}\vec{X}^T$. Replacing rank by trace, this program solves a relaxed version of the ordinal embedding problem, and hence one may not always find the correct embedding. 

{\bf $t$-Distributed Stochastic Triplet Embedding} ($t$-STE, \citealp{maaten12} ) is a more robust version of STE that replaces the Gaussian distribution by the more heavy-tailed $t$-distribution with $\alpha$ degrees of freedom. As the authors suggested, we use $\alpha = d-1$. The final embedding is constructed by minimizing the negative log-likelihood with
\begin{equation}
\label{eq:tSTELikelihood}
p_{ijl} = \frac{\left(1+\frac{\mynorm{\vec{x_i}-\vec{x_j}}^2}{\alpha}\right)^{-\frac{\alpha +1 }{2}}}{ \left(1+\frac{\mynorm{\vec{x_i}-\vec{x_j}}^2}{\alpha}\right)^{-\frac{\alpha +1 }{2}} + \left(1+\frac{\mynorm{\vec{x_i}-\vec{x_l}}^2}{\alpha}\right)^{-\frac{\alpha +1 }{2}}}.
\end{equation}

\begin{figure}[t]
\centering

\subcaptionbox{Gaussian Process for $l = 0.54$.}[.4\textwidth]{
	\centering
	\includegraphics[width=.4\textwidth]{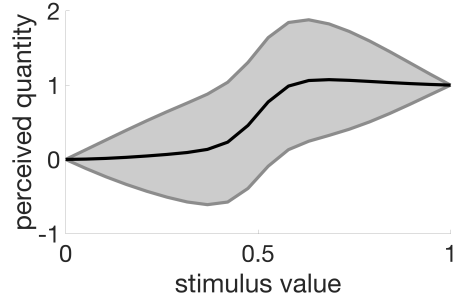}
}\\
\subcaptionbox{Three embeddings with point uncertainties as errorbars.}[.375\textwidth]{
	\centering
	\includegraphics[width=.375\textwidth]{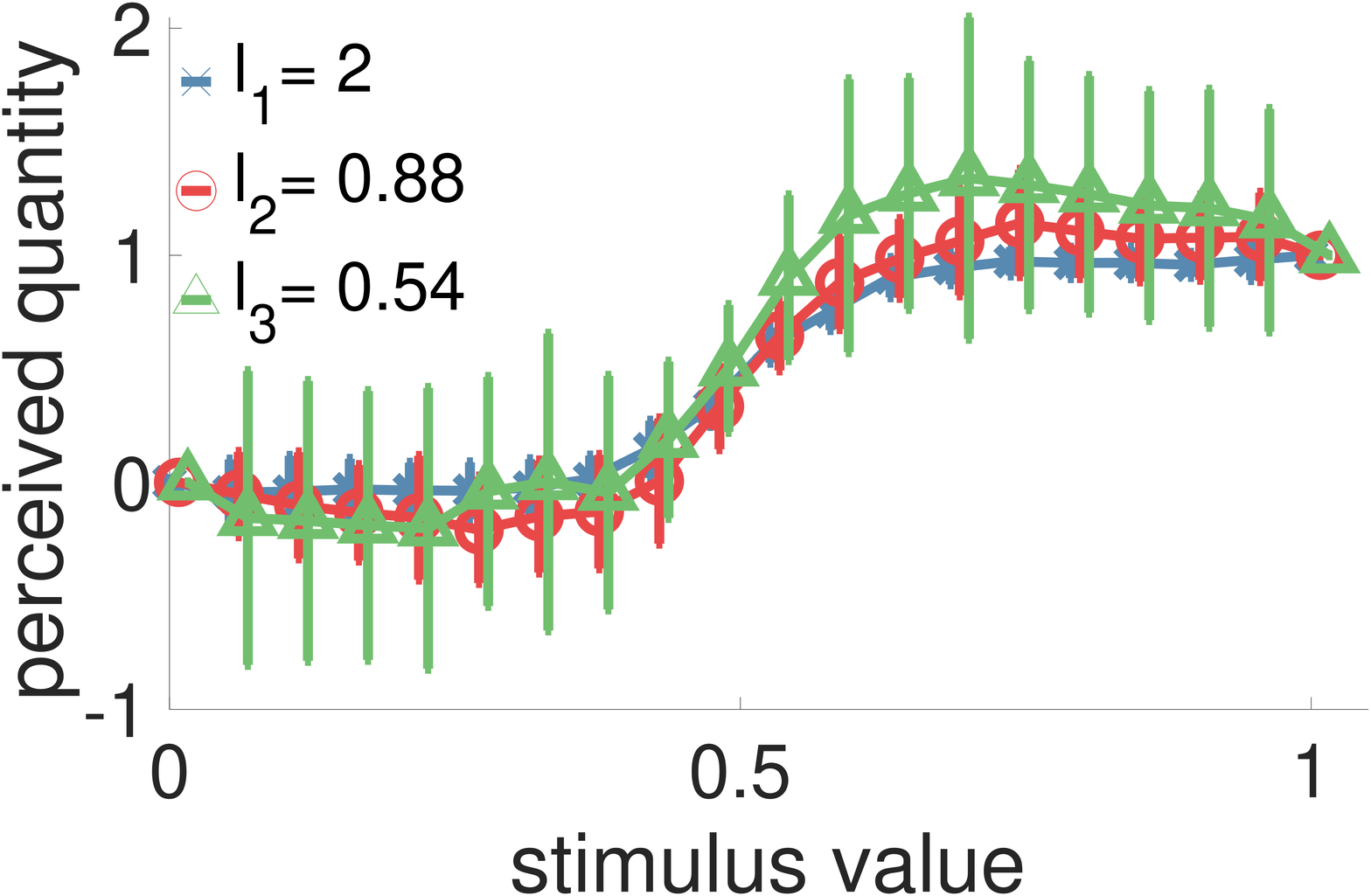}
}
\caption{(a) The upper figure illustrates a Gaussian Process. The solid line is the mean function and the shaded area lies within the standard deviation. (b) For three lengthscales we conduct our psychophysics experiment. We plot the mean embedding of one bootstrap and standard deviation of the points as error bars.}\label{fig:psycho_supp}
\end{figure}

\section{Bayesian approach for sampling embeddings}

For a {\bf prior over the distance matrix $\vec{D}$} we take a simple approach. We employ a multivariate Gaussian prior over the distance matrix and encode symmetry. For a matrix $\vec{D} \in \R^{n \times n}$, we stack the rows into a vector and denote it with $\arrowVec{\vec{D}} \in \R^{n^2}$. The function $\mathbbm{1}_{ik}$ is the indicator function where $\mathbbm{1}_{ik}$ equals $1$ when $i=k$ and $0$ otherwise. The operator $\otimes$ denotes the usual Kronecker product. Following the definition of \citet{bartels16}, symmetry is encoded in the covariance matrix of the Gaussian prior via the symmetric Kronecker product. It uses a matrix $\vec{\Gamma} \in \R^{n^2 \times n^2}$ with $ [\vec{\Gamma}]_{(i,j)(k,l)} = 0.5 ( \mathbbm{1}_{ik} \mathbbm{1}_{jl} + \mathbbm{1}_{il} \mathbbm{1}_{kj})$. The symmetric Kronecker product is defined as $ \vec{A} \circledast \vec{C} := \vec{\Gamma} (\vec{A} \otimes \vec{C}) \vec{\Gamma}$. Any positive semi-definite matrix $\vec{W} \in \R^{n \times n}$, leads to a prior distribution that encodes symmetry of the resulting matrix $\arrowVec{\vec{D}} \sim \mathcal{N}\!(\arrowVec{\vec{D}}_0, \vec{W} \circledast \vec{W})$ with mean $\arrowVec{\vec{D}}_0$. However, as described in the main paper, this approach fails to incorporate more specific properties of distance matrices.

Combining this prior with a triplet likelihood function yields the {\bf posterior} distribution
\begin{equation}
\label{eq:posteriorD}
p\left(\vec{D} | \mathcal{S}\right) \propto \mathcal{N}\!\left(\arrowVec{\vec{D}}_0,\vec{W} \circledast \vec{W}\right) \ell\!\left(\vec{D}\right).
\end{equation}
This posterior distribution is not analytically tractable. However, we can also use the MCMC framework of elliptical slice sampling (ESS) by \citet{murray10} since it is applicable for multivariate Gaussian priors and any likelihood function.

	 \begin{figure*}[t]
\centering

\subcaptionbox*{}[.3\textwidth]{
	\centering
	\includegraphics[width=.3\textwidth]{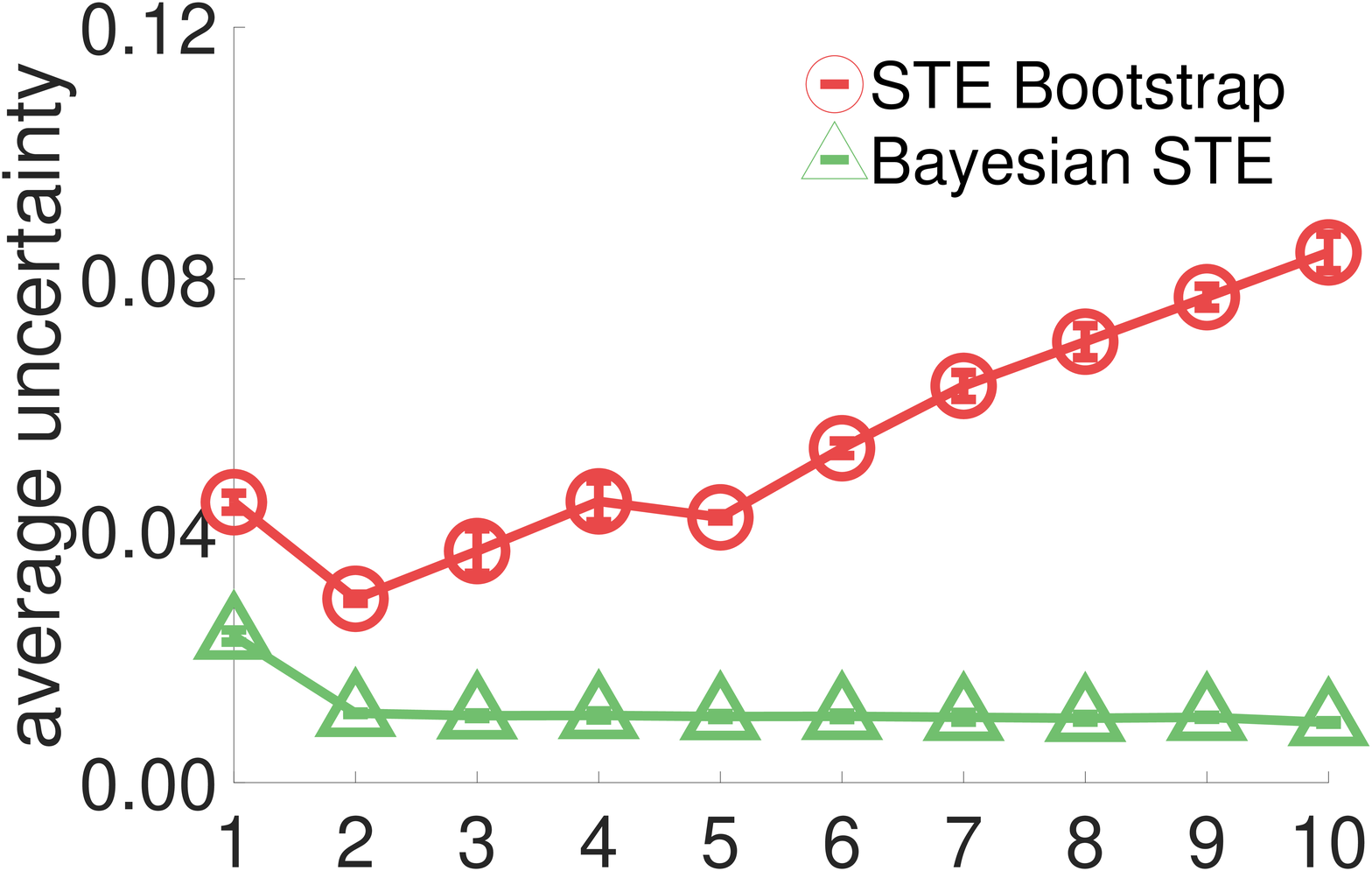}
}\hspace{15pt}
\subcaptionbox*{}[.3\textwidth]{
	\centering
	\includegraphics[width=.3\textwidth]{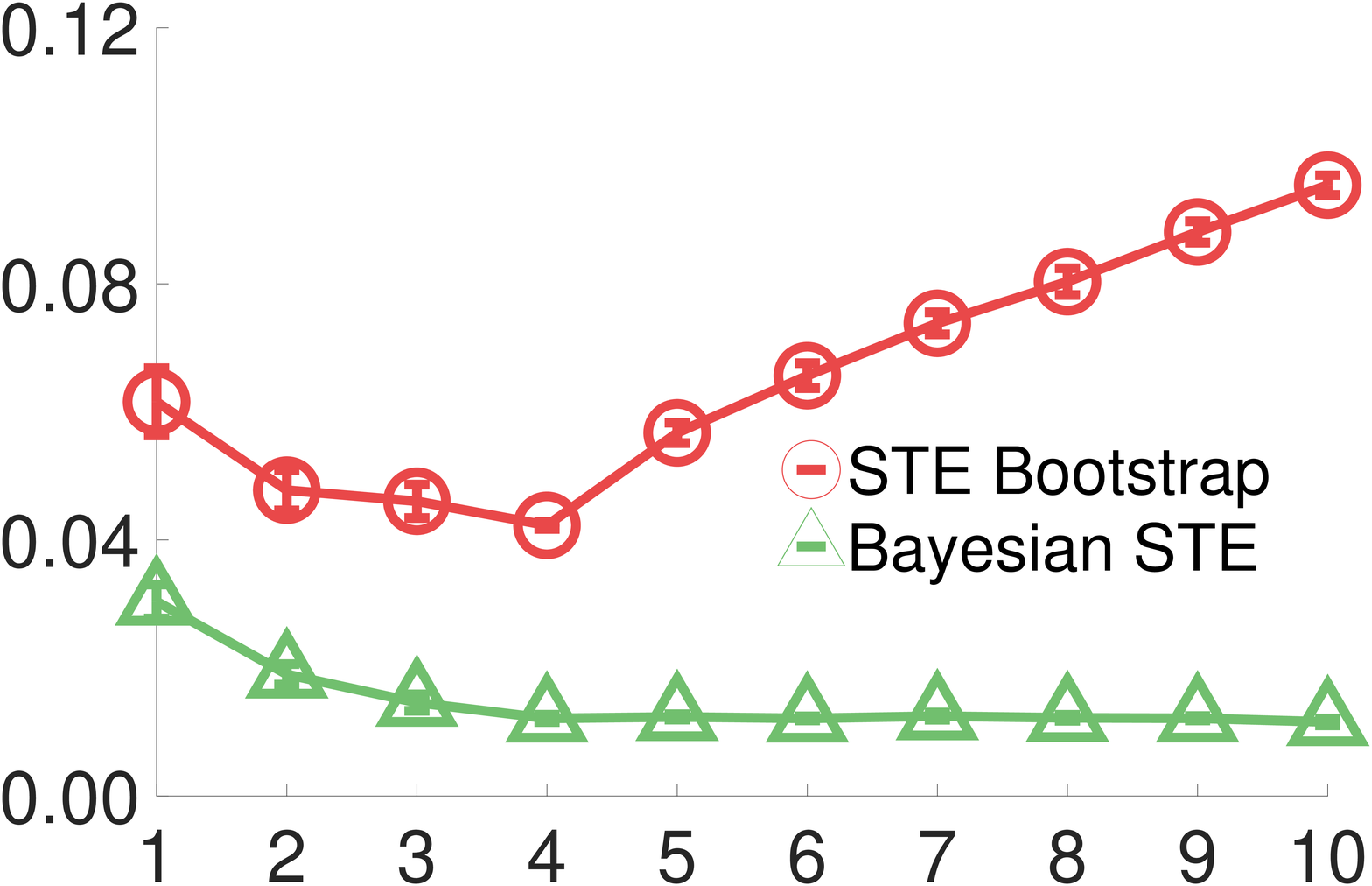}
}\hspace{15pt}
\subcaptionbox*{}[.3\textwidth]{
	\centering
	\includegraphics[width=.3\textwidth]{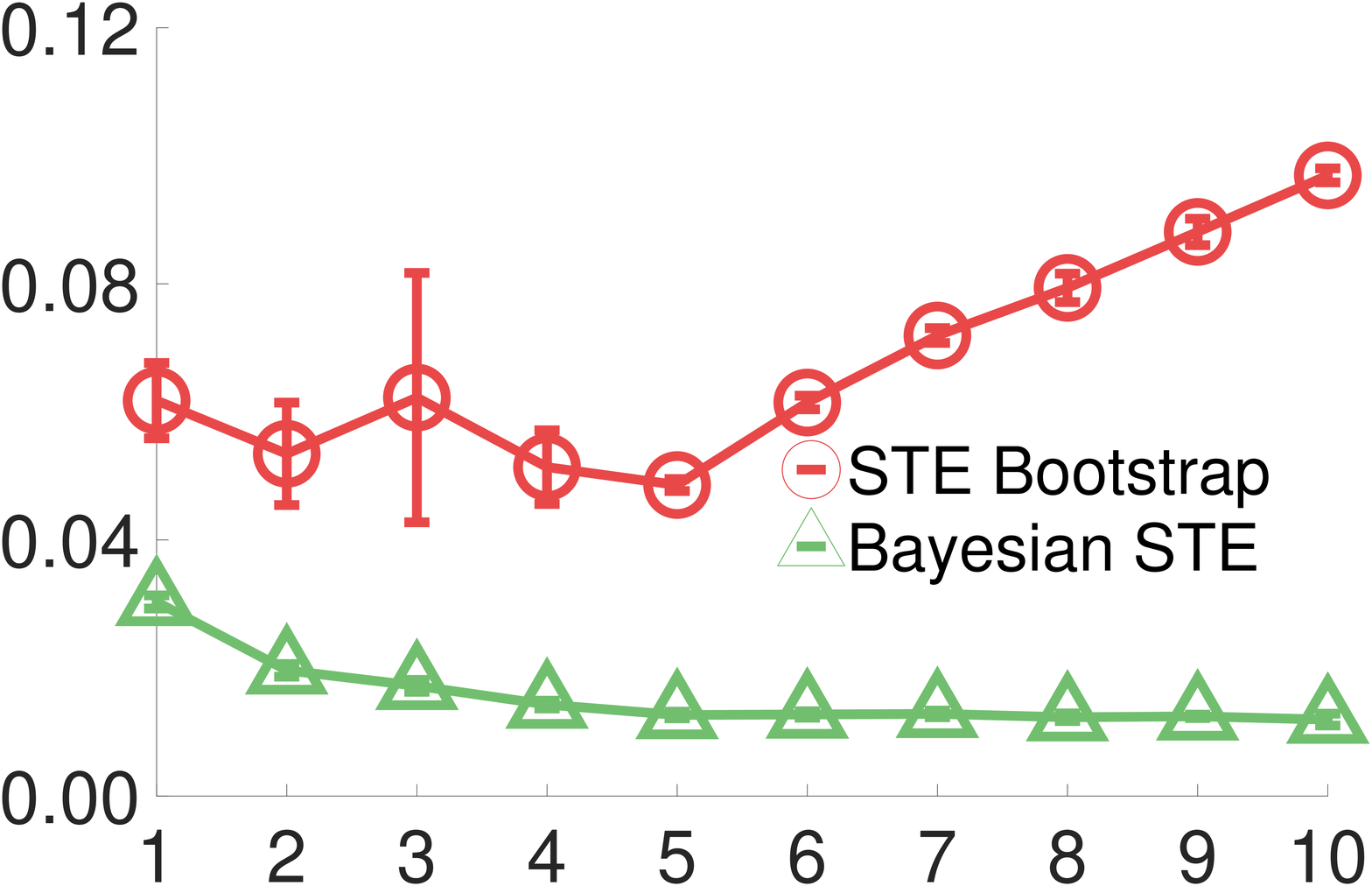}
}\\
\subcaptionbox{$d_{\mathrm{true}} = 2$.\label{fig:increasingDimension_supp}}[.3\textwidth]{
	\centering
	\includegraphics[width=.3\textwidth]{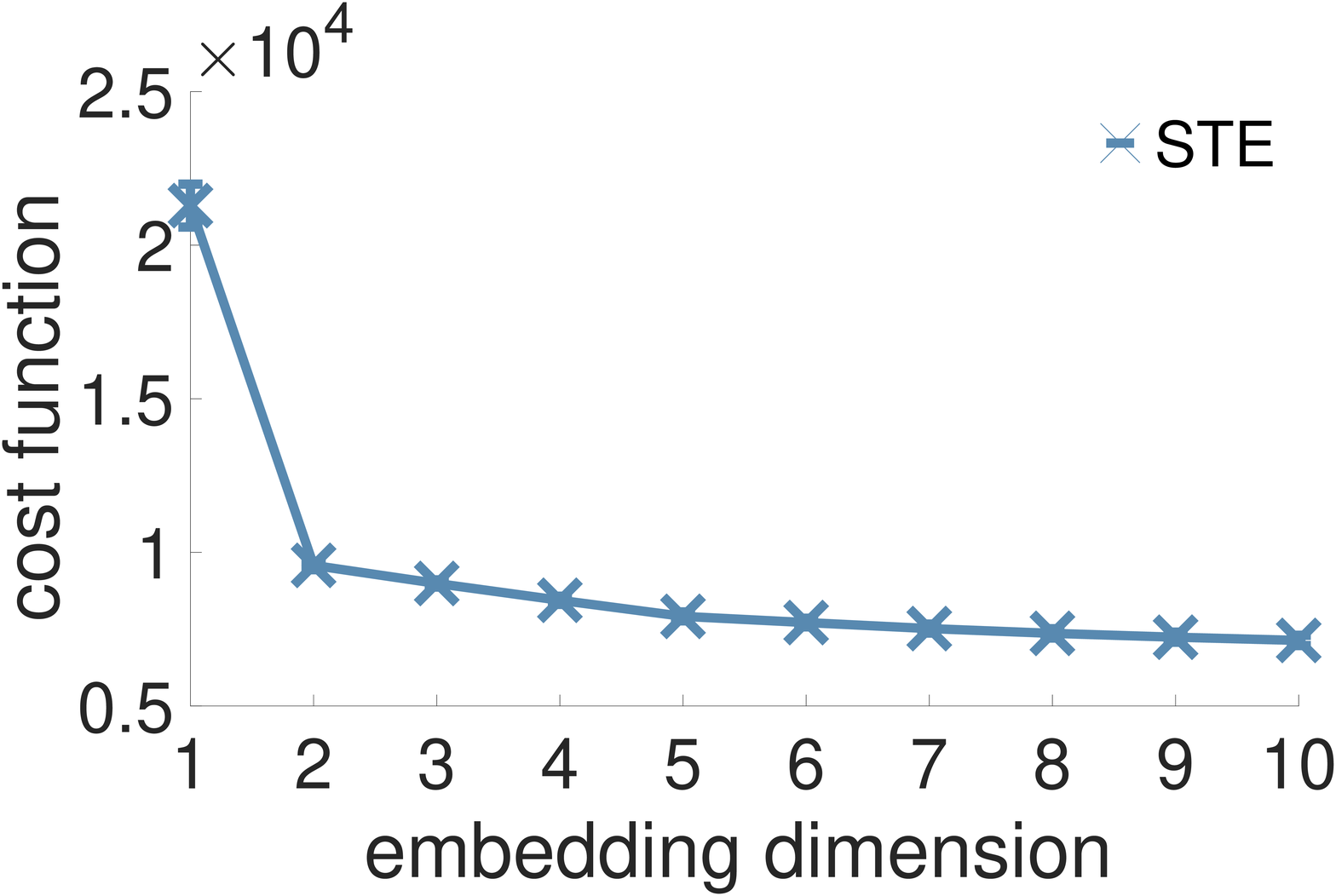}
}\hspace{15pt}
\subcaptionbox{$d_{\mathrm{true}} = 4$.\label{fig:increasingNoise_supp}}[.3\textwidth]{
	\centering
	\includegraphics[width=.3\textwidth]{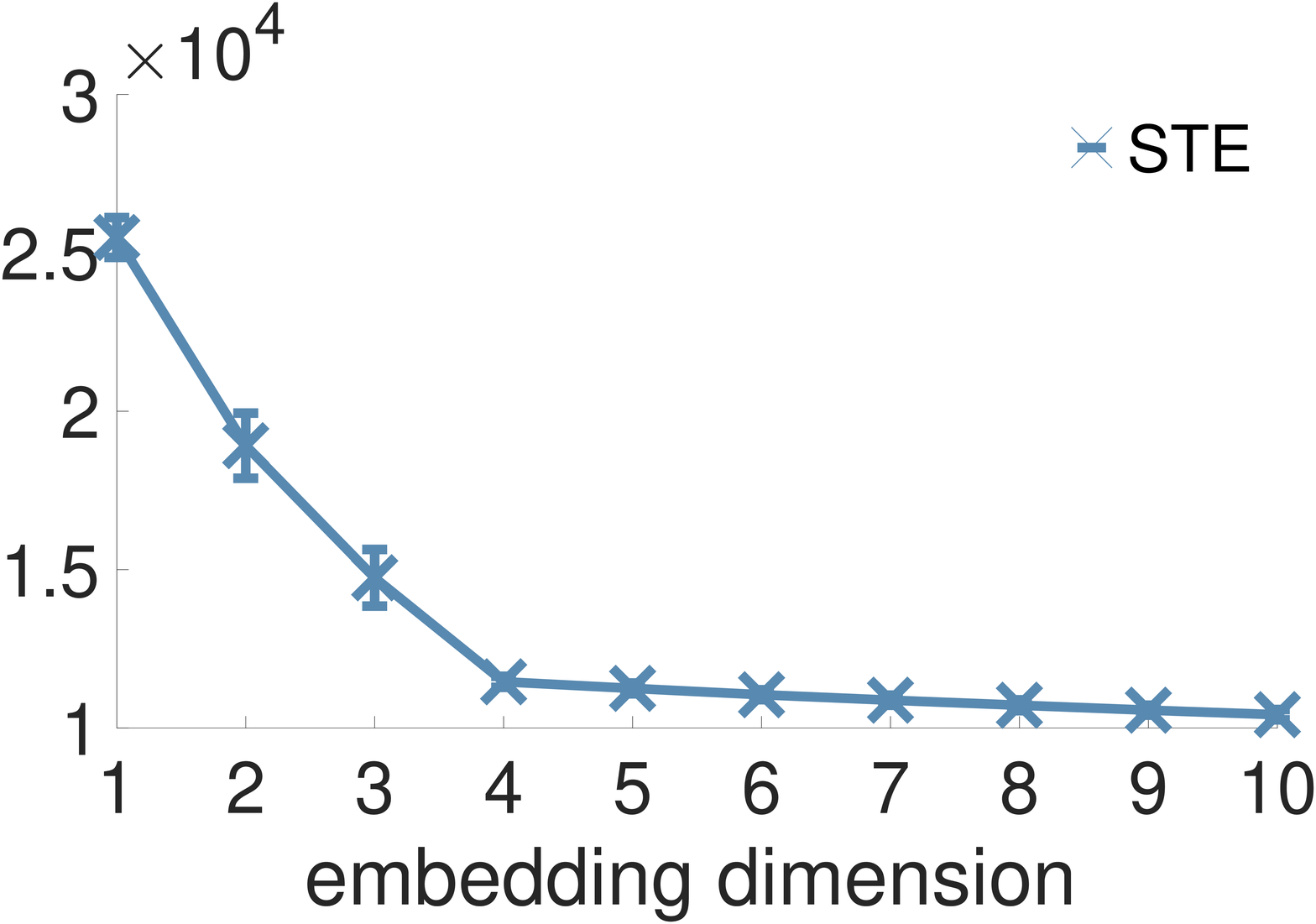}
}\hspace{15pt}
\subcaptionbox{$d_{\mathrm{true}} = 5$.\label{fig:increasingTriplets_supp}}[.3\textwidth]{
	\centering
	\includegraphics[width=.3\textwidth]{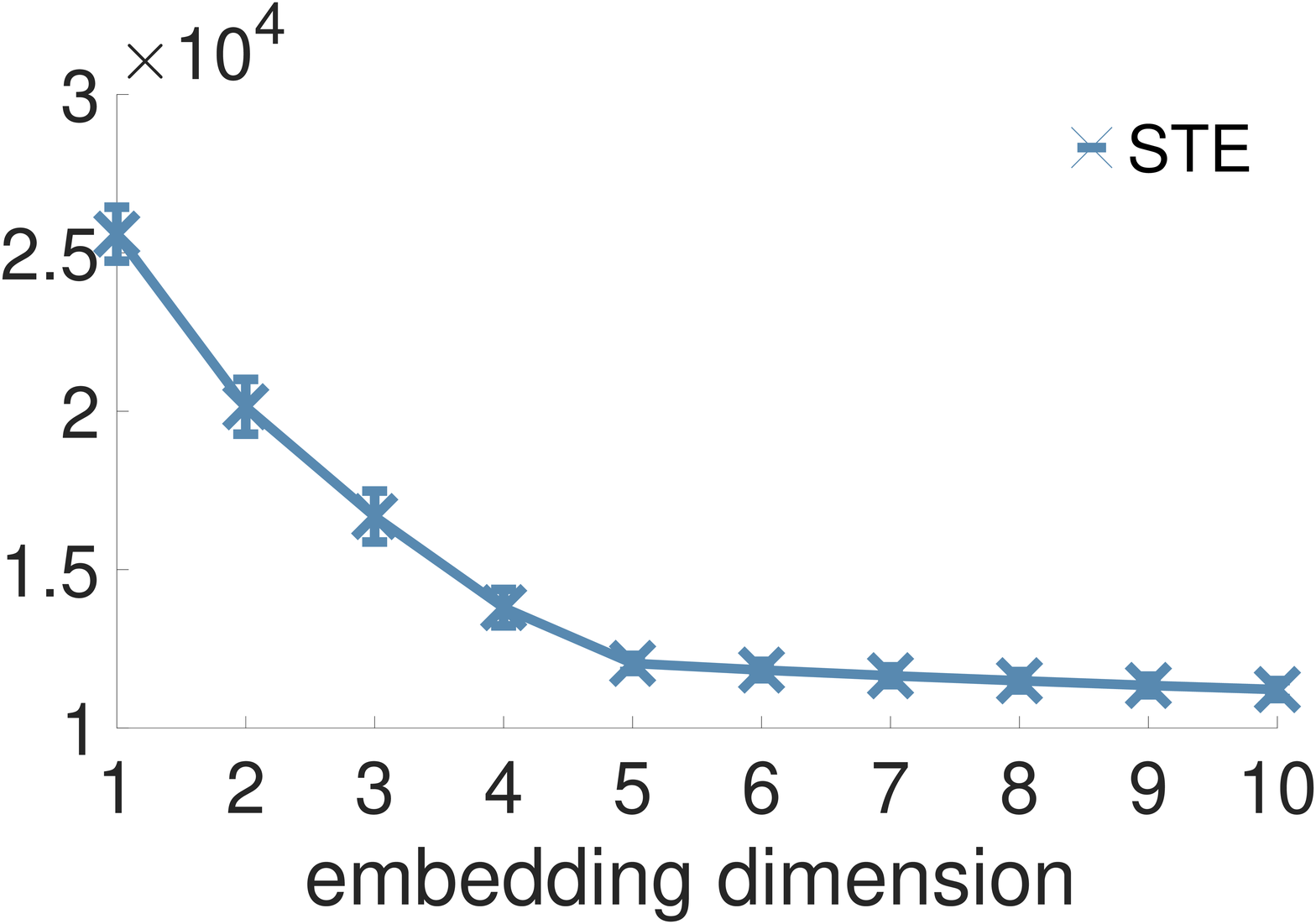}
}
\caption{a) We embed noisy triplets that were created from a projection of MNIST on $d_{\mathrm{true}}$ principal components. The average triplet uncertainty of the boostrap is minimal for $d_{\mathrm{true}}$ whereas the STE cost function decreases further.}
\end{figure*}

	\section{Are the uncertainty estimates well calibrated?}
\begin{figure}[t]
\centering

\subcaptionbox*{}[.4\textwidth]{
	\centering
	\includegraphics[width=.4\textwidth]{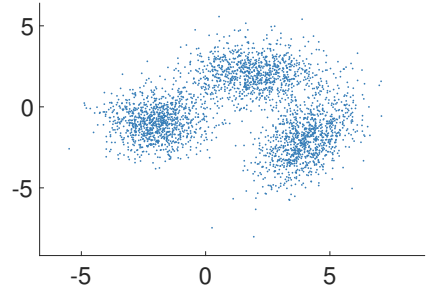}
}
\caption{This two-dimensional mixture of three Gaussians is used to examine the calibration of the uncertainty estimates.}\label{fig:GaussianMixture}
\end{figure}

	As reported in the experiments of Section 5.1, other embedding methods were used to examine if the uncertainty estimates are well calibrated.  As in the main paper, we use a two-dimensional mixture of three Gaussians (see Figure~\ref{fig:GaussianMixture}). The means of the three components are $\mu_1 = \left[2,2\right] , \mu_2 = \left[-2,-1\right], \mu_3 = \left[4,-2\right]$ and the covariance matrices are $\Sigma_1 = [2,0; 0,1], \Sigma_2 = [1,0; 0,1]$ and  $\Sigma_3 = [1, 0.7; 0.7,2] $. We randomly draw $50$ points, generate triplets, and embed again in $\R^2$. We use $t$-STE and GNMDS to see how our uncertainty estimates behave under more noise (see Figure \ref{fig:increasingNoise_supp}) or an increasing number of training triplets (see Figure \ref{fig:increasingTriplets_supp}). Note, that GNMDS does not use a probabilistic model, and therefore, we just perform a bootstrap for GNMDS. For the overall uncertainty, we compute the average over the uncertainties of all true triplets, and, for visualization purposes, flip the average on $0.5$ such that the uncertainty decreases to $0$ rather than increases to $1$. 

\begin{figure*}
	 	\centering
	 	\subcaptionbox{Triplet prediction error.}{
	 		\centering
	 		\includegraphics[width=0.46\textwidth]{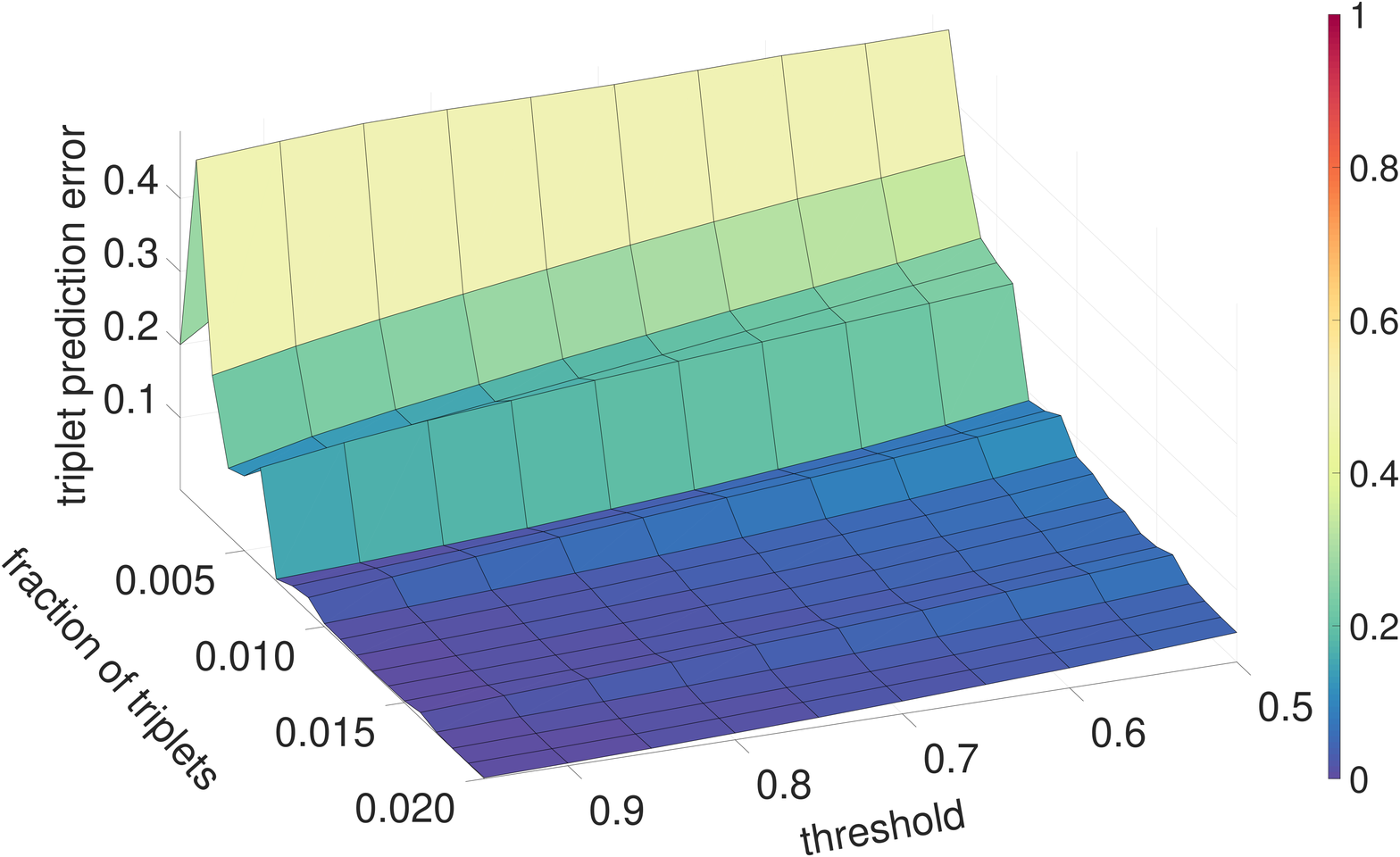}
	 	}
	 	\subcaptionbox{Abstention rate.}{
	 		\centering
	 		\includegraphics[width=0.46\textwidth]{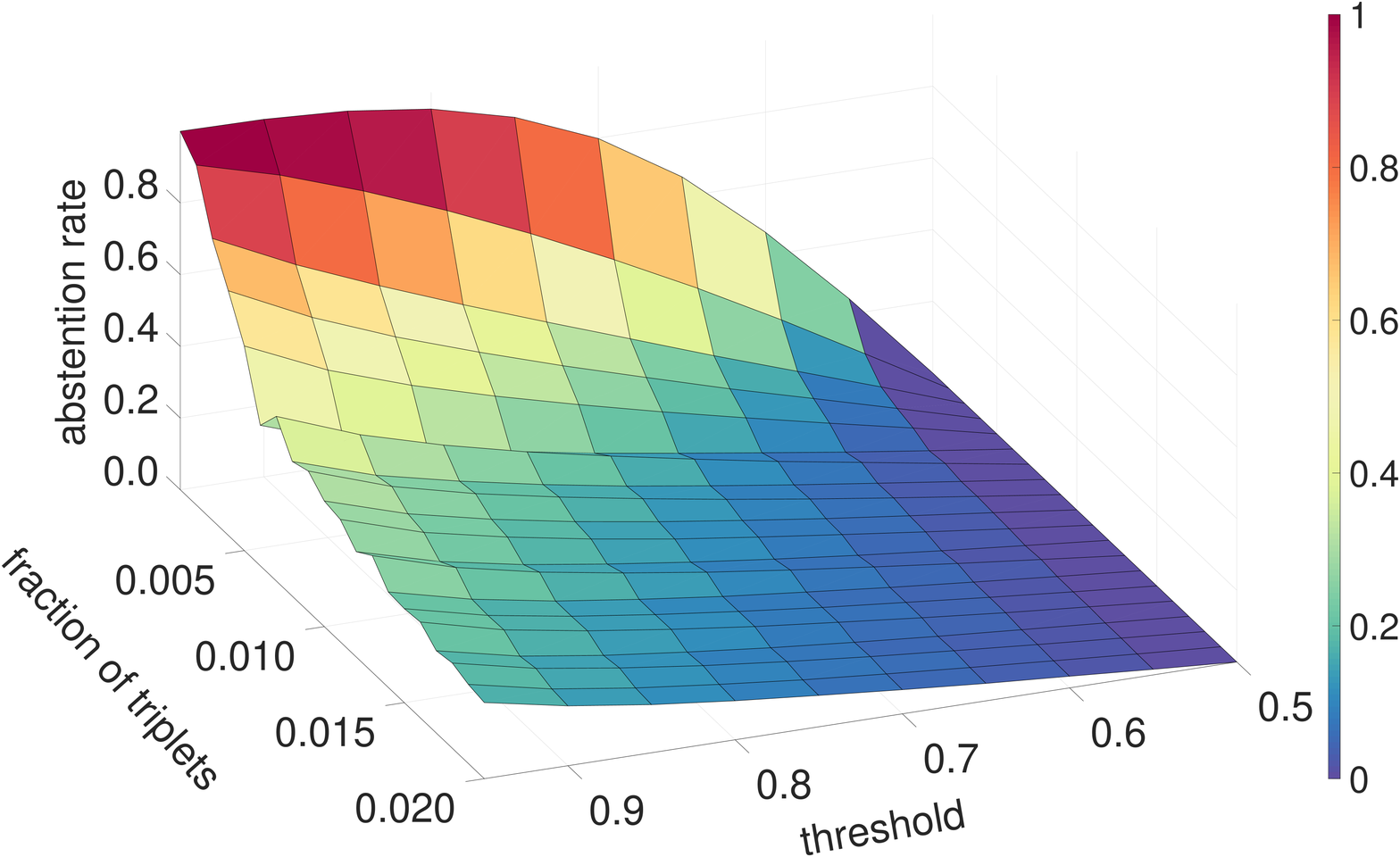}
	 	}
	 	\caption{Triplet prediction with Bayesian STE performed on $50$ points from a two-dimensional mixture of three Gaussians. A triplet set of variable size is used to generate uncertainty estimates for all triplets. If the estimate is above a threshold $t$ (or below $1-t$), we make a prediction, otherwise we abstain. a) The triplet prediction error decreases when a higher threshold for certainty is required, or when the number of triplets increases. b) The abstention rate increases, when the threshold increases, but decreases when more triplets are used.} \label{fig:tripletPrediction_supp}
	 \end{figure*}

	Additionally, we performed the task of triplet prediction, as described in Section~\ref{sec:UsingEstimates}, on the same data set. Here, we report the results for our Bayesian STE method in Figure \ref{fig:tripletPrediction_supp}. The abstention rate decreases faster than for the STE Bootstrap with an increasing number of triplets. On the other hand, the prediction error for Bayesian STE is quite high when only very few triplets are available, whereas the STE Bootstrap abstains more in this regime, which leads to a low triplet prediction error. 
	
	 \section{Estimating the embedding dimension}
	 
    In order to evaluate how well our uncertainty estimates are calibrated, we report the overall uncertainty which is the average over the uncertainties of all true triplets. When estimating the embedding dimension, we often have not access to all ground-truth triplets. Here, we proceed slightly differently when computing the average uncertainty: first, we compute the triplet uncertainty of every possible triplet $(i,j,l)$. If $\pi_{ijl}>0.5$, we include $1-\pi_{ijl}$ into the average, otherwise we include $\pi_{ijl}$. Since uncertainty is expressed only by the distance of $\pi_{ijl}$ to $0.5$, the average uncertainty computed this way is meaningful.         
	
	\begin{table}[h]
	\centering
	\caption{Summary Of Different Data Sets}
	\begin{tabular}{@{}cccc@{}}\toprule
		Name & Dimension & Size & Classes \\ 
		\midrule
		MNIST ($6$ and $8$) & $784$ & $11,769$ & $2$\\         
		Breast Cancer & $30$ & $399$ & $2$ \\         
		Satellite & $36$ &$4,435$ & $6$\\         
		\bottomrule
	\end{tabular}
	\label{tab:Datasets}
    \end{table}
	
	\section{Application in psychophysics}
	
	We model the perception functions with a Gaussian Process as described in Section~\ref{sec:ExpPsycho}. We fix all functions to be the identity function to map $0$ on $0$ and $1$ on $1$. We summarize our training points in $\vec{x} := (0,1)^T$ with the outcome $\vec{y} := (0,1)^T$. When we use the kernel function on two vectors, we evaluate covariances of all pairs of entries, hence, $k(\vec{x},\vec{x^\prime})$ denotes the matrix of pairwise covariances. Then, our posterior kernel is given by $k_{\mathrm{post}}(x_i,x_j) = k(x_i,x_j) - k(x_i,\vec{x}) \left( \vec{K}\left(\vec{x},\vec{x}\right)\right)^{-1}  k(\vec{x},x_j)$. The posterior mean function is given by  $m_{\mathrm{post}}= m(x)$.

	\section{Active Learning}

	\citet{tamuz11} suggest an adaptive approach to select the next query using the triplet probabilities (\ref{eq:CKprobability}) developed for the Crowd Kernel embedding algorithm. This sampling strategy can also be used with the STE or $t$-STE triplet probabilities. By \emph{STE--IG} (or \emph{$t$-STE--IG}) we denote that we use the information gain (IG) sampling scheme with the STE (or $t$-STE) likelihood.  
	
\begin{figure*}[ht]
\subcaptionbox*{}[.175\textwidth]{
	\begin{minipage}{.175\textwidth}
	 i) Triplet Prediction
	\end{minipage}
	\vspace{40pt}
}
\subcaptionbox*{}[.32\textwidth]{
	\centering
	\includegraphics[width=.32\textwidth]{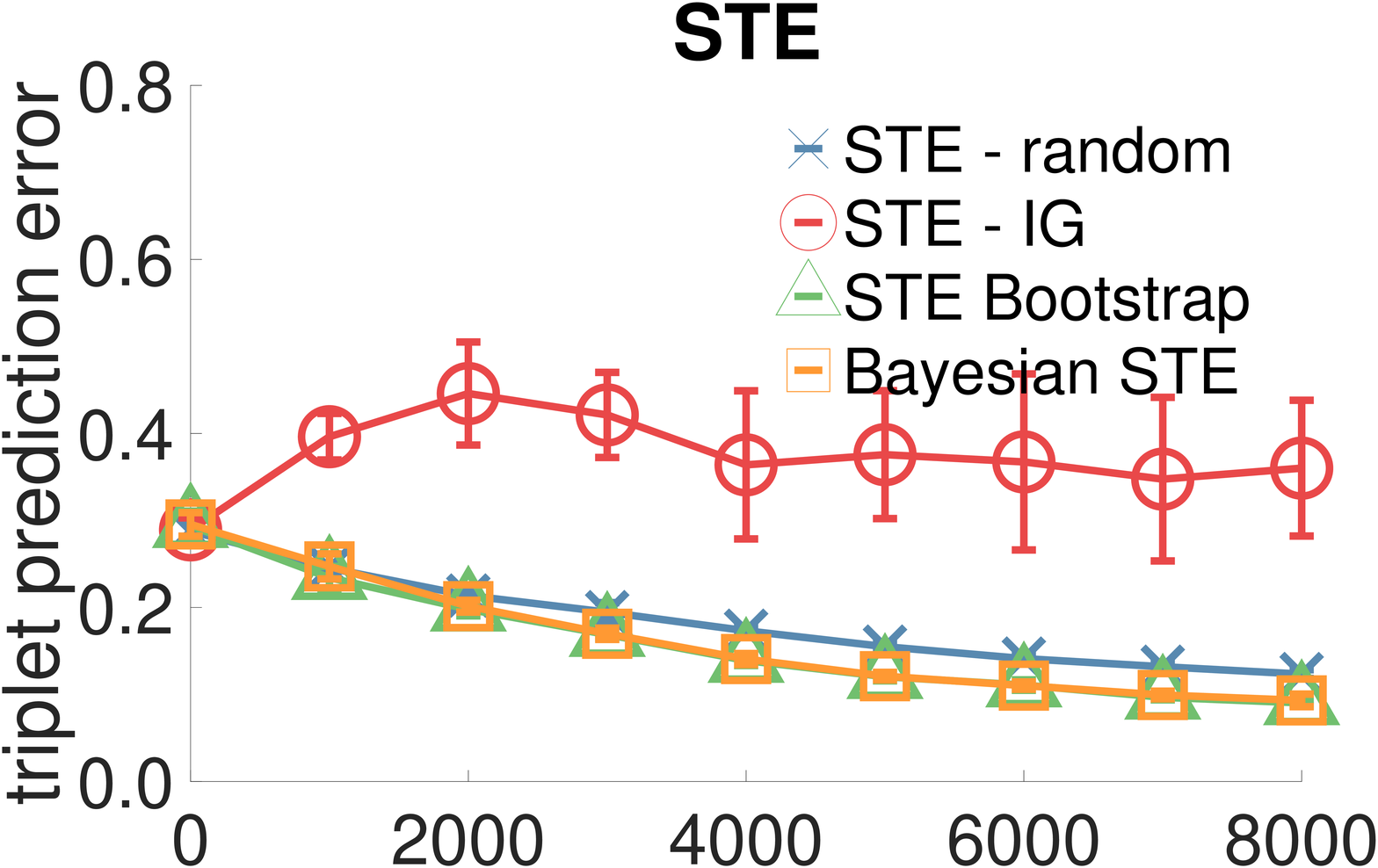}
}\hspace{4pt}
\subcaptionbox*{}[.32\textwidth]{
	\centering
	\includegraphics[width=.32\textwidth]{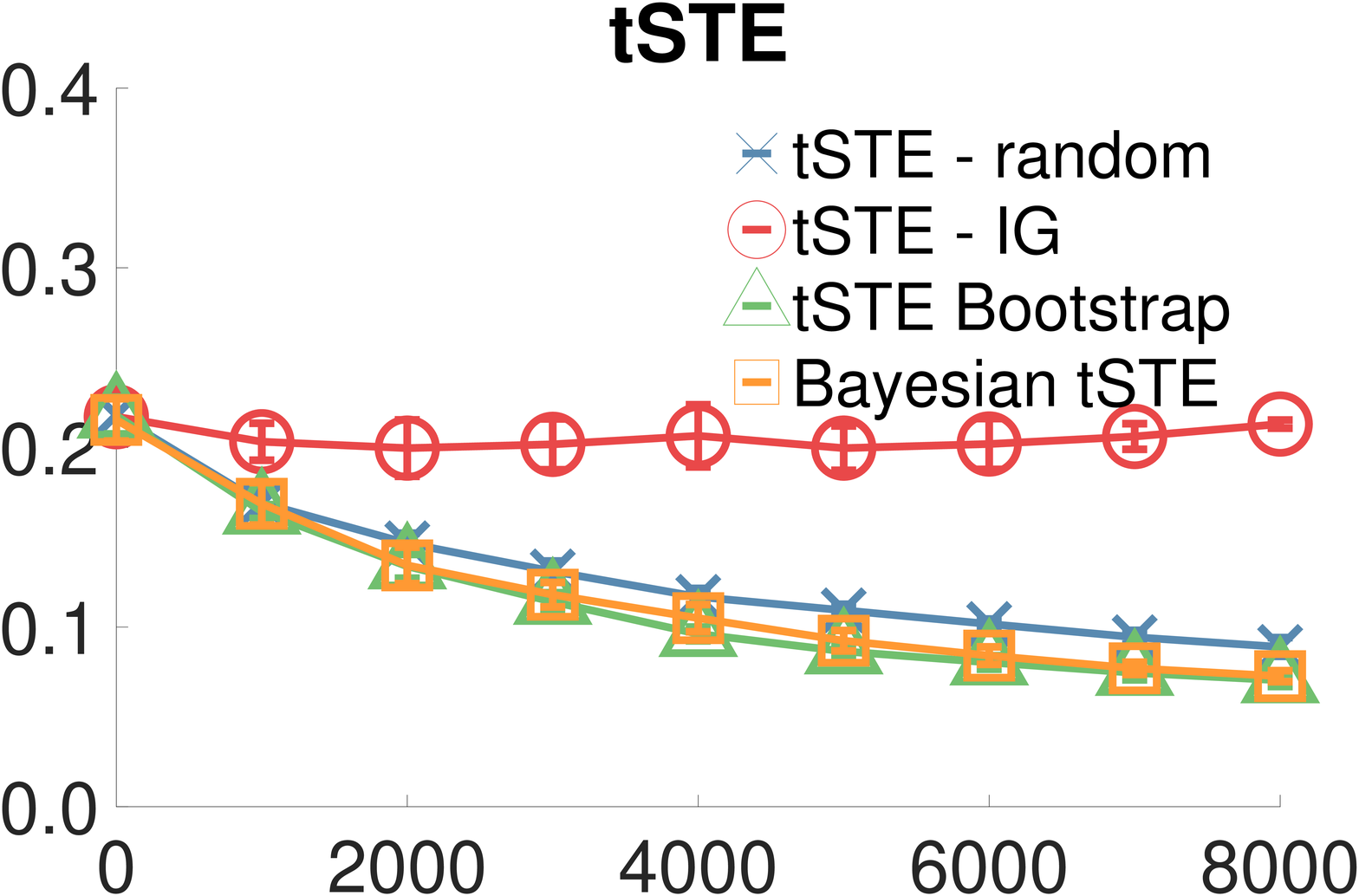}
}
\subcaptionbox*{}[.175\textwidth]{
	\begin{minipage}{.175\textwidth}
	 ii) Classification
	\end{minipage}
	\vspace{40pt}
}
\subcaptionbox*{}[.32\textwidth]{
	\centering
	\includegraphics[width=.32\textwidth]{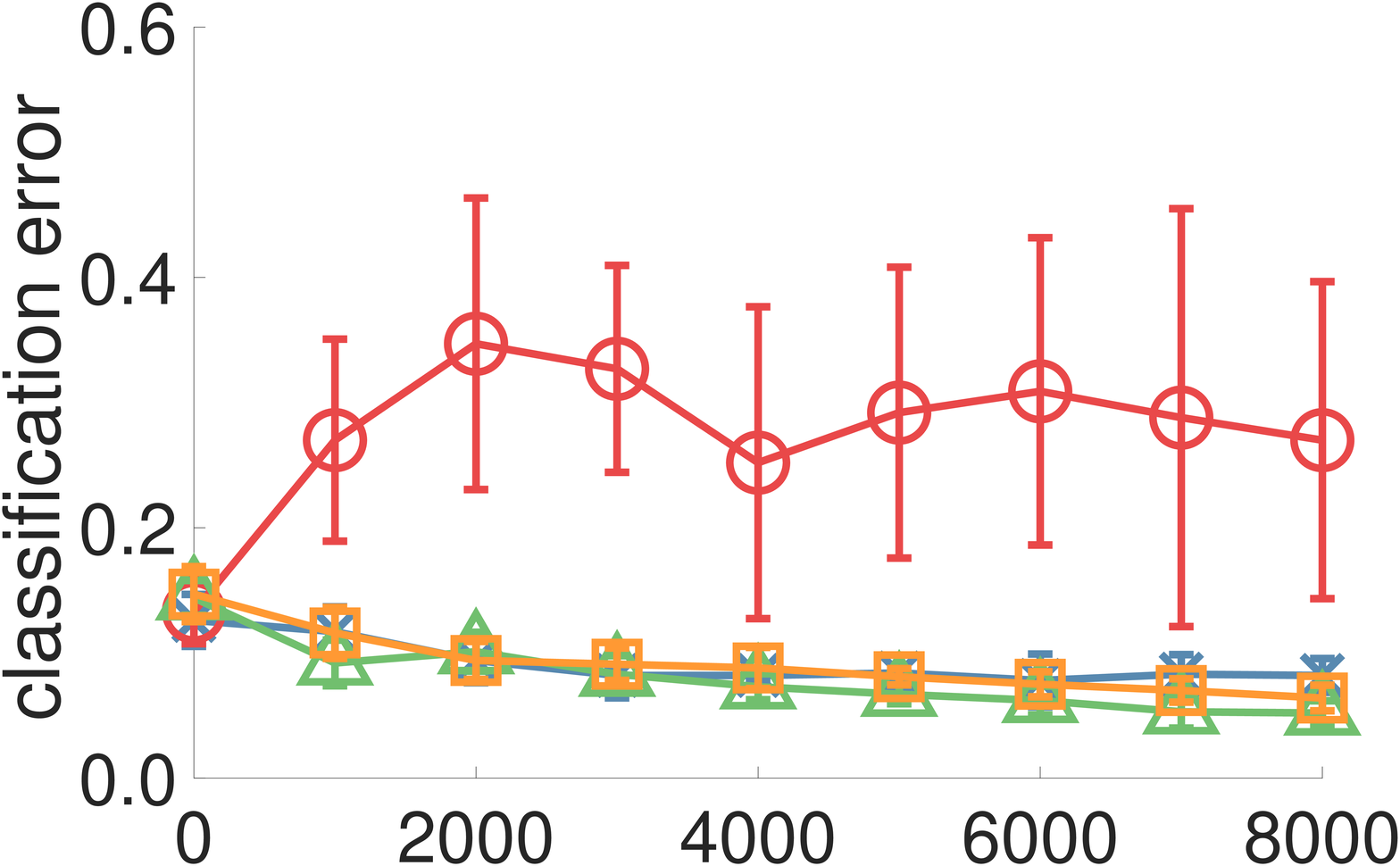}
}\hspace{4pt}
\subcaptionbox*{}[.32\textwidth]{
	\centering
	\includegraphics[width=.32\textwidth]{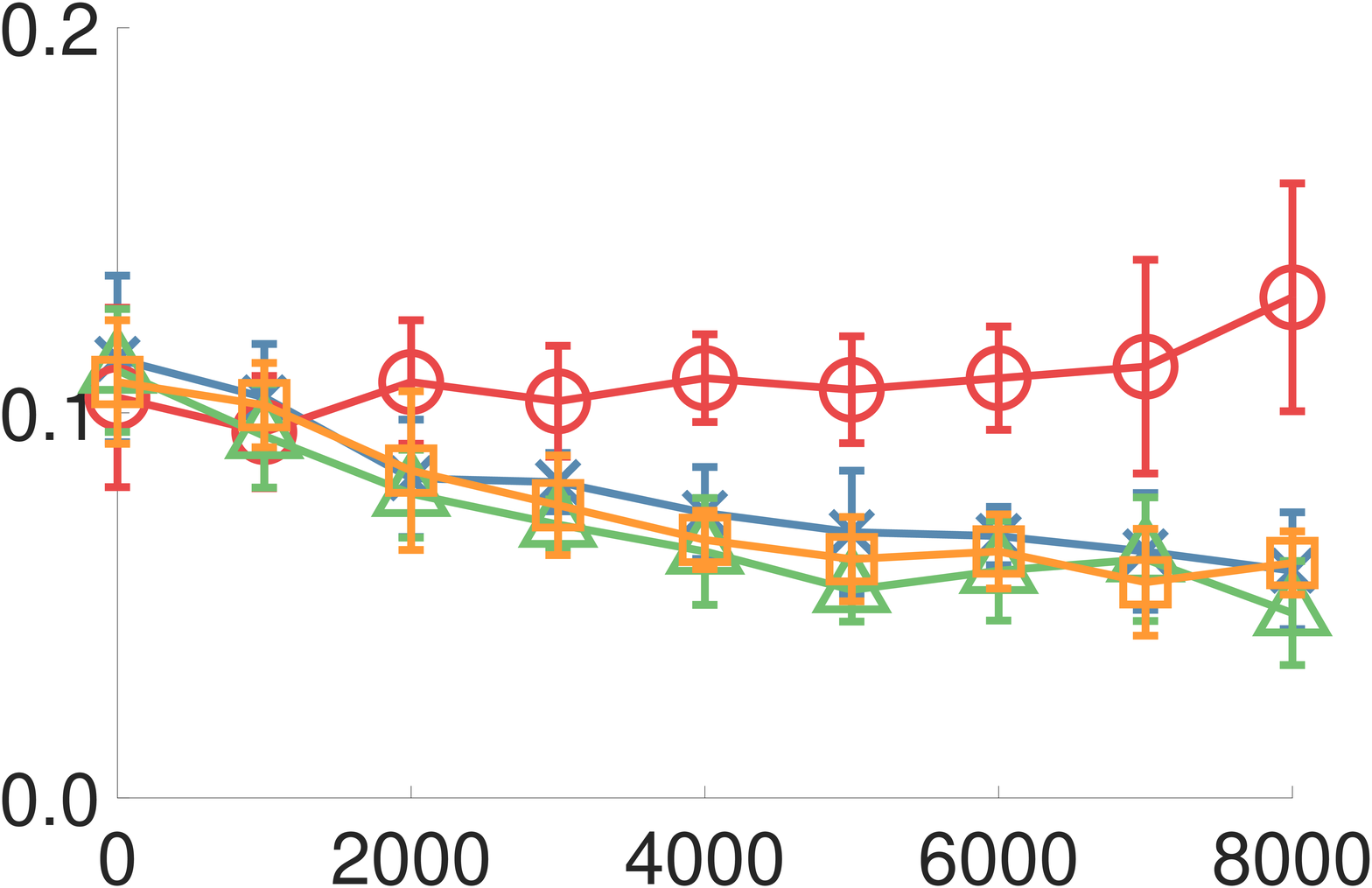}
}
\subcaptionbox*{}[.175\textwidth]{
	\begin{minipage}{.175\textwidth}
	 iii) Clustering
	\end{minipage}
	\vspace{50pt}
}\hspace{40pt}
\subcaptionbox{Comparing STE with its active approaches.}[.32\textwidth]{
	\centering
	\includegraphics[width=.32\textwidth]{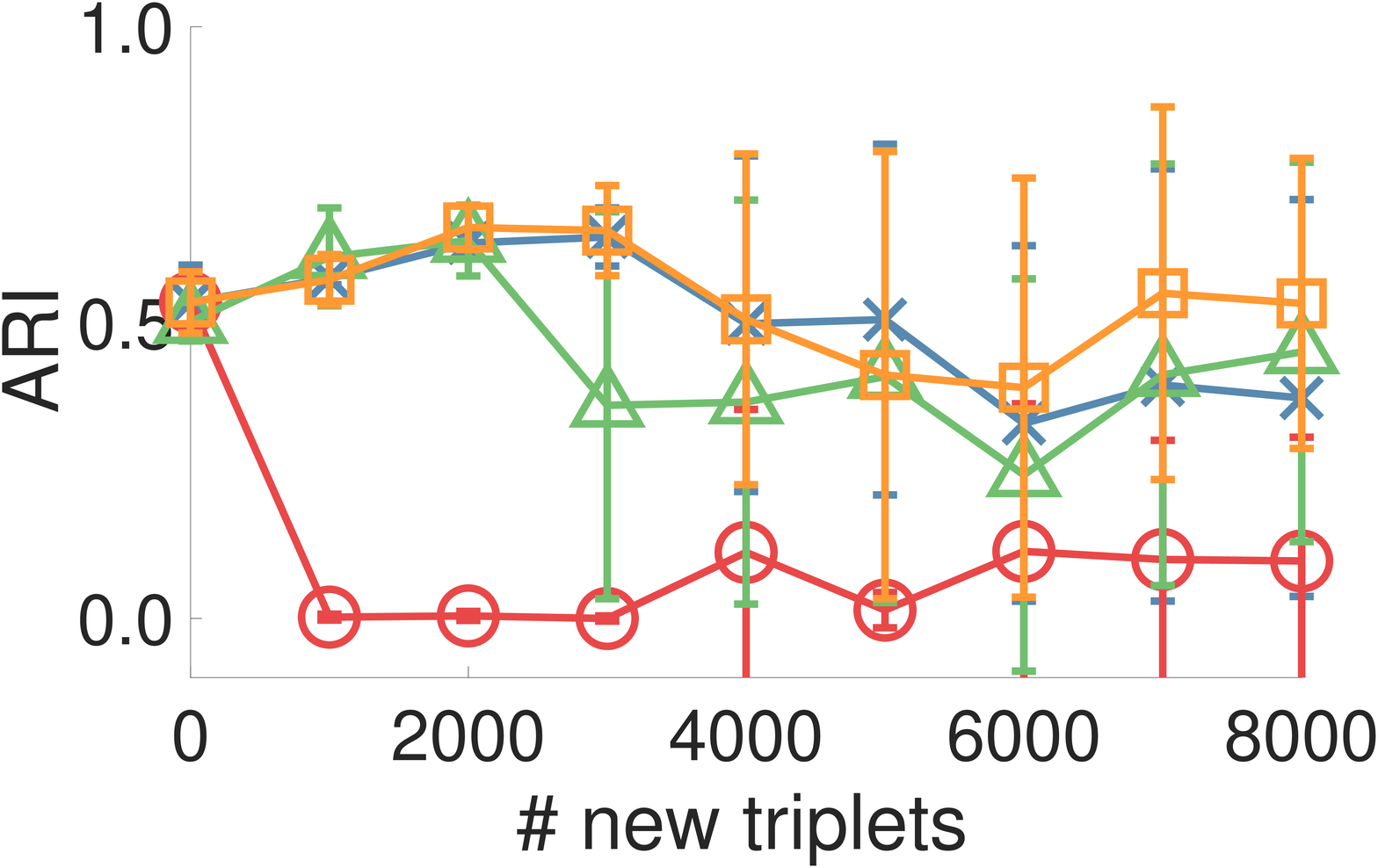}
}\hspace{45pt}
\subcaptionbox{ Comparing $t$-STE with its active approaches.}[.32\textwidth]{
	\centering
	\includegraphics[width=.32\textwidth]{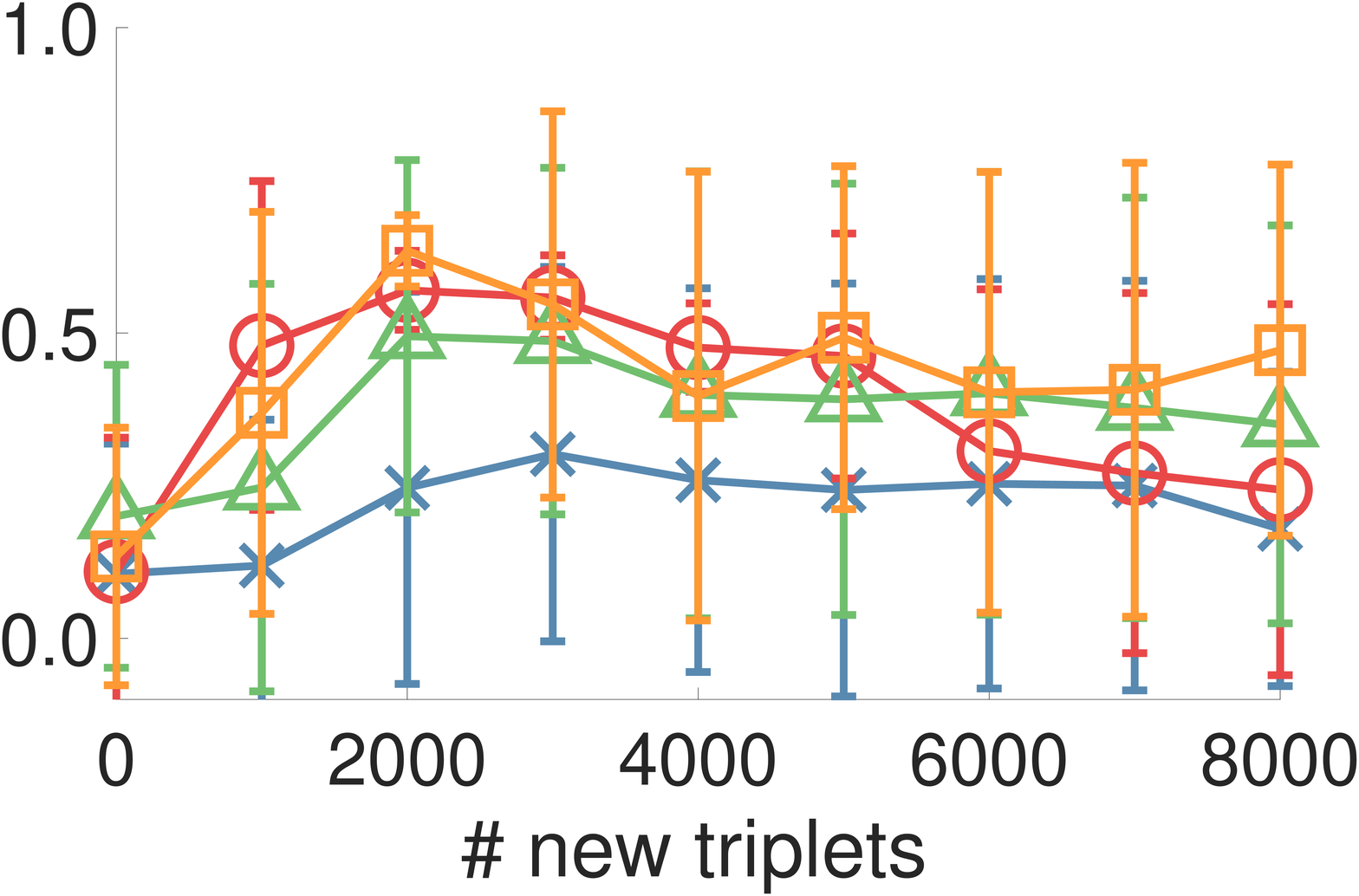}
}
\caption{Breast Cancer. The original dimension is $30$, the embedding dimension is $d = 5$. } \label{fig:Cancer}
\end{figure*}

\begin{figure*}
	 	\subcaptionbox*{}[.175\textwidth]{
	 		\begin{minipage}{.175\textwidth}
	 			i) Triplet Prediction
	 		\end{minipage}
	 		\vspace{40pt}
	 	}\hspace{40pt}
	 	\subcaptionbox*{}[.255\textwidth]{
	 		\centering
	 		\includegraphics[width=.255\textwidth]{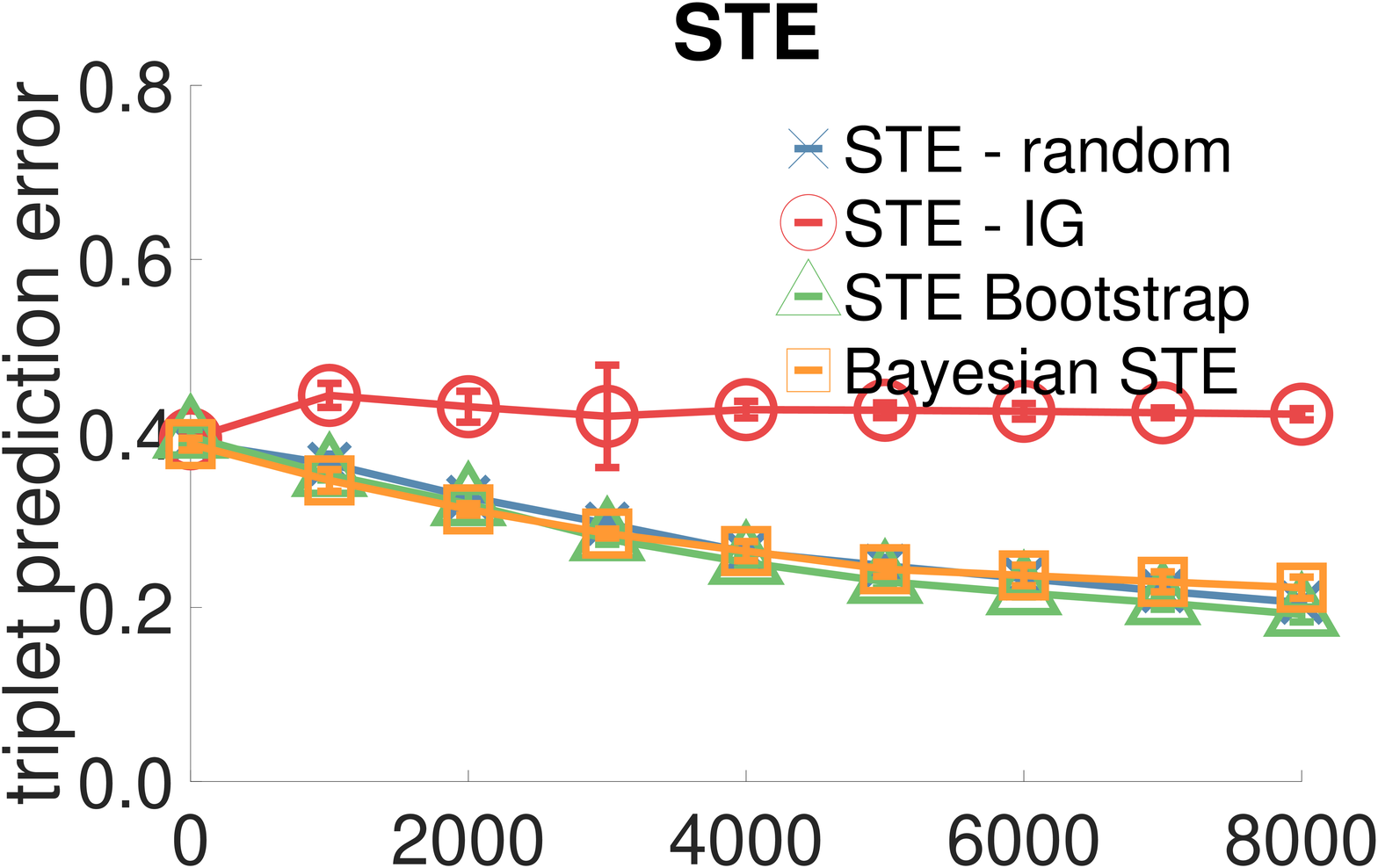}
	 	}\hspace{45pt}
	 	\subcaptionbox*{}[.255\textwidth]{
	 		\centering
	 		\includegraphics[width=.255\textwidth]{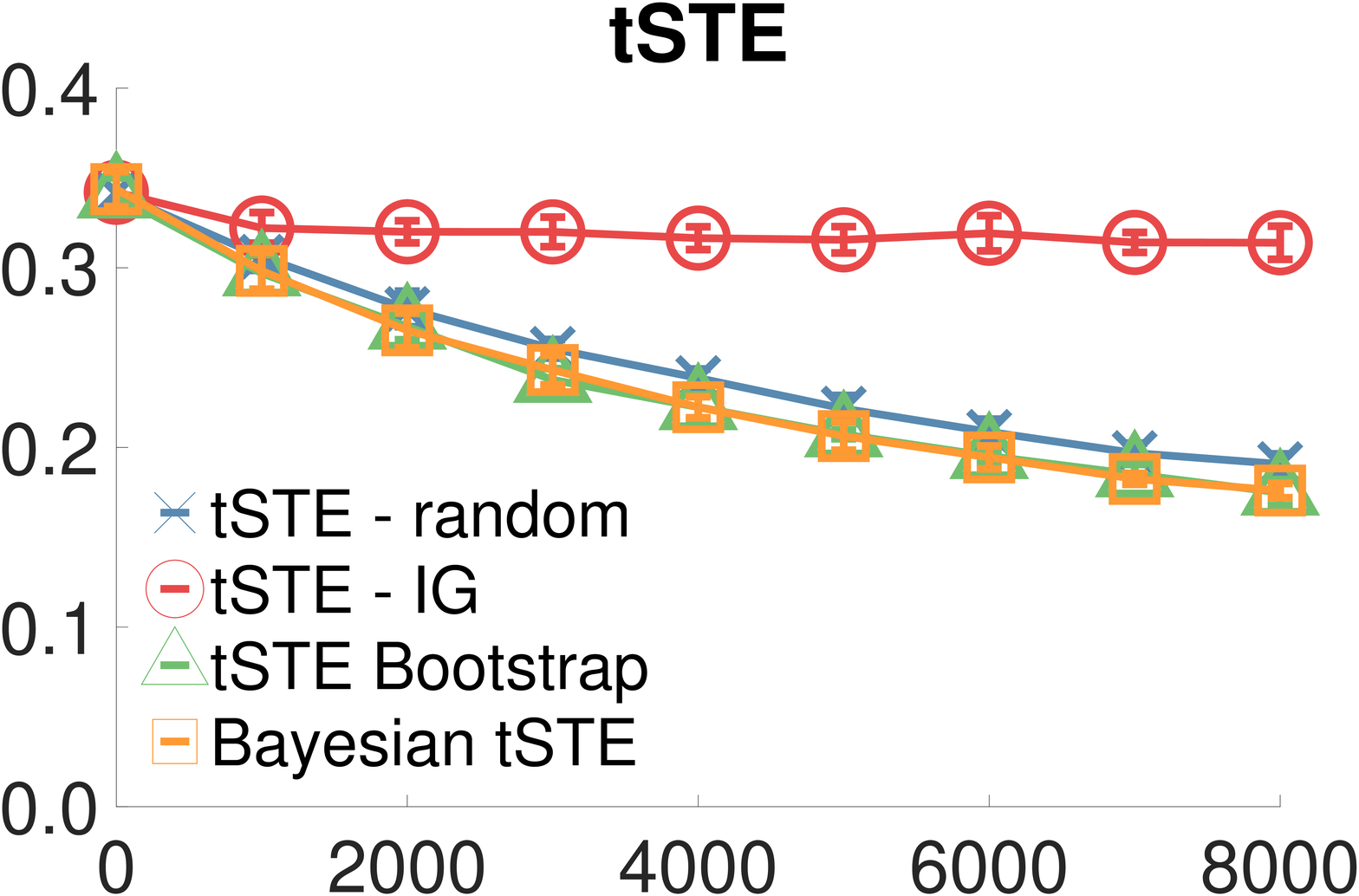}
	 	}\\
 	\subcaptionbox*{}[.175\textwidth]{
 		\begin{minipage}{.175\textwidth}
 			ii) Classification
 		\end{minipage}
 		\vspace{40pt}
 	}\hspace{40pt}
	 	\subcaptionbox*{}[.255\textwidth]{
	 		\centering
	 		\includegraphics[width=.255\textwidth]{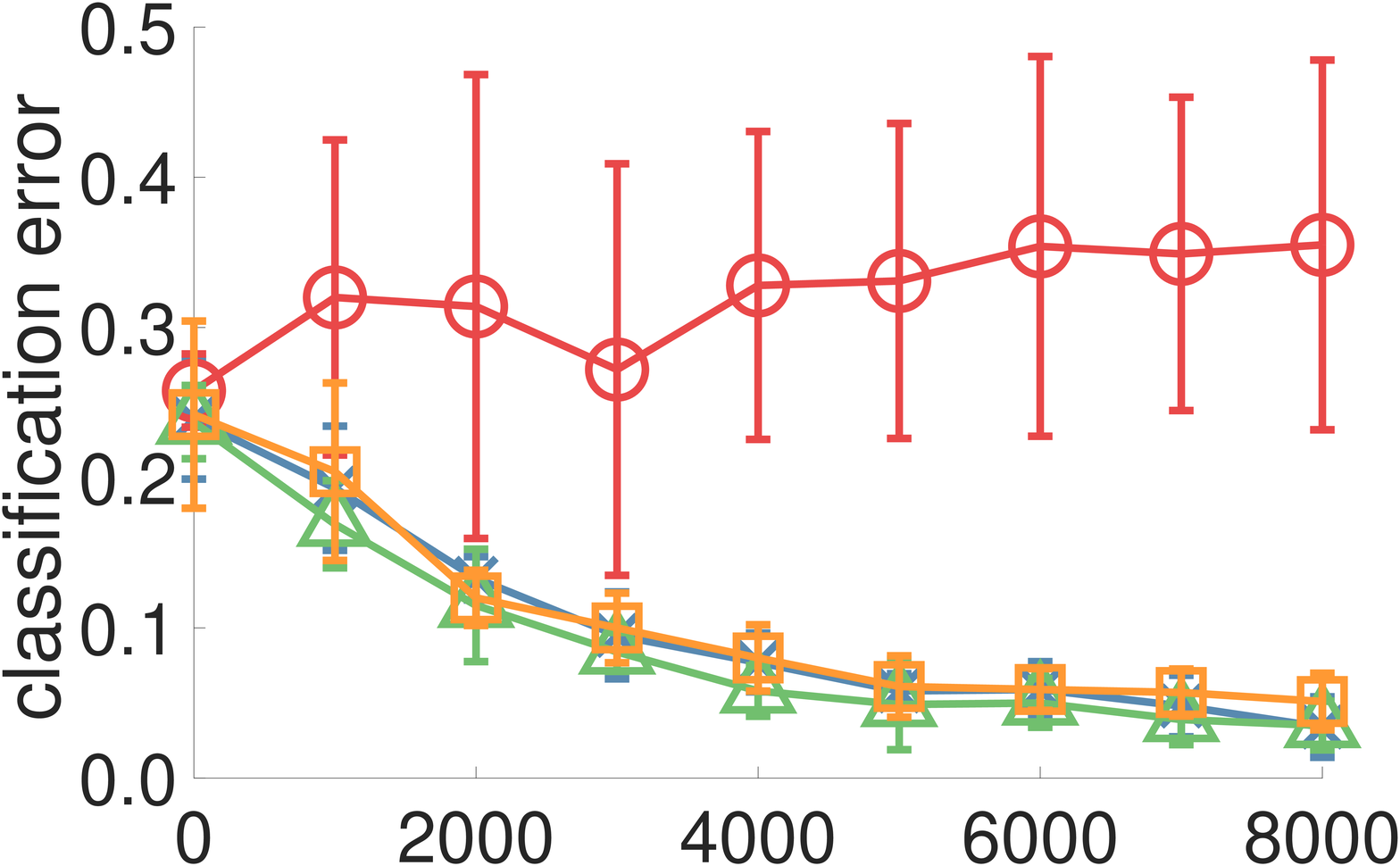}
	 	}\hspace{45pt}
	 	\subcaptionbox*{}[.255\textwidth]{
	 		\centering
	 		\includegraphics[width=.255\textwidth]{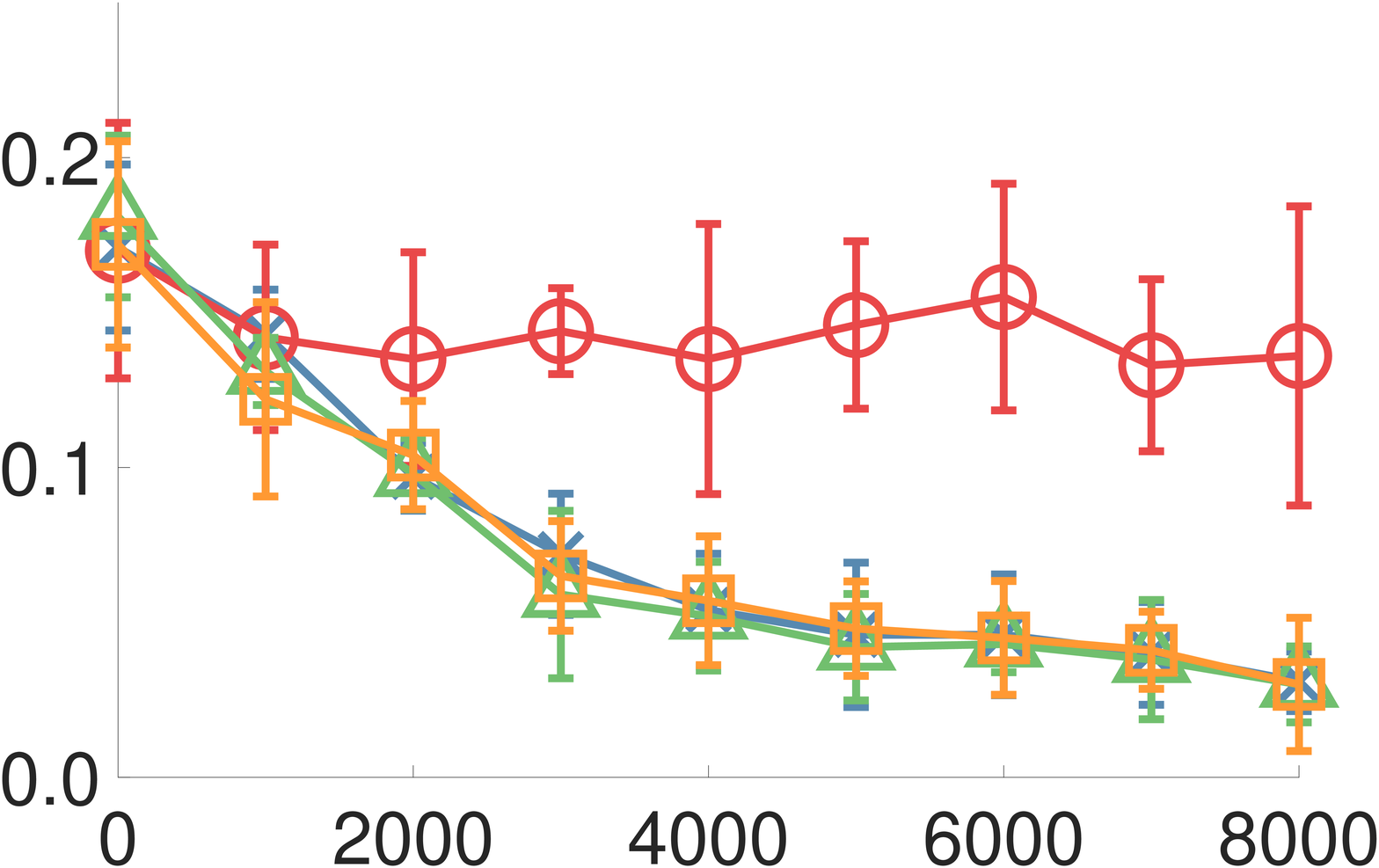}
	 	}\\
 	\subcaptionbox*{}[.175\textwidth]{
 		\begin{minipage}{.175\textwidth}
 			iii) Clustering
 		\end{minipage}
 		\vspace{50pt}
 	}\hspace{40pt}
	 	\subcaptionbox{Comparing STE with its active approaches.}[.255\textwidth]{
	 		\centering
	 		\includegraphics[width=.255\textwidth]{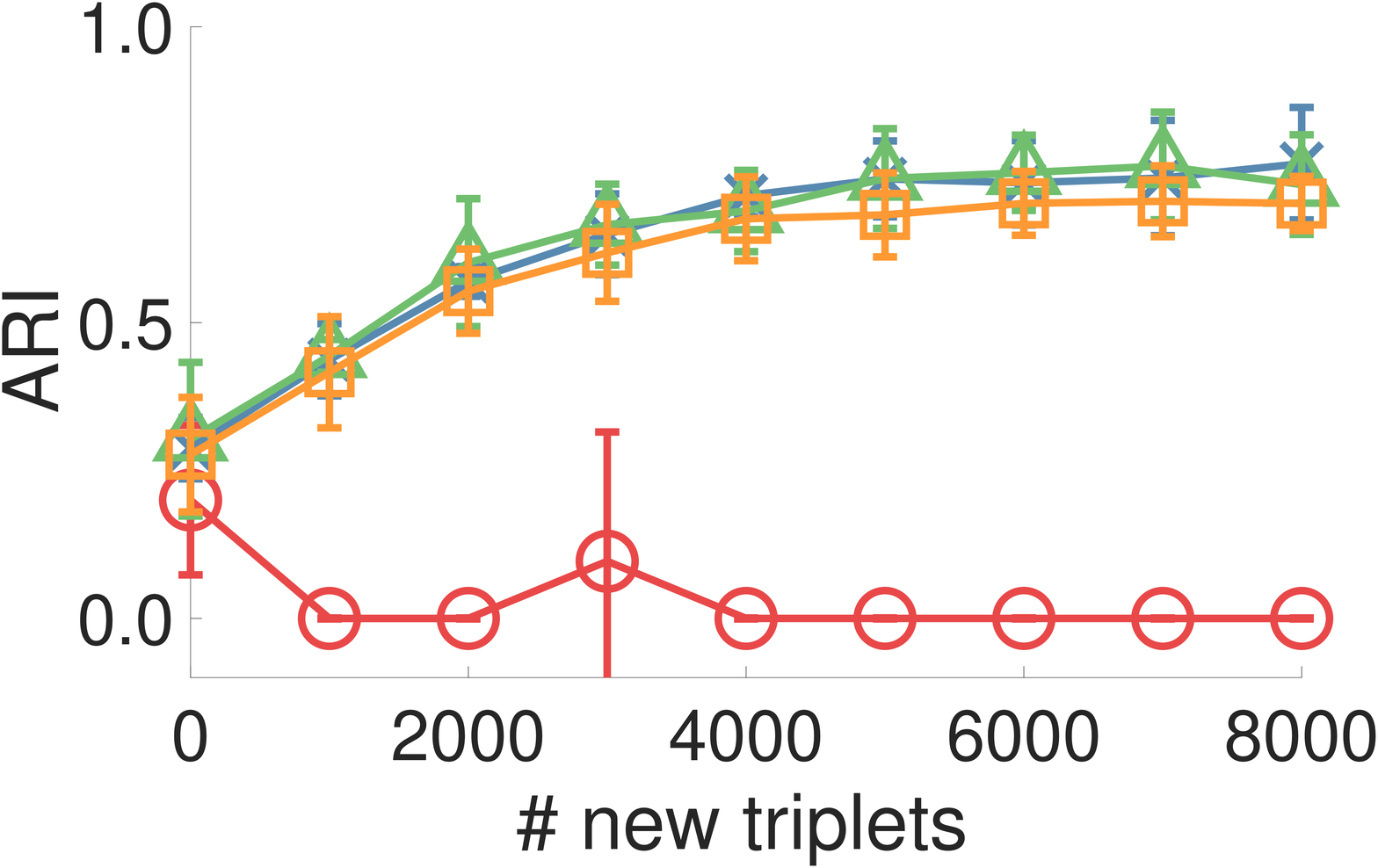}
	 	}\hspace{45pt}
	 	\subcaptionbox{ Comparing $t$-STE with its active approaches.}[.255\textwidth]{
	 		\centering
	 		\includegraphics[width=.255\textwidth]{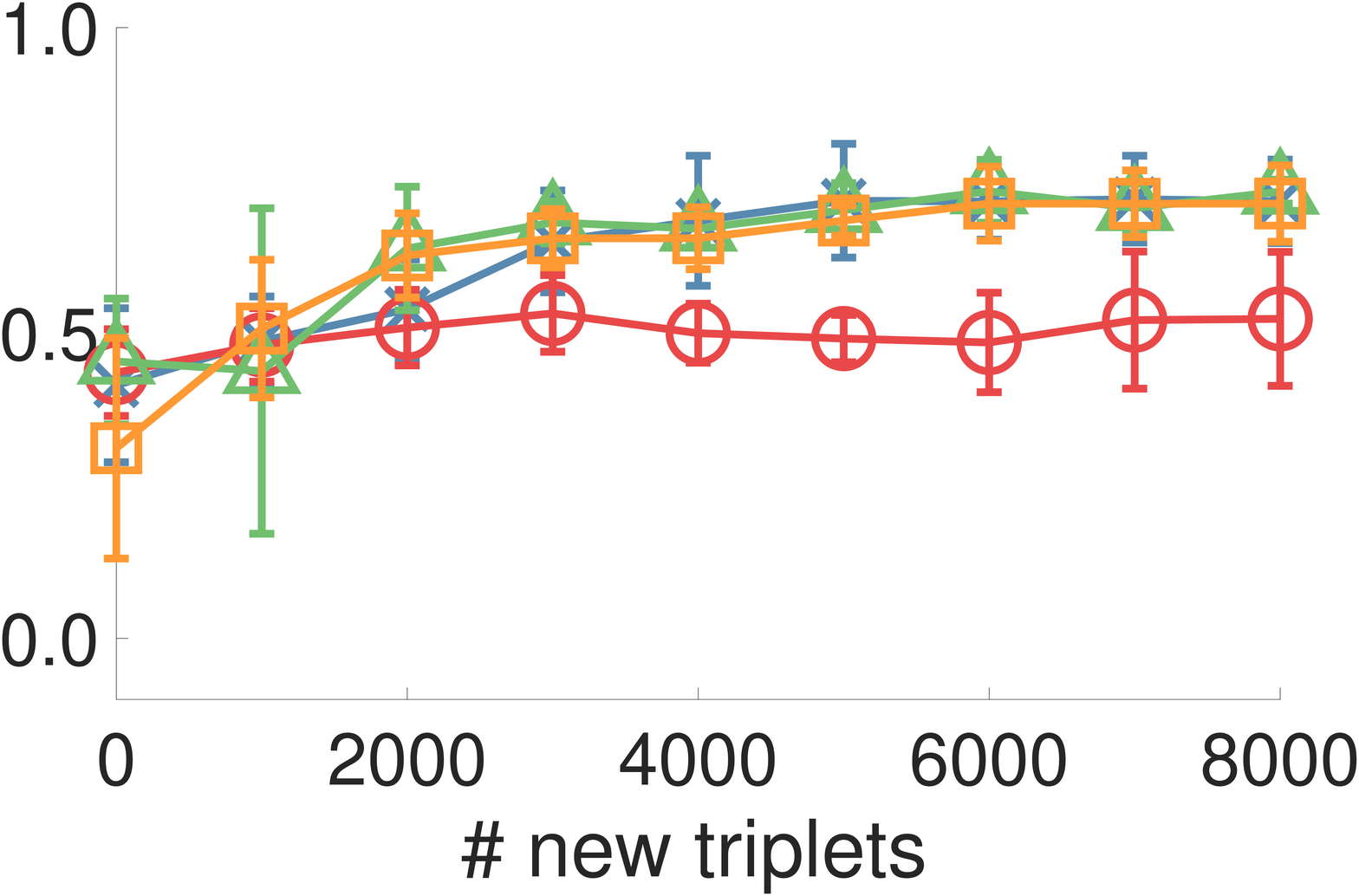}
	 	}
	 	\caption{MNIST. The original dimension is $784$, the embedding dimension is $d = 5$.  } \label{fig:MNIST}
	 \end{figure*}

\begin{figure*}
 	\subcaptionbox*{}[.175\textwidth]{
 		\begin{minipage}{.175\textwidth}
 			i) Triplet Prediction
 		\end{minipage}
 		\vspace{40pt}
 	}\hspace{40pt}
 	\subcaptionbox*{}[.255\textwidth]{
 		\centering
 		\includegraphics[width=.255\textwidth]{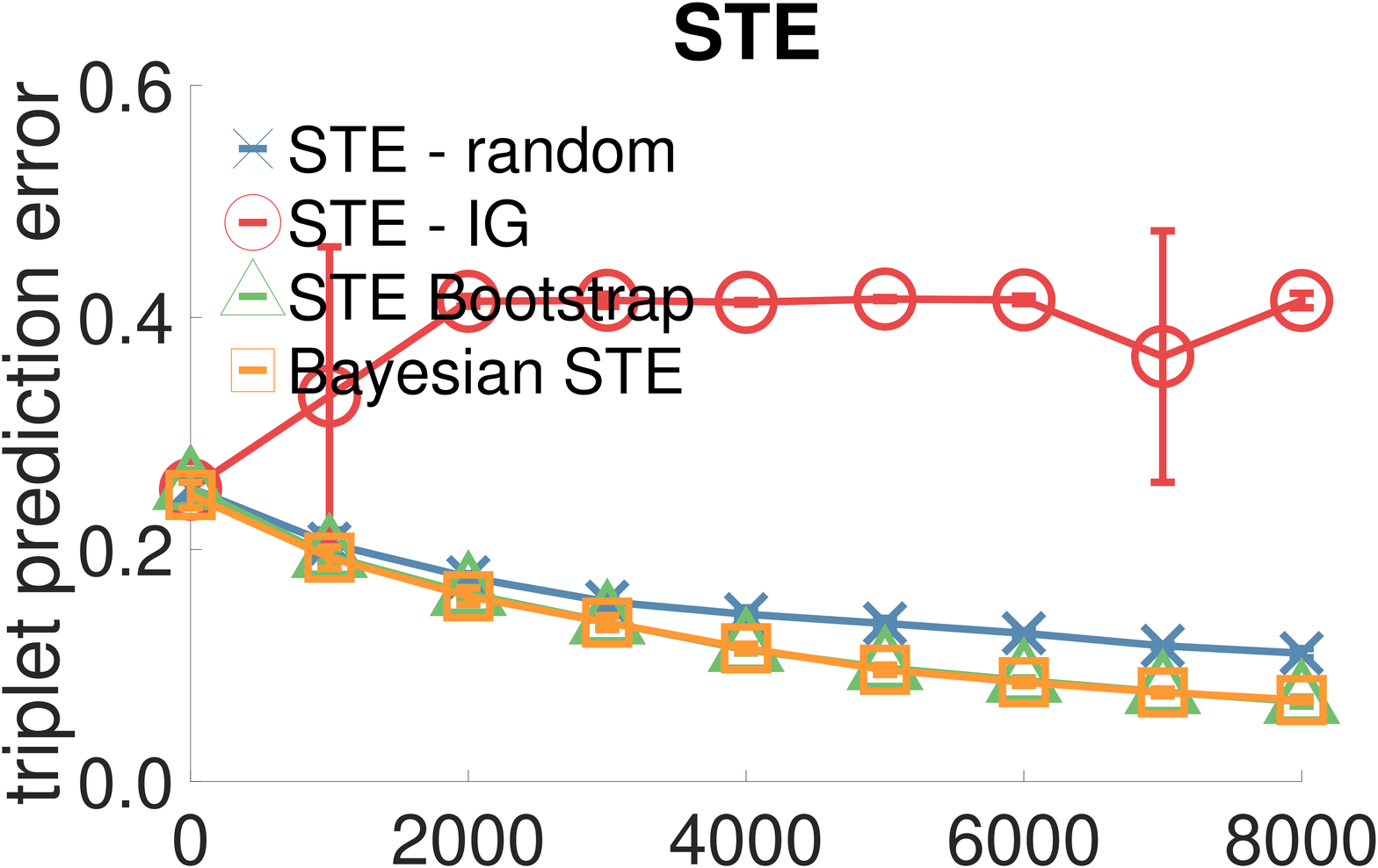}
 	}\hspace{45pt}
 	\subcaptionbox*{}[.255\textwidth]{
 		\centering
 		\includegraphics[width=.255\textwidth]{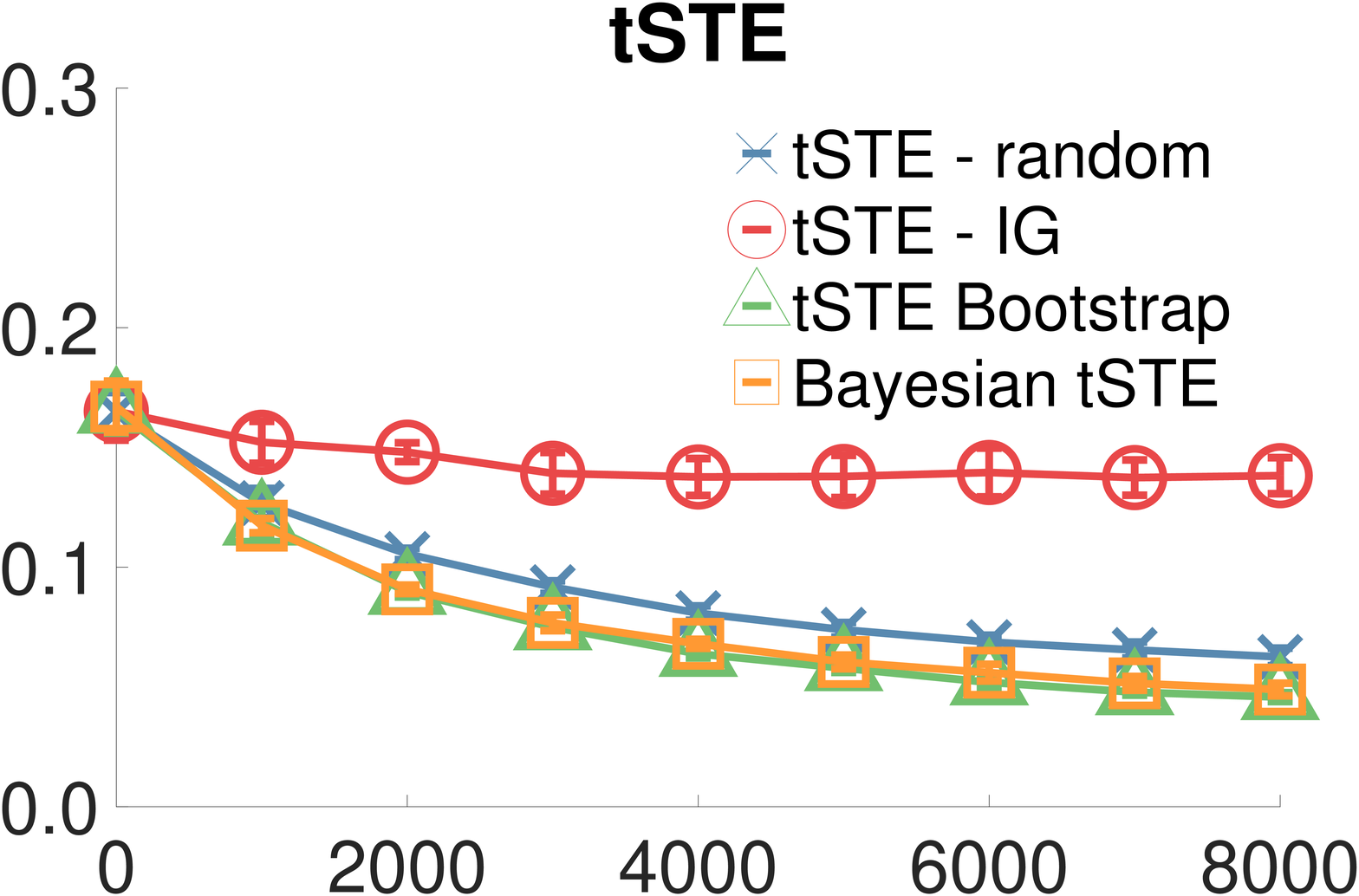}
 	}\\
 	\subcaptionbox*{}[.175\textwidth]{
 		\begin{minipage}{.175\textwidth}
 			ii) Classification
 		\end{minipage}
 		\vspace{40pt}
 	}\hspace{40pt}
 	\subcaptionbox*{}[.255\textwidth]{
 		\centering
 		\includegraphics[width=.255\textwidth]{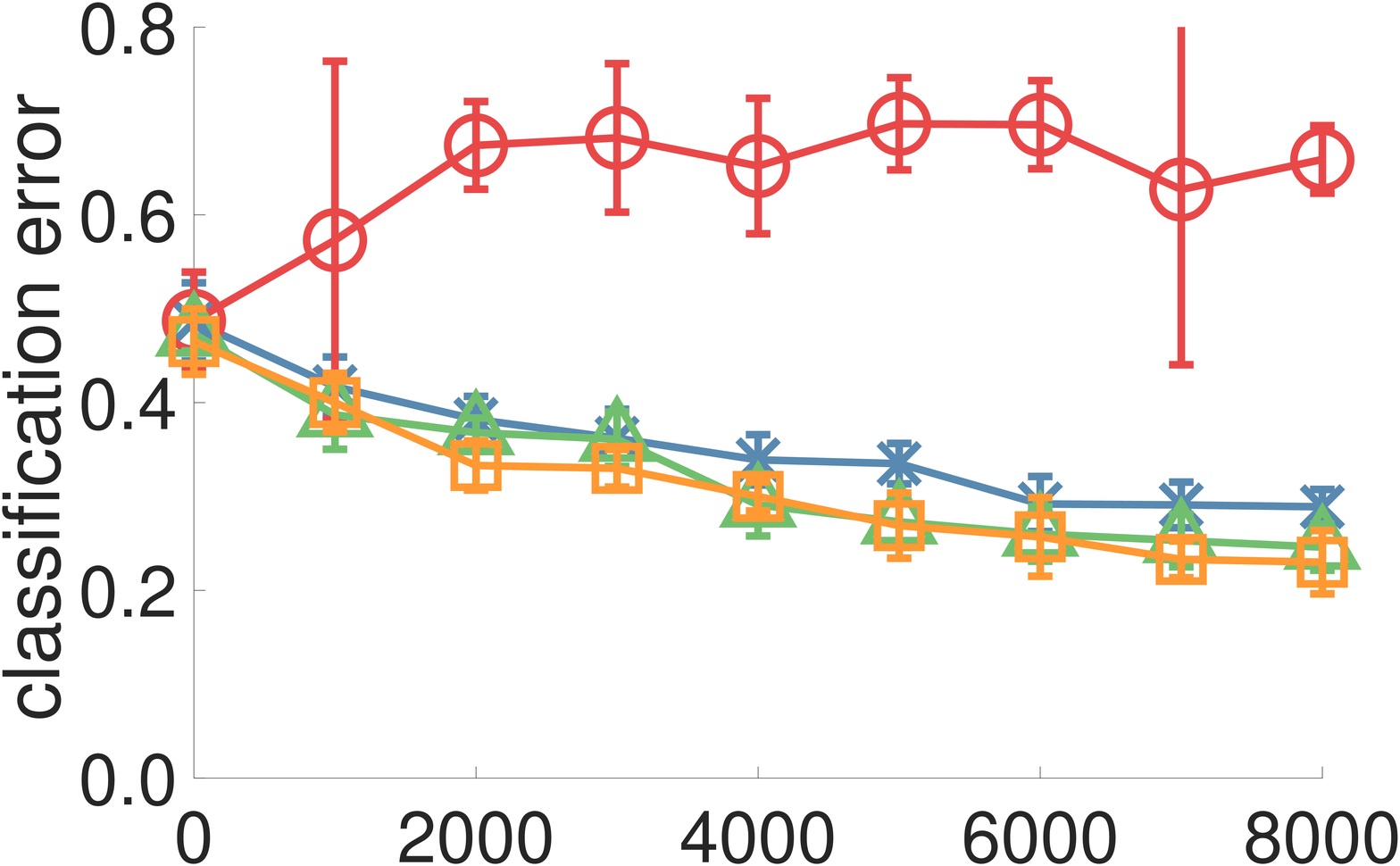}
 	}\hspace{45pt}
 	\subcaptionbox*{}[.255\textwidth]{
 		\centering
 		\includegraphics[width=.255\textwidth]{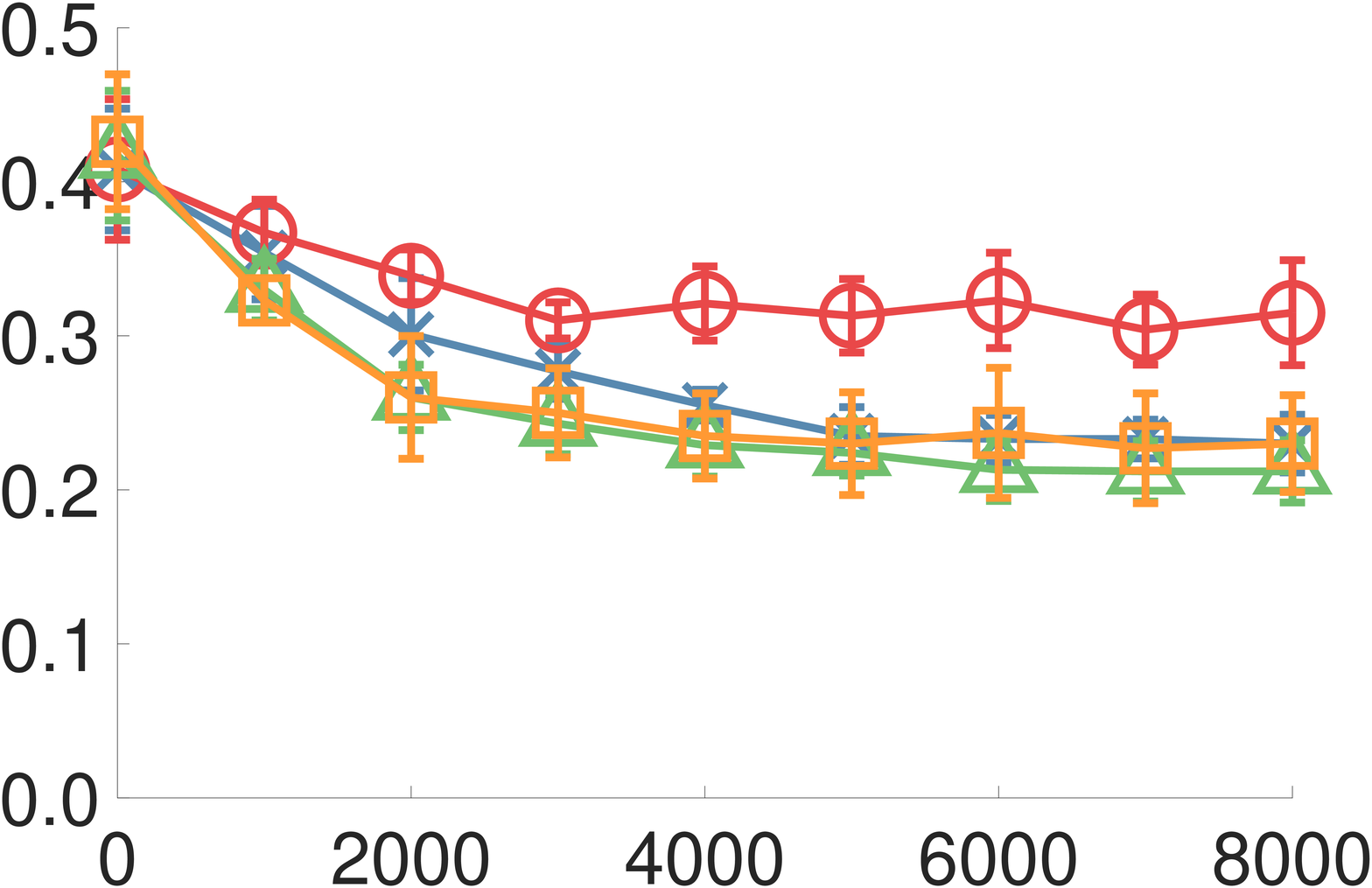}
 	}\\
 	\subcaptionbox*{}[.175\textwidth]{
 		\begin{minipage}{.175\textwidth}
 			iii) Clustering
 		\end{minipage}
 		\vspace{50pt}
 	}\hspace{40pt}
 	\subcaptionbox{Comparing STE with its active approaches.}[.255\textwidth]{
 		\centering
 		\includegraphics[width=.255\textwidth]{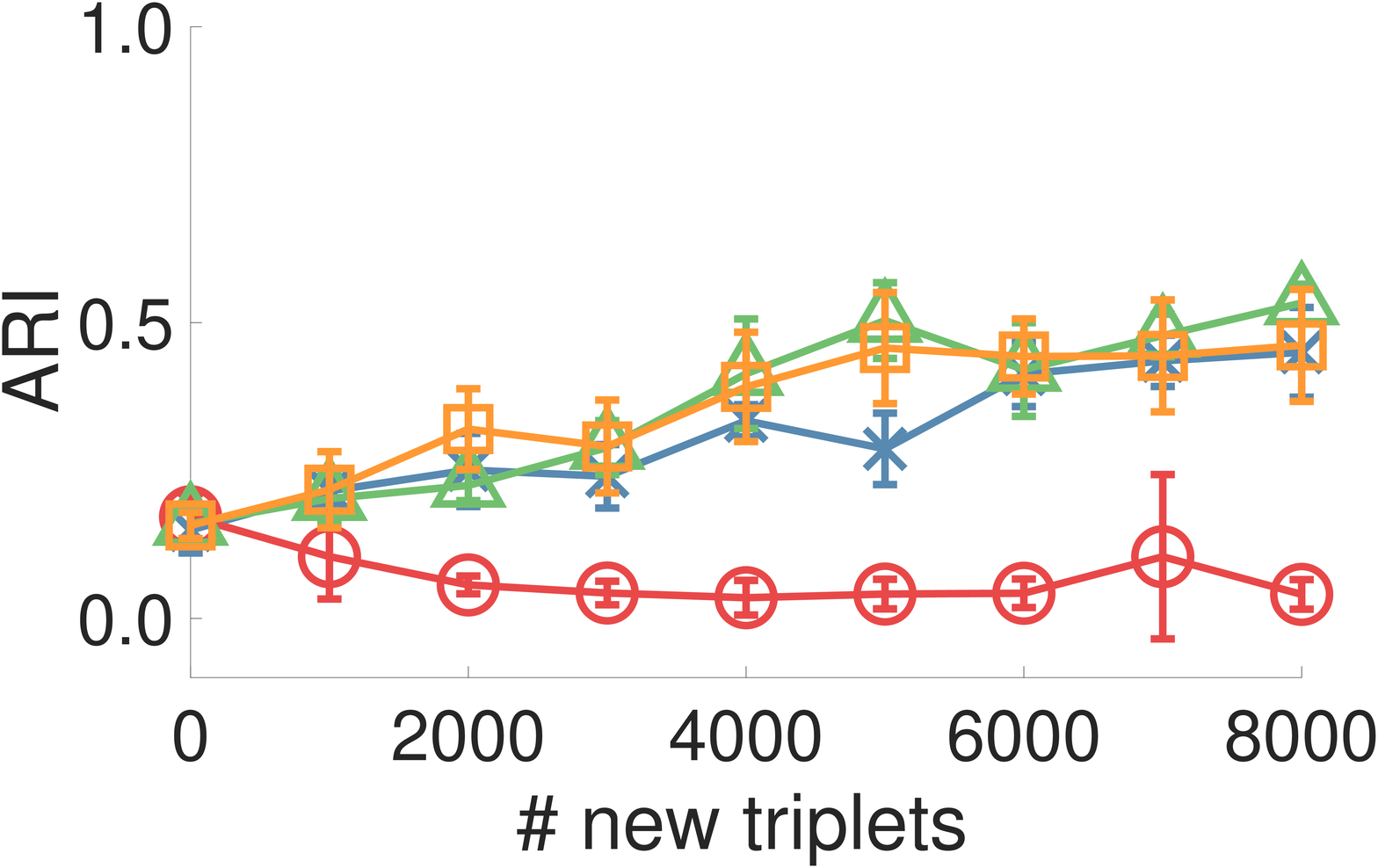}
 	}\hspace{45pt}
 	\subcaptionbox{ Comparing $t$-STE with its active approaches.}[.255\textwidth]{
 		\centering
 		\includegraphics[width=.255\textwidth]{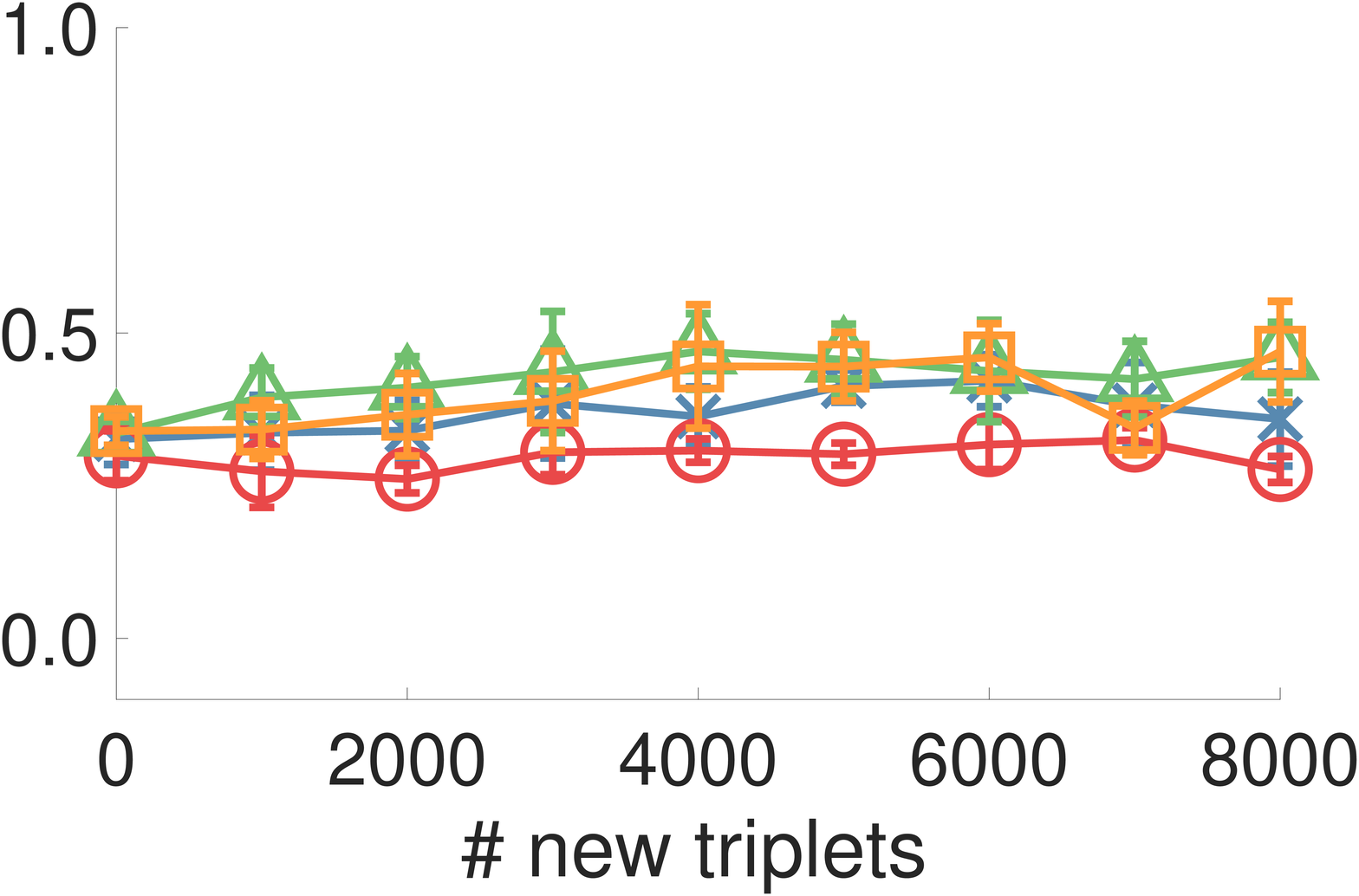}
 	}
 		\caption{Landsat Satellite. The original dimension is $36$, the embedding dimension is $d = 5$.  } \label{fig:Satellite}
 \end{figure*}

	For the active learning experiment described in Section~\ref{sec:UsingEstimates}, we use three data sets --- MNIST, Breast Cancer, and Landsat Satellite (see Table \ref{tab:Datasets}). Here, we report all results for all three data sets. For Breast Cancer see Figure~\ref{fig:Cancer}, for MNIST see Figure~\ref{fig:MNIST}, and for Landsat Satellite see Figure~\ref{fig:Satellite}. 
	
	Recall, that we use $n = 200$ points and the noise level is $\sigma = 0.1$. We start with a random seed of $2,000$ triplets and repeatedly add $1,000$ triplets to the training set by determining the $1,000$ most uncertain triplet comparisons. In the case of the information gain criterion, we add those $1,000$ triplets that have the highest information gain. We measure the embeddings after each step by performing the three tasks: triplet prediction, classification and clustering. 
	
	For the following results we denote our bootstrap approach based on the $t$-STE embedding method with \emph{$t$-STE Bootstrap} and our bootstrap approach based on the GNMDS embedding method with \emph{GNMDS Bootstrap}. By \emph{Bayesian $t$-STE} we denote our Bayesian approach using the $t$-STE likelihood (\ref{eq:tSTELikelihood}).
	
\begin{figure*}
		\centering
		\subcaptionbox*{}[.3\textwidth]{
			\centering
			\includegraphics[width=.3\textwidth]{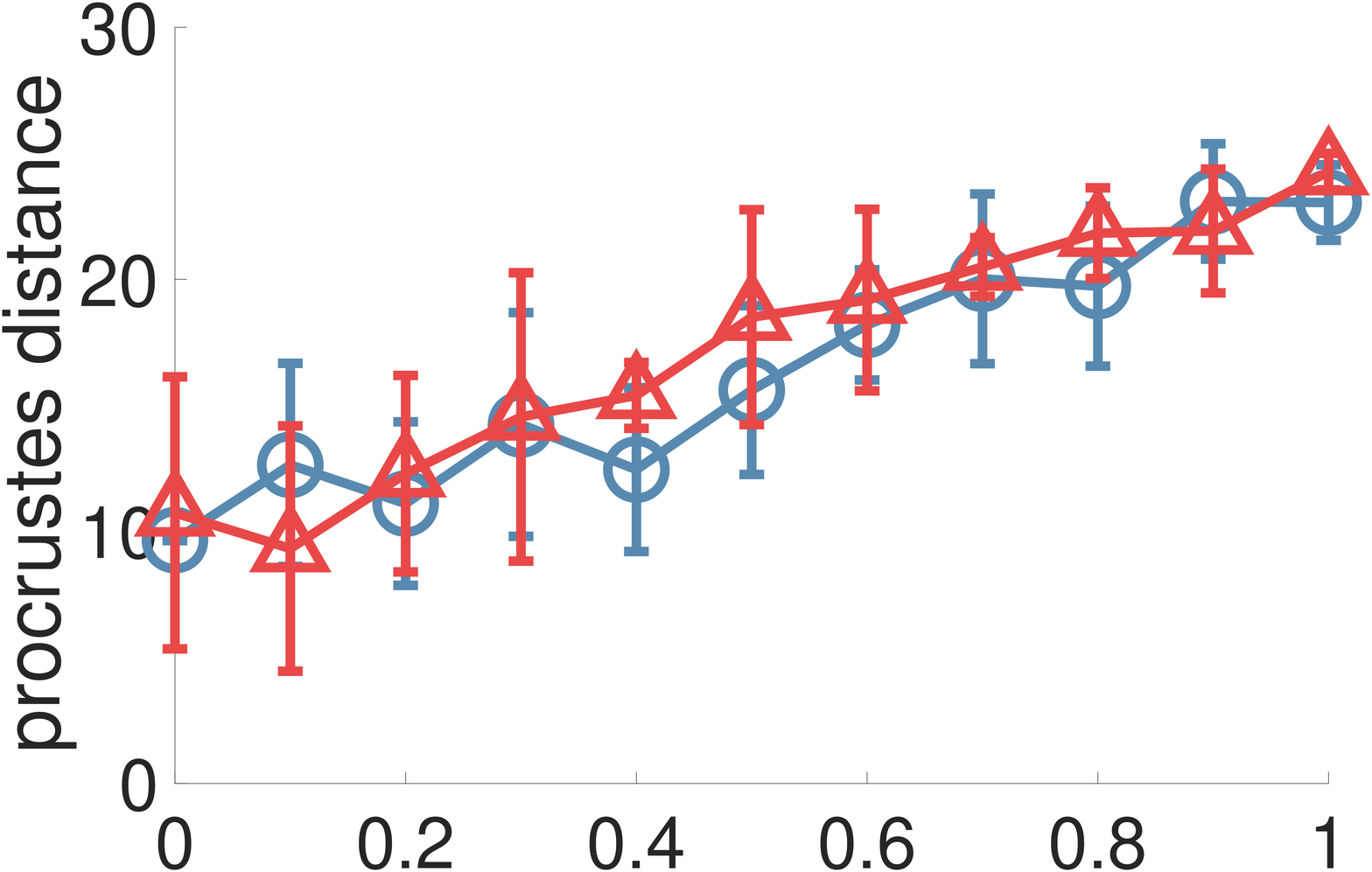}
		}\hspace{15pt}
		\subcaptionbox*{}[.3\textwidth]{
			\centering
			\includegraphics[width=.3\textwidth]{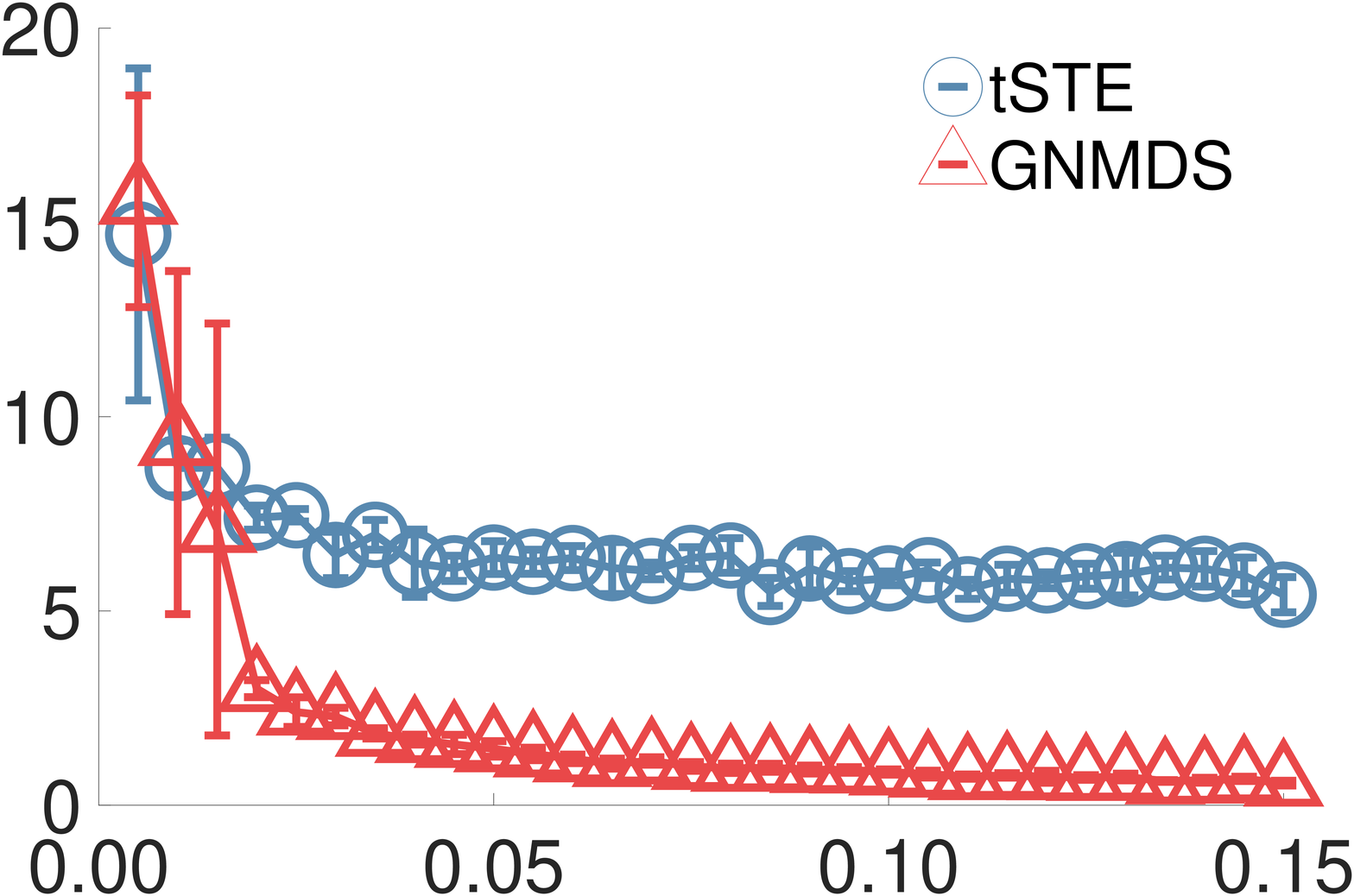}
		}\\
		\subcaptionbox{Increasing noise.\label{fig:increasingNoise_supp}}[.3\textwidth]{
			\centering
			\includegraphics[width=.3\textwidth]{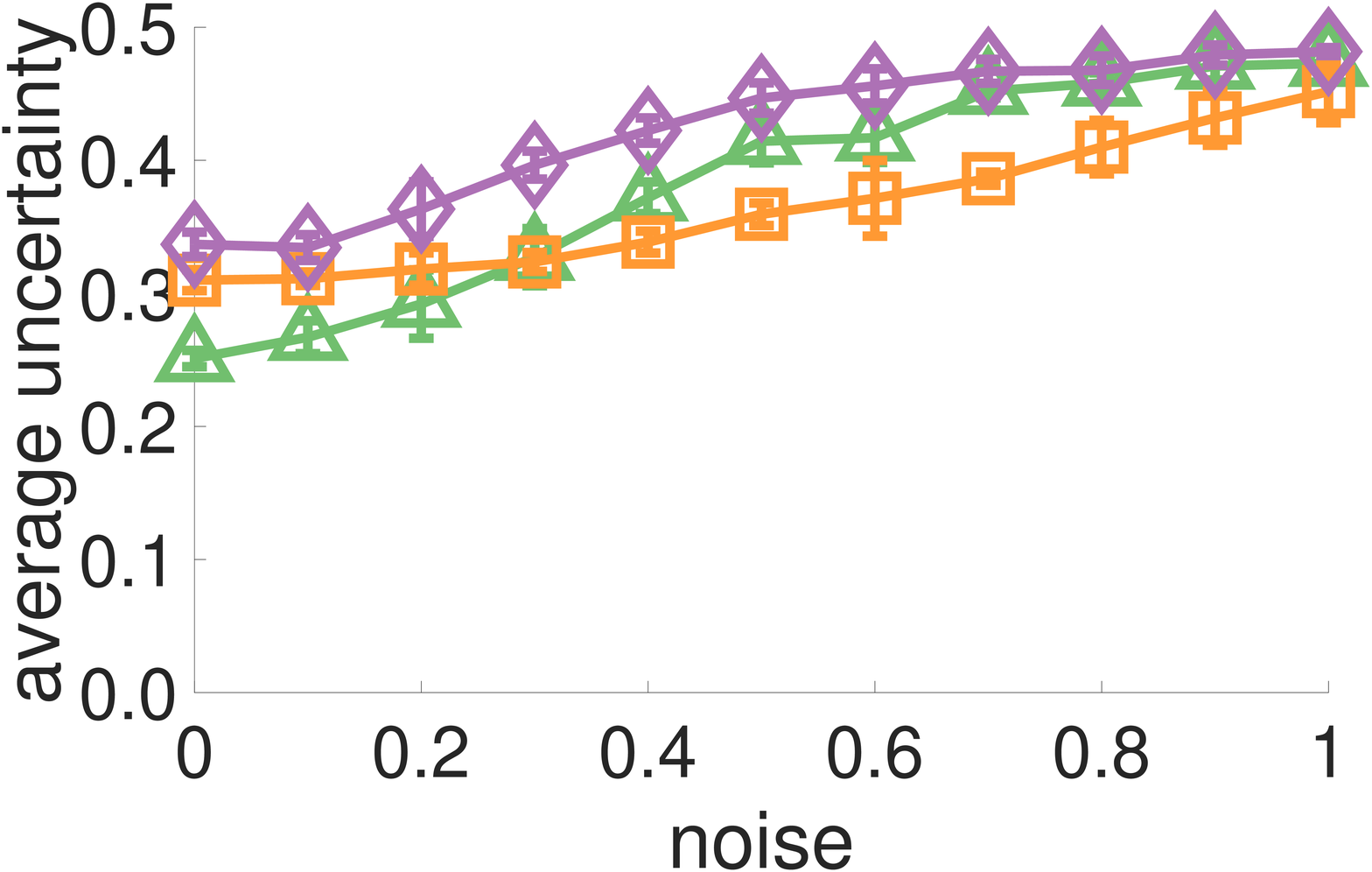}
		}\hspace{15pt}
		\subcaptionbox{Increasing fraction of triplets.\label{fig:increasingTriplets_supp}}[.3\textwidth]{
			\centering
			\includegraphics[width=.3\textwidth]{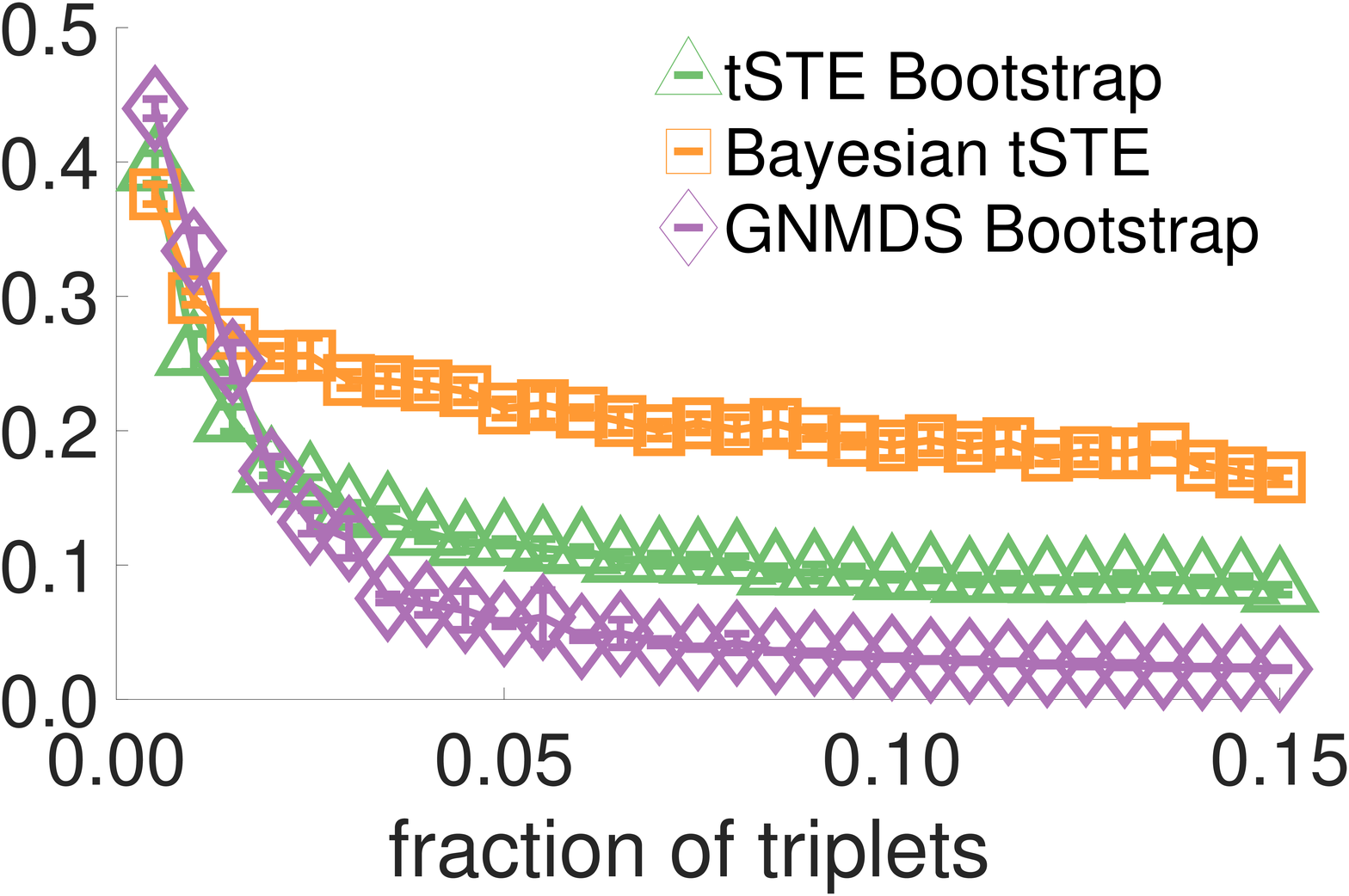}
		}
		\caption{We selected $50$ points from the mixture of Gaussians. The top row shows the mean and standard deviation of the embedding error measured by the procrustes distance by the standard embedding algorithms.  The second row shows the mean and standard deviation of the average uncertainty. a) We selected one percent of all triplets. Increasing the noise induces a higher embedding error. Simultaneously, the uncertainty estimates increase and converge to 0.5 (completely uncertain).  b) Here, we have no noise in the training triplets. The error decreases when more triplets are available for the embedding algorithm. Along with it, the uncertainty decreases.}
	\end{figure*}

\end{appendices}

\end{document}